\documentclass[runningheads]{llncs}

\usepackage{eccv}

\usepackage{eccvabbrv}

\usepackage{graphicx}
\usepackage{booktabs}
\usepackage{bm}
\usepackage{mathtools}
\usepackage{multirow}
\usepackage{makecell}
\usepackage{float}
\usepackage{units}

\usepackage[accsupp]{axessibility}  

\usepackage[pagebackref,breaklinks,colorlinks,citecolor=eccvblue]{hyperref}

\usepackage{orcidlink}

\newcommand{\red}[1]{{\color{red}#1}}

\usepackage{xcolor}
\usepackage{colortbl}
\definecolor{color3}{RGB}{255, 255, 200}
\definecolor{color2}{RGB}{255, 220, 200}
\definecolor{color1}{RGB}{255, 181, 163}
\newcommand{\cc}[1]{\cellcolor{color#1}}

\begin{document}

\title{Analytic-Splatting: Anti-Aliased 3D Gaussian Splatting via Analytic Integration}

\titlerunning{Analytic-Splatting}

\author{Zhihao Liang\inst{1,*} \and
Qi Zhang\inst{2,\dag} \and
Wenbo Hu\inst{2} \and
Lei Zhu\inst{3} \and
Ying Feng\inst{2} \and
Kui Jia\inst{4,\dag}
}

\authorrunning{Zhihao Liang et al.}

\institute{
$^1$South China University of Technology \\
$^2$Tencent AI Lab \quad
$^3$City University of Hong Kong \\
$^4$School of Data Science, The Chinese University of Hong Kong, Shenzhen
\href{https://lzhnb.github.io/project-pages/analytic-splatting/}{https://lzhnb.github.io/project-pages/analytic-splatting/}}

\maketitle

\begin{abstract}
    The 3D Gaussian Splatting (3DGS) gained its popularity recently by combining the advantages of both primitive-based and volumetric 3D representations, resulting in improved quality and efficiency for 3D scene rendering.
    However, 3DGS is not alias-free, and its rendering at varying resolutions could produce severe blurring or jaggies. This is because 3DGS treats each pixel as an isolated, single point rather than as an area, causing insensitivity to changes in the footprints of pixels. Consequently, this discrete sampling scheme inevitably results in aliasing, owing to the restricted sampling bandwidth.
    In this paper, we derive an analytical solution to address this issue. More specifically, we use a conditioned logistic function as the analytic approximation of the cumulative distribution function (CDF) in a one-dimensional Gaussian signal and calculate the Gaussian integral by subtracting the CDFs. We then introduce this approximation in the two-dimensional pixel shading, and present \emph{Analytic-Splatting}, which analytically approximates the Gaussian integral within the 2D-pixel window area to better capture the intensity response of each pixel.
    Moreover, we use the approximated response of the pixel window integral area to participate in the transmittance calculation of volume rendering, making Analytic-Splatting sensitive to the changes in pixel footprint at different resolutions.
    Experiments on various datasets validate that our approach has better anti-aliasing capability that gives more details and better fidelity.
\end{abstract}

\renewcommand{\thefootnote}{\ifcase\value{footnote}\or*\or
 \dag \or(***)\or(****)\or(\#)\or(\#\#)\or(\#\#\#)\or(\#\#\#\#)\or($\infty$)\fi}
\footnotetext[1]{Work was done during an internship at Tencent AI Lab.}
\footnotetext[2]{Corresponding authors.}

\begin{figure}[!htbp]
\centering
\includegraphics[width=1\textwidth]{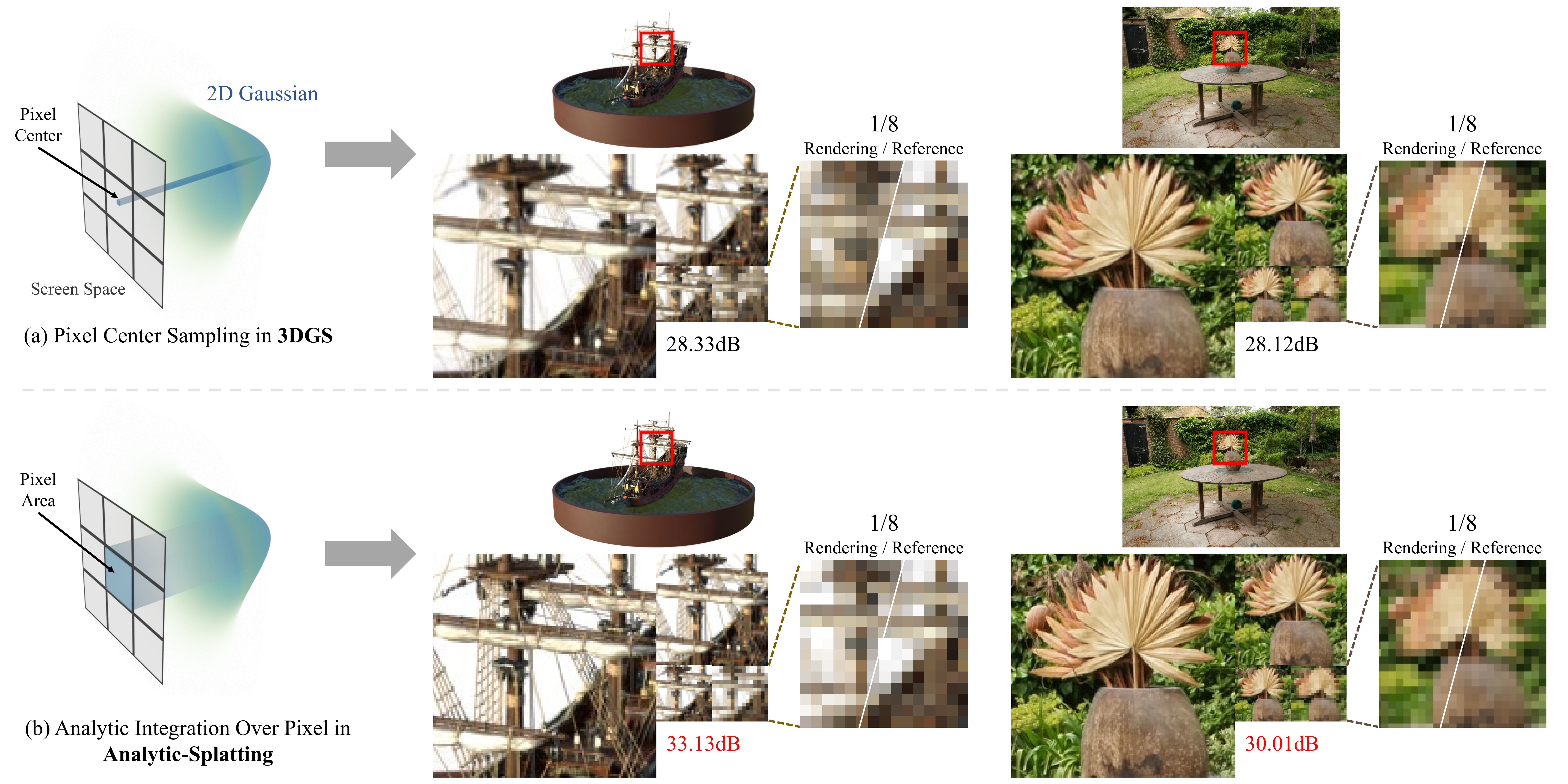}
\caption{For shading a pixel by a Gaussian signal, 3DGS (a) only treats the Gaussian signal value corresponding to the pixel center as the intensity response. Analytic-Splatting (b) instead considers an analytic approximation of the integral over the pixel window area as the intensity response. Compared to 3DGS, Analytic-Splatting has anti-aliasing capability and better detail fidelity.}
\label{fig:teaser}
\vspace{-0.6cm}
\end{figure}

\renewcommand{\thefootnote}{\arabic{footnote}}

\section{Introduction}
\label{sec:intro}

Novel view synthesis of a scene captured from multiple images has achieved great progress due to the rapid advancements of neural rendering. As a prominent representative, Neural Radiance Field (NeRF)~\cite{mildenhall2020nerf} models the scene using a neural volumetric representation, enabling photorealistic rendering of novel views via ray marching. Ray marching trades off rendering efficiency with quality, and subsequent works~\cite{xu2022point, fridovich2022plenoxels, muller2022instant} are proposed to have a better quality-efficiency balance. More recently, 3D Gaussian Splatting (3DGS)~\cite{kerbl20233d} proposes a GPU-friendly differentiable rasterization pipeline that incorporates an explicit point-based representation, achieving high-quality and real-time renderings for novel view synthesis. In contrast to ray marching in NeRF, which renders a pixel by accumulating the radiance of samples along the ray that intersects the image plane at the pixel, 3DGS employs a forward-mapping technique that can be rasterized very efficiently. Specifically, 3DGS represents the scene as a set of anisotropic 3D Gaussians with scene properties; when rendering a pixel, 3DGS orders and projects these 3D Gaussians onto the image plane as 2D Gaussians, and then queries values and scene properties associated with the Gaussians that have overlaps with the pixel, and finally shades the pixel by cumulatively compositing these queried values and properties.

3DGS works for scene representation learning and novel view synthesis at constant resolutions; however, its performance degrades greatly either when the multi-view images are captured at varying distances, or when the novel view to be rendered has a resolution different from those of the captured images. The main reason is that the footprint \footnote{The footprint is defined as the ratio between the pixel window area in screen space and its covered Gaussian signals region in the world space.} of the pixel changes at different resolutions and 3DGS is insensitive to such changes since it treats each pixel as an isolated point (\ie merely pixel center) when retrieving the corresponding Gaussian values; Fig. \ref{fig:teaser}\red{a} gives an illustration. 
As a result, 3DGS could produce significant artifacts (\eg blurry or jaggies) especially when pixel footprints change drastically (\eg synthesizing novel views with zooming-in and zooming-out effects).

By delving into the details, we know that 3DGS represents a continuous signal in the image space as a set of $\alpha$-blended 2D Gaussians, and the pixel shading is a process of integrating the signal response within each pixel area; artifacts in 3DGS are caused by the limited sampling bandwidth for the Gaussians that retrieves erroneous responses, especially when the pixel footprint changes drastically. It is possible to increase sampling bandwidth (\ie via super sampling) or use prefiltering techniques to alleviate this problem; for example, Mip-Splatting~\cite{yu2023mip} employs the prefiltering technique and presents a hybrid filtering mechanism to regularize the high-frequency components of 2D and 3D Gaussians to achieve anti-aliasing. While Mip-Splatting overcomes most aliasing in 3DGS, it is limited in capturing details and synthesizes over-smoothing results. Consequently, solving the integral of Gaussian signals within the pixel window area as intensity responses is crucial for both anti-aliasing and capturing details.

In this paper, we revisit pixel shading in 3DGS and introduce an analytic approximation of the window integral response of Gaussian signals for anti-aliasing. Rather than discrete sampling in 3DGS and prefiltering in Mip-Splatting, we analytically approximate the integral within each pixel area as shown in \cref{fig:teaser}\red{b}. We term our method as \textit{Analytic-Splatting}. Compared with Mip-Splatting, which approximates the pixel window as a 2D Gaussian low-pass filter, our proposed method does not suppress the high-frequency components in Gaussian signals and can better preserve high-quality details.
Experiments show that our method removes the aliasing existing in 3DGS and other methods while synthesizing more details with better fidelity. We summarize our contributions as follows. 
\begin{itemize}
    \item We revisit the causes of aliasing in 3D Gaussian Splatting from the perspective of signal window response and derive an analytic approximation of the window response for Gaussian signals;
    \item Based on the derivation, we present \emph{Analytic-Splatting} that improves the pixel shading in 3D Gaussian Splatting to achieve anti-aliasing and better detail fidelity.
    \item Our experiments on challenging datasets demonstrate the superiority of our method to other approaches in terms of anti-aliasing and synthesizing results.
\end{itemize}

\section{Related Works}
\label{sec:related_works}
\noindent\textbf{Neural Rendering.} Recently, neural rendering techniques exemplified by Neural Radiance Field (NeRF)~\cite{mildenhall2020nerf} have achieved impressive results in novel view synthesis, and further enhanced several advanced tasks~\cite{wang2021neus, yariv2021volume, lin2021barf, huang2022hdr, ma2022deblur, park2021nerfies}.
Nevertheless, the backward-mapping volume rendering used in NeRF hinders the real-time rendering performance, restricting the application prospects of NeRF. While several NeRF variants adopt efficient sampling strategies~\cite{yu2021plenoctrees, muller2022instant, li2022nerfacc} or use explicit/hybrid representations~\cite{fridovich2022plenoxels, sun2022direct, chen2022tensorf, gao2023strivec} with higher capacities, they still suffer from the tough sampling problem and struggle with real-time rendering.
To overcome these limitations, 3DGS~\cite{kerbl20233d} employs forward mapping volume rendering technology and implements GPU-friendly tile-based rasterization to achieve real-time rendering and impressive rendering results.
Due to its real-time rendering capability and impressive rendering performance, 3DGS has been widely used in advanced tasks such as Human/Avatar modeling~\cite{saito2023relightable, hu2023gaussianavatar, zheng2023gps, zielonka2023drivable}, surface reconstruction~\cite{guedon2023sugar, chen2023neusg}, inverse rendering~\cite{liang2023gs, jiang2023gaussianshader, shi2023gir}, physical simulation~\cite{xie2023physgaussian, feng2024gaussian}, etc.
Although rasterization makes 3DGS avoid tough sampling problems along rays and achieve promising results, it also introduces aliasing caused by restricted sampling bandwidth when shading pixels using 2D Gaussians.
And the aliasing will be noticeable when the pixel footprint changes drastically (\eg zooming in and out). In this paper, we study the errors introduced by the discrete sampling scheme used in 3DGS and introduce our advanced resolution.

\noindent\textbf{Anti-Aliasing.}
Aliasing is the phenomenon of overlapping frequency components when the discrete sampling rate is below the Nyquist rate.
Anti-aliasing is critical for rendering high-fidelity results, which has been extensively explored in the computer graphics and vision community~\cite{akeley1993reality,wu2019accurate,kuznetsov2021neumip}.
In the neural rendering context, MipNeRF~\cite{barron2021mip} and Zip-NeRF~\cite{barron2023zip} pioneer the use of prefiltering and multi-sampling to address the aliasing issue in neural radiance fields (NeRF).
Recent works also explored the anti-aliased NeRF for unbounded scenes~\cite{barron2022mip}, efficient reconstruction~\cite{hu2023tri}, and surface reconstruction~\cite{zhuang2023anti}.
All these works are built upon the backward-mapping volume rendering to consider the pixel footprint, by replacing the original ray-casting with cone-casting.
However, the backward-mapping volume rendering is too computationally expensive to achieve real-time rendering.
On the other hand, 3DGS~\cite{kerbl20233d} introduced real-time forward-mapping volume rendering but suffers from aliasing artifacts due to the discrete sampling during shading pixels using projected Gaussians.
To this end, Mip-Splatting~\cite{yu2023mip} presents a hybrid filtering mechanism to restrict the high-frequency components of 2D and 3D Gaussians to achieve anti-aliasing.
Nevertheless, this low-pass filtering strategy hinders the capability to preserve high-quality details.
In contrast, our approach introduces an analytic approximation of the integral within the pixel area to better capture the intensity response of each pixel, harvesting both aliasing-free and detail-preserving rendering results.


\section{Preliminary}
\label{sec:preliminary}
In this section, we give the technical background necessary for presentation of our proposed method.

3D Gaussian Splatting (3DGS) explicitly represents 3D scene as a set of points $\{\bm{p}_i\}^N_{i=1}$. Given any point $\bm{p} \in \{\bm{p}_i\}^N_{i=1}$, 3DGS models it as a 3D Gaussian signal with mean vector $\bm{\mu}$ and covariance matrix $\bm{\Sigma}$:
\begin{equation}
\vspace{-0.1cm}
g^\text{3D}(\bm{x} | \bm{\mu}, \bm{\Sigma}) =
\exp\left(
-\frac{1}{2}(\bm{x}-\bm{\mu})^\top\bm{\Sigma}^{-1}(\bm{x}-\bm{\mu})
\right),
\label{eq:gaussian_3d}
\vspace{-0.1cm}
\end{equation}
where $\bm{\mu} \in \mathbb{R}^3$ is the position of point $\bm{p}$, and $\bm{\Sigma} \in \mathbb{R}^{3\times 3}$ is an anisotropic covariance matrix, which is factorized into a scaling matrix $\bm{S}$ and a rotation matrix $\bm{R}$ as
$\bm{\Sigma} = \bm{R}\bm{S}\bm{S}^\top\bm{R}^\top$.
Note that $\bm{S}$ indicates a $3\times3$ diagonal matrix and $\bm{R}$ refers to a $3\times3$ matrix constructed from a unit quaternion $\bm{q}$.

Given a viewing transformation with extrinsic matrix $\bm{T}$ and projection matrix $\bm{K}$, we get the projected position $\hat{\bm{\mu}}$ and covariance matrix $\hat{\bm{\Sigma}}$ in 2D screen space as:
\begin{equation}
\hat{\bm{\mu}} = \bm{KT}[\bm{\mu}, 1]^\top,\quad
\hat{\bm{\Sigma}} = \bm{J}\bm{T}\bm{\Sigma}\bm{T}^\top\bm{J}^\top,
\label{eq:projection}
\end{equation}
where $\bm{J}$ is the Jacobian matrix of the affine approximation of the perspective projection.
Note that 3DGS only retains the second-order values of $\hat{\bm{\mu}}$ and $\hat{\bm{\Sigma}}$ as $\hat{\bm{\mu}} \in \mathbb{R}^2$ and $\hat{\bm{\Sigma}} \in \mathbb{R}^{2\times 2}$ respectively.
The projected 2D Gaussian signal for the pixel $\bm{u}$ is given:
\begin{equation}
\vspace{-0.1cm}
g^\text{2D}(\bm{u} | \hat{\bm{\mu}}, \hat{\bm{\Sigma}}) = \exp\left(
-\frac{1}{2}(\bm{u}-\hat{\bm{\mu}})^\top\hat{\bm{\Sigma}}^{-1}(\bm{u}-\hat{\bm{\mu}})
\right),
\label{eq:gaussian_2d}
\vspace{-0.1cm}
\end{equation}
using the projected 2D Gaussian signal, 3DGS derives the volume transmittance and shades the color of pixel $\bm{u}$ through:
\begin{equation}
\vspace{-0.1cm}
\bm{C}(\bm{u}) = \sum_{i\in N} T_i g^\text{2D}_i(\bm{u} | \hat{\bm{\mu}_i}, \hat{\bm{\Sigma}_i})\alpha_i \bm{c}_i,\quad
T_i = \prod^{i-1}_{j=1}(1 - g^\text{2D}_{j}(\bm{u} | \hat{\bm{\mu}_j}, \hat{\bm{\Sigma}_j}) \alpha_j),
\label{eq:alpha_blending}
\vspace{-0.1cm}
\end{equation}
where the symbols with subscripts $i$ indicate the attributes related to the point $\bm{p}_i$. Specifically, $\alpha_i$ and $\bm{c}_i$ respectively denote the opacity and view-dependent color of point $\bm{p}_i$.

For better understanding, we further formulate the 2D Gaussian signal $g^\text{2D}$ in \cref{eq:gaussian_2d} as a flattened expression. Considering $\hat{\bm{\Sigma}}$ is a real-symmetric $2\times 2$ matrix, we numerically express $\hat{\bm{\Sigma}}$ and $\hat{\bm{\Sigma}}^{-1}$ as:
\begin{equation}
\vspace{-0.1cm}
\begin{aligned}
    \hat{\bm{\Sigma}} =
    \begin{bmatrix}
        \hat{\Sigma}_{11}\  & \ \hat{\Sigma}_{12} \\
        \hat{\Sigma}_{12}\  & \ \hat{\Sigma}_{22}
    \end{bmatrix} & ,\quad
    \hat{\bm{\Sigma}}^{-1} = \frac{1}{|\hat{\bm{\Sigma}}|}
    \begin{bmatrix}
        \hat{\Sigma}_{22}\   & \ -\hat{\Sigma}_{12} \\
        -\hat{\Sigma}_{12}\  & \ \hat{\Sigma}_{11}
    \end{bmatrix} =
    \begin{bmatrix}
        a\  & \ b \\
        b\  & \ c
    \end{bmatrix},       \\
    a   & = \frac{\hat{\Sigma}_{22}}{|\hat{\bm{\Sigma}}|}, \quad
    b = -\frac{\hat{\Sigma}_{12}}{|\hat{\bm{\Sigma}}|}, \quad
    c = \frac{\hat{\Sigma}_{11}}{|\hat{\bm{\Sigma}}|},
\end{aligned}
\label{eq:sigma}
\vspace{-0.1cm}
\end{equation}
given the pixel $\bm{u} = [u_x, u_y]^\top$ and the projected position $\hat{\bm{\mu}} = [\hat{\mu}_x, \hat{\mu}_y]^\top$ in 2D screen space, \cref{eq:gaussian_2d} can be rewrited as:
\begin{equation}
\begin{aligned}
    g^\text{2D}(\bm{u} & | \hat{\bm{\mu}}, \hat{\bm{\Sigma}}) = \exp\left(
    -\frac{a}{2} \hat{u}_x^2
    -\frac{c}{2} \hat{u}_y^2
    -b \hat{u}_x \hat{u}_y
    \right),   \\
    \hat{u}_x          & = u_x - \hat{\mu}_x, \quad
    \hat{u}_y = u_y - \hat{\mu}_y
    .
\end{aligned}
\label{eq:gaussian_2d_rewrited}
\end{equation}

\noindent\textbf{Remark.} It is worth noting that 3DGS treats each pixel as an isolated, single point when calculating its corresponding Gaussian value, as shown in Eq. \eqref{eq:gaussian_2d_rewrited}. This approximating scheme functions effectively when training and testing images to capture the scene content from a relatively consistent distance. However, when the pixel footprint changes due to focal length or camera distance adjustments, 3DGS renderings exhibit considerable artifacts, such as thin Gaussians observed during zooming in. As a result, it becomes crucial to define the pixel as a window area and calculate the corresponding intensity response by integrating the Gaussian signal within this domain. Rather than using the intuitive but time-consuming super sampling, it would be better to tackle the problem more analytically, given that the Gaussian Signal is a continuous function.

\section{Methods}
\label{sec:methods}
In \cref{sec:preliminary}, we observe that 3DGS ignores the window area of each pixel, and only considers the Gaussian value corresponding to the pixel center as its intensity response. This approach would inevitably produce artifacts due to fluctuations of the pixel footprint under different resolutions.
To address this problem, we are motivated to derive an analytical approximation of a 2D Gaussian signal within the pixel window area to accurately describe the intensity response of the imaging pixel. Subsequently, we plan to apply this derived integration to replace $g^\text{2D}$ in the 3DGS framework.

\begin{figure}[tbp]
    \centering 
    \begin{subfigure}{0.32\textwidth}
        \includegraphics[width=\linewidth]{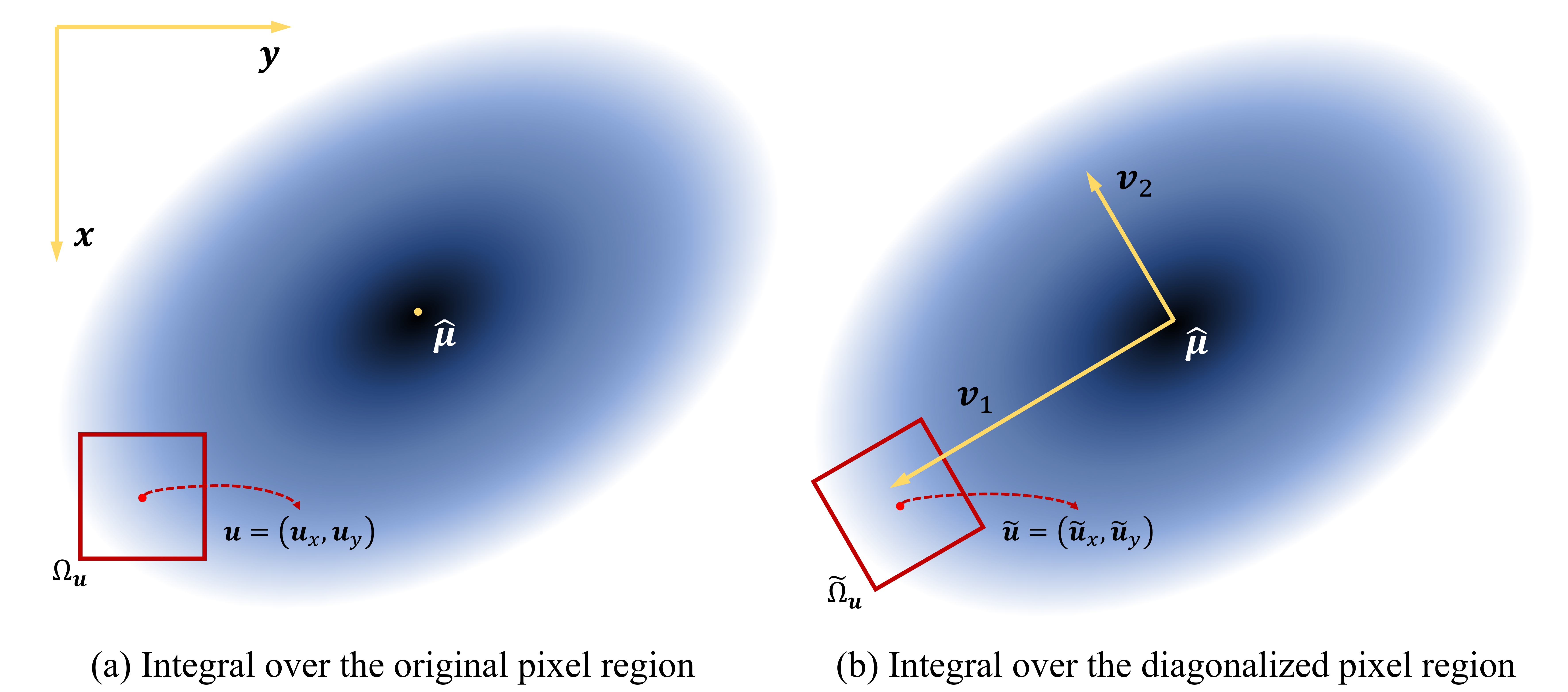}
        \caption{Integration (Ground-truth)}
        \label{fig:Integration}
    \end{subfigure}\hfil 
    \begin{subfigure}{0.32\textwidth}
        \includegraphics[width=\linewidth]{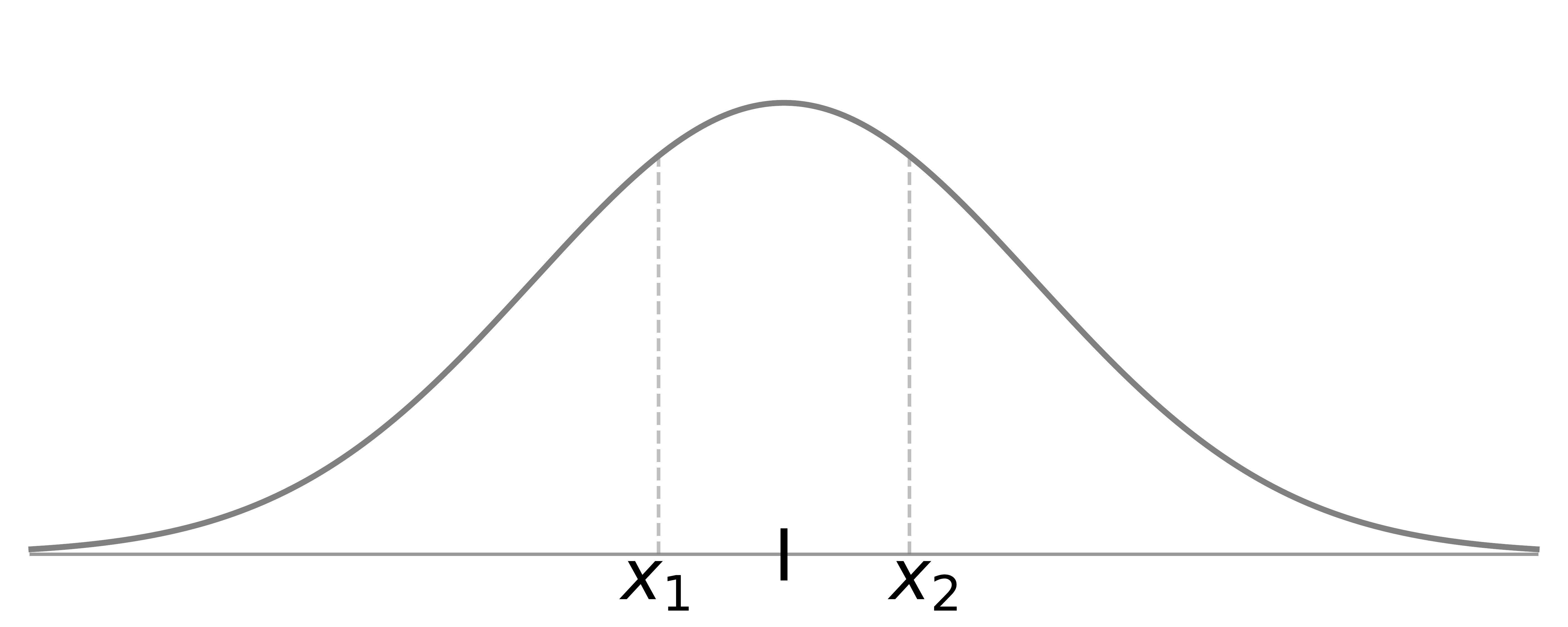}
        \caption{Baseline (3DGS)}
        \label{fig:Baseline}
    \end{subfigure}\hfil 
    \begin{subfigure}{0.32\textwidth}
        \includegraphics[width=\linewidth]{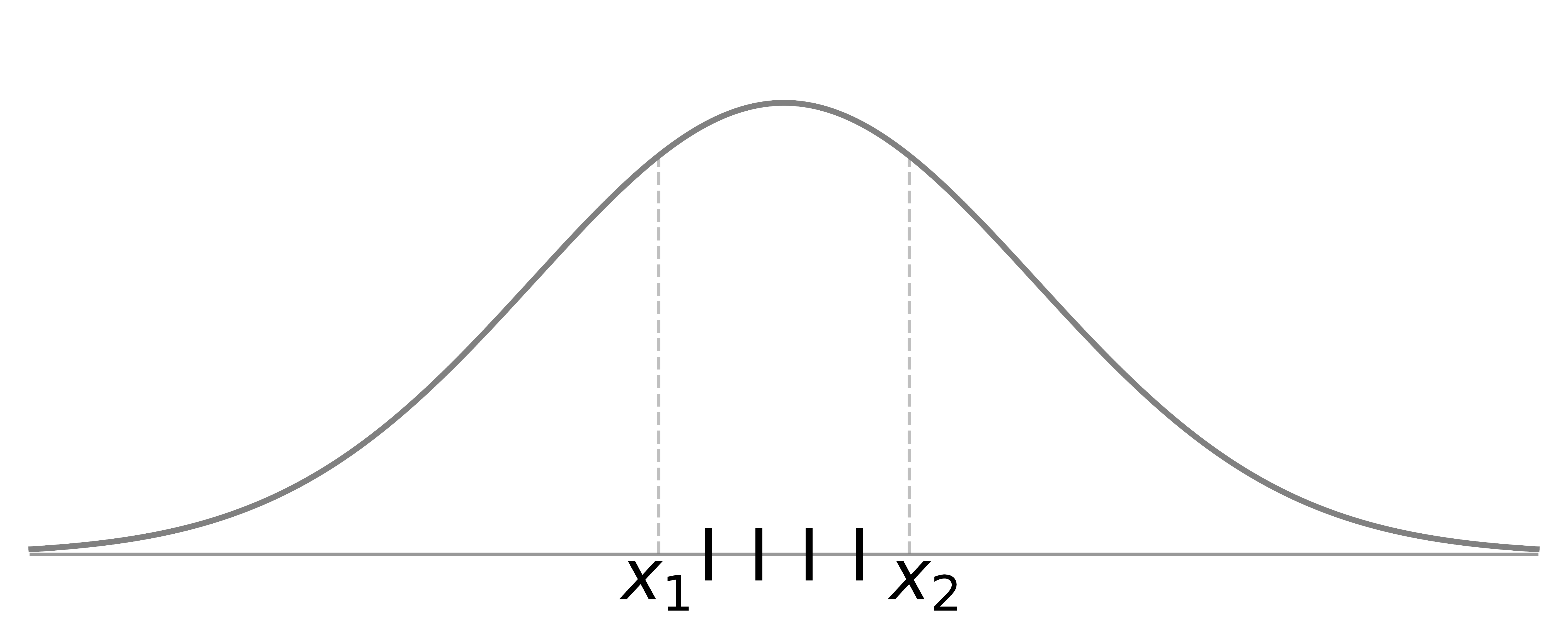}
        \caption{Super Sampling (3DGS-SS)}
        \label{fig:SuperSampling}
    \end{subfigure}

    \medskip
    \begin{subfigure}{0.32\textwidth}
        \includegraphics[width=\linewidth]{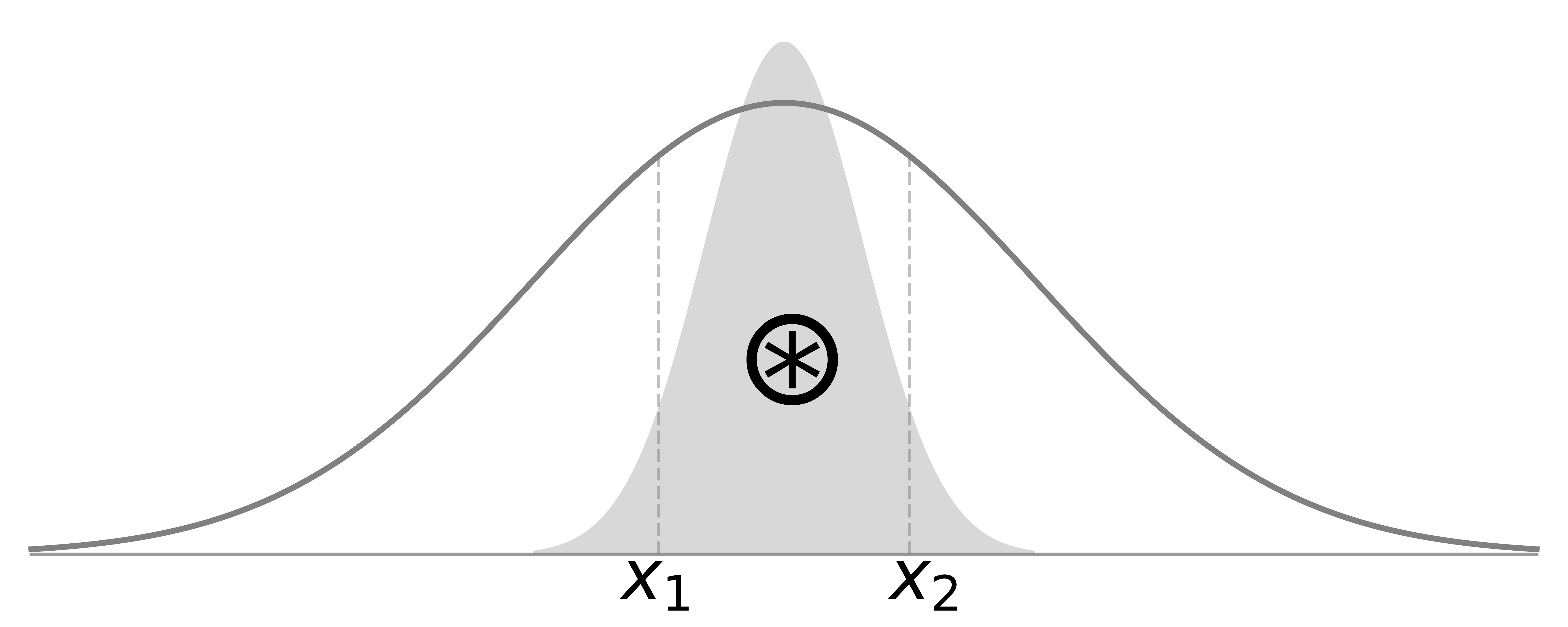}
        \caption{Filtering (Mip-Splatting)}
        \label{fig:Prefilter}
    \end{subfigure}\hfil 
    \begin{subfigure}{0.448\textwidth}
        \includegraphics[width=\linewidth]{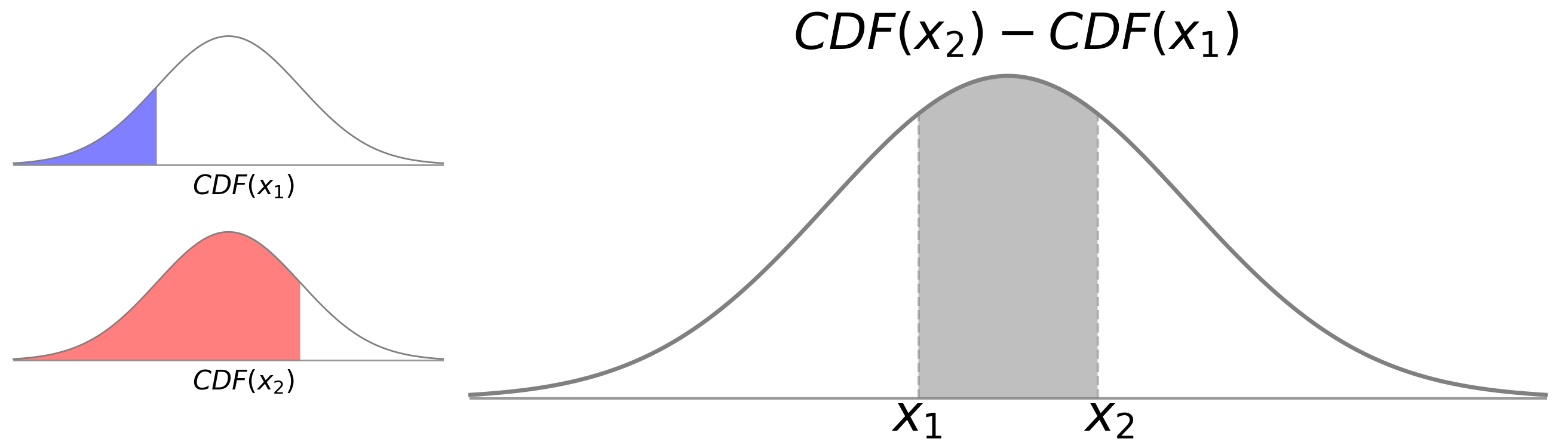}
        \caption{Analytic Approximation (Ours)}
        \label{fig:Analytic}
    \end{subfigure}
    \vspace{-0.2cm}
    \caption{Example diagram of the signal integration within window area and the approximation schemes used in different methods.}
    \label{fig:comparison}
    \vspace{-0.6cm}
\end{figure}

\subsection{Revisit One-dimensional Gaussian Signal Response}
\label{subsec:revisit}
Before describing our proposed method, we first revisit the example of the integrated response of a one-dimensional Gaussian Signal within a window area for better understanding.
Given a signal $g(x)$ and a window area $[x_1, x_2]$, we aim to get the response by integrating the signal $\mathcal{I}_g = \int^{x_2}_{x_1} g(x) \text{d}x$ within this domain as shown in \cref{fig:Integration}. For an unknown signal, Monte Carlo sampling within the window area is a feasible approach to approximate the integral as demonstrated in \cref{fig:Baseline} and \cref{fig:SuperSampling}, and the approximation result $\mathcal{I}_g \approx \frac{x_2 - x_1}{N} \sum^N_{i=1} g(x_i),\  x_i \in [x_1, x_2]$ will be more accurate as the number of samples $N$ increases. Nonetheless, increasing the number of samples (i.e., super sampling) increases the computation burden significantly.

Fortunately, our goal in the 3DGS framework is to obtain the intensity response of the Gaussian signal within the window area $[x_1, x_2]$. Given that the Gaussian signal is a continuous real-valued function, it is natural to derive an analytical approximation to the Gaussian definite integral (\cref{fig:Integration}) which is more accurate compared to the numerical integration (\cref{fig:Baseline} and \cref{fig:SuperSampling}).
For instance, in Mip-Splatting, the window area is treated as a Gaussian kernel $g_\text{w}$, and the integral is approximated as the result of sampling after convolving the Gaussian signal with the Gaussian kernel $\mathcal{I}_g \approx g \circledast g_\text{w}$ \footnote{the standard deviation $\sigma$ in $g_\text{w}$ is set to $0.1$. Note that the result of convolving a Gaussian signal with a Gaussian kernel is still a Gaussian signal} as shown in \cref{fig:Prefilter}. While this prefiltering uses the great convolution properties of Gaussian signals, this approximation will introduce a large gap when the Gaussian signal $g$ mainly consists of high-frequency components (\ie, with small standard variation $\sigma$).

To overcome these shortcomings, we are motivated to calculate the integral within the window area analytically. Specifically, the problem of evaluating the definite integral within $[x_1, x_2]$ can be simplified to the subtraction of two improper integrals by applying the first part of the fundamental theorem of calculus. Let $G(x)$ be the cumulative distribution function (CDF) of the standard Gaussian distribution $g(x)$ defined by,
\begin{equation}
\begin{aligned}
    G(x) = \int^x_{-\infty} g(x) \text{d}x,
    \quad
    g(x) & = \frac{1}{\sqrt{2\pi}} \exp\left(-
    \frac{x^2}{2}
    \right),
\end{aligned}
\label{eq:CDF}
\end{equation}
and the definite integral of $g(x)$ within $[x_1, x_2]$ can be expressed as,
\begin{equation}
\label{eq:int_by_CDF}
\mathcal{I}_g = G(x_2) - G(x_1).
\end{equation}

However, this CDF of the Gaussian function (defined as the error function $\text{erf}$) is not closed-form. We start by approximating the CDF $G(x)$ and find an important corollary, \textit{i.e.},
\begin{definition}
    The logistic function $S(x)$ is the analytic approximation of the CDF $G(x)$ with standard deviation $\sigma = 1$, which is defined as,
    \begin{equation}
        S(x) = \frac{1}{
            1 + \exp\left(
            -1.6 \cdot x - 0.07 \cdot x^3
            \right)
        },
        \label{eq:logistic_approximation}
    \end{equation}
    and could be derivative-friendly.
    \label{coro:one}
\end{definition}
This analytic approximation also contains similar properties of the CDF $G(x)$:
\begin{enumerate}
    \item $S(x)$ is \textit{non-decreasing} and \textit{right-continuous}, satisfying
          \begin{equation*}
\lim_{x\rightarrow-\infty} G(x) = \lim_{x\rightarrow-\infty} S(x) = 0 \quad \text{and} \quad \lim_{x\rightarrow+\infty} G(x) = \lim_{x\rightarrow+\infty} S(x) = 1.
          \end{equation*}
    \item The curve of $S(x)$ has \textit{2-fold rotational symmetry} around the point $(0, \nicefrac{1}{2})$,
          \begin{equation*}
              G(x) + G(-x) = S(x) + S(-x) = 1,\ \forall x\in\mathbb{R}.
          \end{equation*}
\end{enumerate}

For any Gaussian signals with different standard deviations, we can approximate their CDFs by scaling $x$ in \cref{eq:logistic_approximation} by the reciprocals of their standard deviations. Once $x$ in $S(x)$ scales by the reciprocal of $\sigma$, we express it as $S_\sigma(x)$. For more details, please refer to the \textit{supplementary material}.

In summary, given the sample $u$ and setting the window area as $1$, the integral $\mathcal{I}_g(u)$ of Gaussian signal $g(x)$ within the area $[u - \frac{1}{2}, u + \frac{1}{2}]$ is defined as:
\begin{equation}
\mathcal{I}_g(u) = \int^{u + \frac{1}{2}}_{u - \frac{1}{2}} g(x) \text{d}x = G(u + \frac{1}{2}) - G(u - \frac{1}{2}).
\label{eq:window_integral}
\end{equation}

Besides, according to \cref{coro:one}, we apply \cref{eq:logistic_approximation} and approximate the integral of \cref{eq:window_integral} as:
\begin{equation}
\mathcal{I}_g(u) \approx S(u + \frac{1}{2}) - S(u - \frac{1}{2}).
\label{eq:window_logistic}
\end{equation}

In the following section, we will discuss how to employ the above approximation to 2D-pixel shading in Analytic-Splatting.

\begin{figure}[tbp]
    \centering
    \begin{subfigure}[b]{0.45\textwidth}
        \centering
        \includegraphics[width=\textwidth]{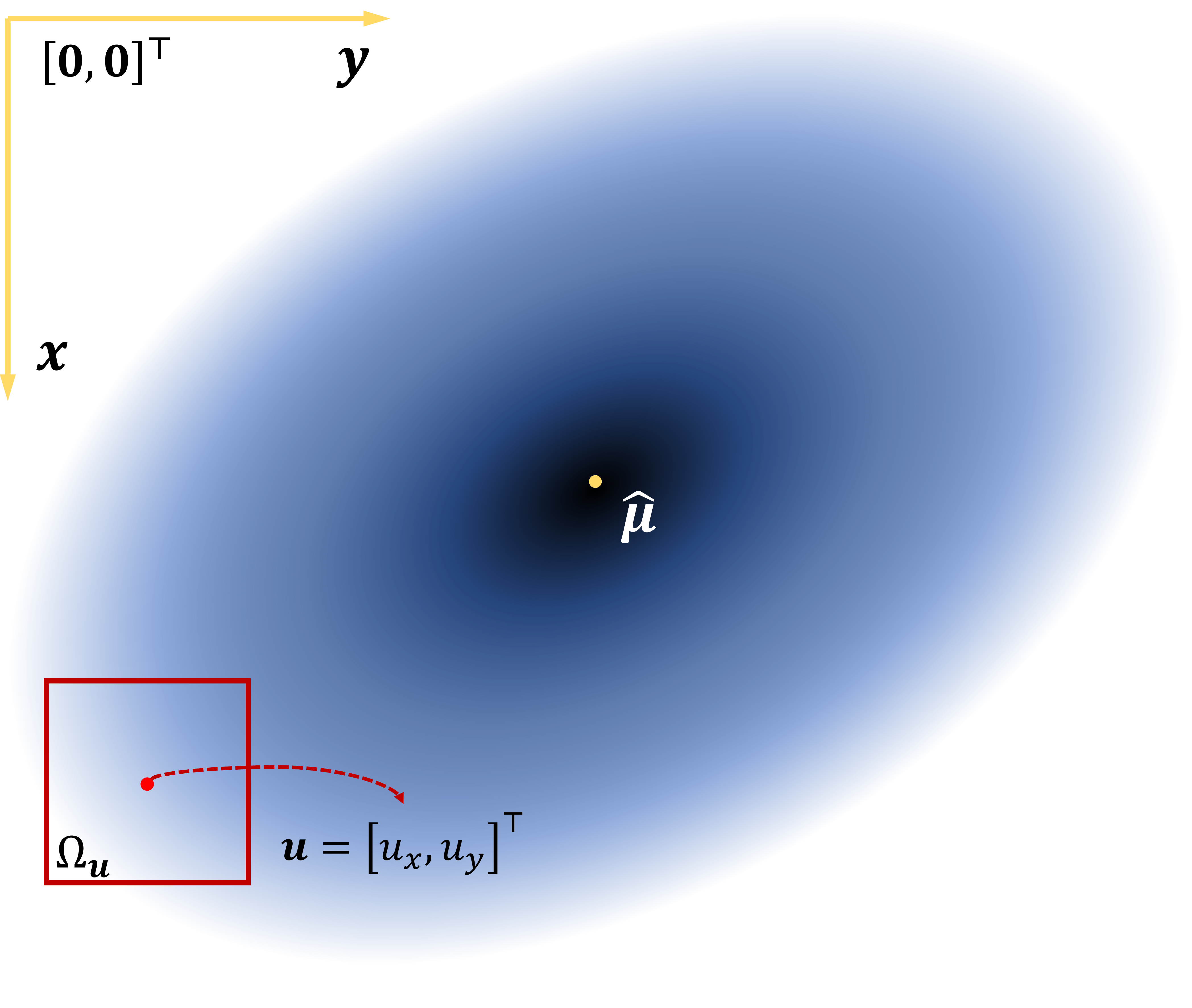}
        \caption{Integral over the original pixel area.}
        \label{fig:integral_pixel}
    \end{subfigure}
    \hfill
    \begin{subfigure}[b]{0.45\textwidth}
        \centering
        \includegraphics[width=\textwidth]{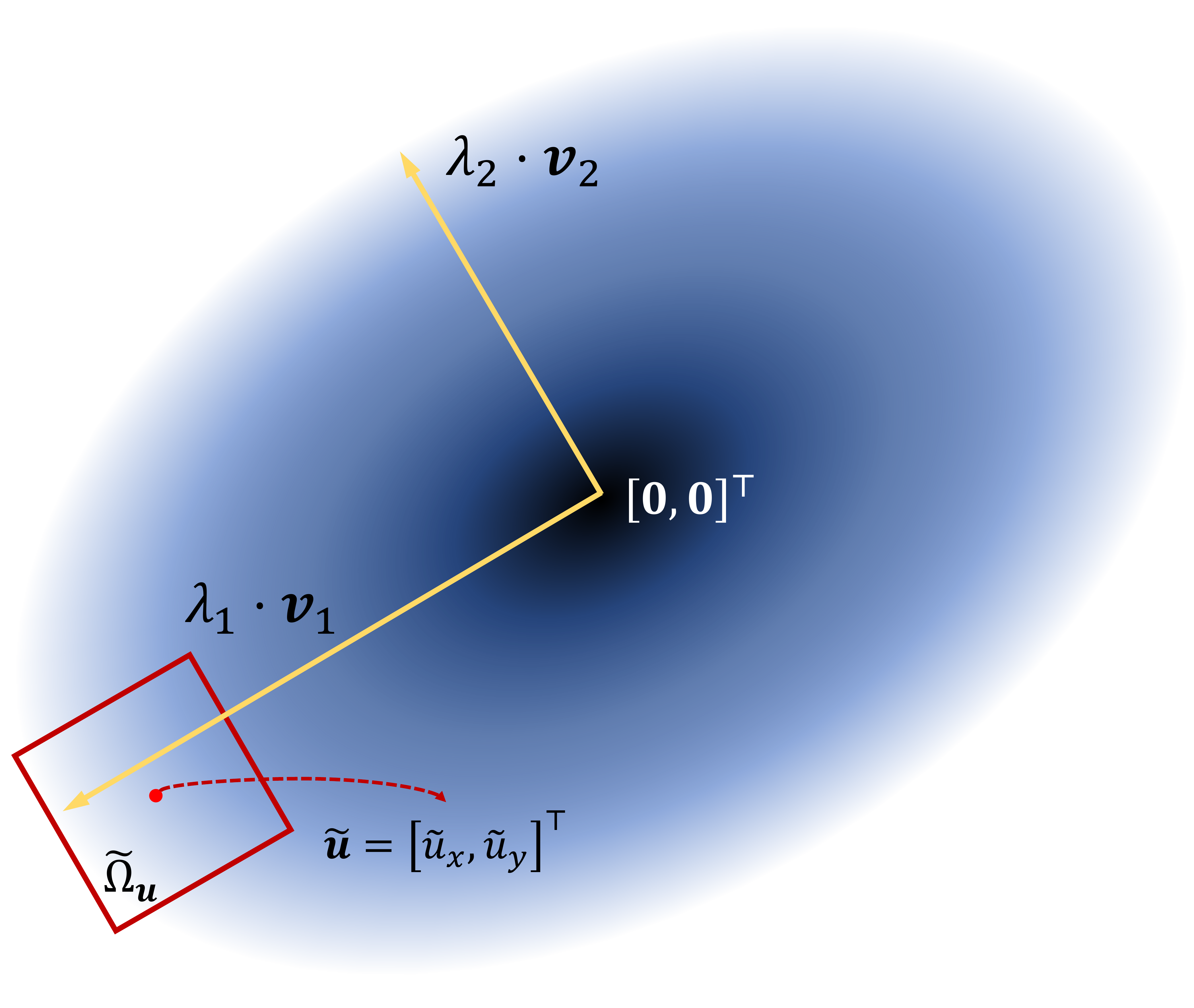}
        \caption{Integral over the rotated pixel area.}
        \label{fig:integral_rotated_pixel}
    \end{subfigure}
    \vspace{-0.2cm}
    \caption{Example diagram of the pixel integration domain in \cref{eq:integration_2D} and the domain after rotation in \cref{eq:integration_2D_approximation}. The yellow lines in \cref{fig:integral_pixel} are the coordinate axes of 2D screen space; And the yellow lines in \cref{fig:integral_rotated_pixel} are the eigenvectors scaled by the eigenvalues. }
    \label{fig:2D_approximation}
    \vspace{-0.6cm}
\end{figure}

\subsection{Analytic-Splatting}
\label{subsec:analytic_splatting}
After revisiting the one-dimensional Gaussian signal integration, we tend to approximate the integral of projected 2D Gaussians within each pixel window area in Analytic-Splatting. Mathematically, we replace the sampled $g^\text{2D}(\bm{u})$ in \cref{eq:alpha_blending} with the approximated integral $\mathcal{I}_g^\text{2D}(\bm{u})$.
For the pixel $\bm{u} = [u_x, u_y]^\top$ in 2D screen space, which corresponds to the window area $\Omega_{\bm{u}}$ as shown in \cref{fig:integral_pixel}, the integration of Gaussian signal in \cref{eq:gaussian_2d} is represented as:
\begin{equation}
\scriptsize
\begin{aligned}
    \mathcal{I}_g^\text{2D}(\bm{u}) =
    \int^{u_x + \frac{1}{2}}_{u_x - \frac{1}{2}}
    \int^{u_y + \frac{1}{2}}_{u_y - \frac{1}{2}} \exp\left(
    -\frac{a}{2} (x - \hat{\mu}_x)^2
    -\frac{c}{2} (y - \hat{\mu}_y)^2
    -\underbrace{b (x - \hat{\mu}_x) (y - \hat{\mu}_y)}_\text{correlation term}
    \right)\ \text{d}x\text{d}y.
\end{aligned}
\label{eq:integration_2D}
\vspace{-0.2cm}
\end{equation}

Notably, handling the correlation term in this integral is intractable. To unravel the correlation term and thus feasibly solve the integral, we diagonalize the covariance matrix $\hat{\bm{\Sigma}}$ of the 2D Gaussian $g^\text{2D}$ and slightly rotate the integration domain as shown in \cref{fig:integral_rotated_pixel}, thus approximating the integral by multiplying two independent 1D Gaussian integrals.

In detail, we first perform eigenvalue decomposition on the covariance matrix $\hat{\bm{\Sigma}}$ (refer to \cref{eq:sigma}) to obtain eigenvalues $\{\lambda_1, \lambda_2\}$ and the corresponding eigenvectors $\{\bm{v}_1, \bm{v}_2\}$.
After diagonalization, for better description, we take $\hat{\bm{\mu}} = [\hat{\mu}_x, \hat{\mu}_y]^\top$ (the mean vector of $g^\text{2D}$) as the origin and the eigenvectors $\{\bm{v}_1, \bm{v}_2\}$ as the axis to construct a new coordinate system.

In this coordinate system, given a pixel $\bm{u} = [u_x, u_v]^\top$, we rewrite the $g^\text{2D}$ in \cref{eq:gaussian_2d_rewrited} as the multiplication of two independent 1D Gaussians:
\begin{equation}
\small
\begin{aligned}
g^\text{2D}(\bm{u}) & = \exp\left(
-\frac{1}{2\lambda_1}\tilde{u}_x^2
-\frac{1}{2\lambda_2}\tilde{u}_y^2
\right) = \exp\left(
-\frac{1}{2\lambda_1}\tilde{u}_x^2
\right) \exp\left(
-\frac{1}{2\lambda_2}\tilde{u}_y^2
\right),        \\
\tilde{\bm{u}}      & =
\begin{bmatrix}
    \tilde{u}_x \\ \tilde{u}_y
\end{bmatrix} =
\begin{bmatrix}
- & \bm{v}_1 & - \\
- & \bm{v}_2 & -
\end{bmatrix}
\left(\bm{u} - \hat{\bm{\mu}}\right)
=
\begin{bmatrix}
- & \bm{v}_1 & - \\
- & \bm{v}_2 & -
\end{bmatrix}
\begin{bmatrix}
u_x - \hat{\mu}_x \\
u_y - \hat{\mu}_y
\end{bmatrix},
\end{aligned}
\label{eq:gaussian_2d_transformed}
\vspace{-0.2cm}
\end{equation}
where $\tilde{\bm{u}} = [\tilde{u}_x, \tilde{u}_y]^\top$ denotes the diagonalized coordinate of the pixel center. After diagonalization, we further rotate the pixel integration domain $\Omega_{\bm{u}}$ along the pixel center to align it with the eigenvectors and get $\tilde{\Omega}_{\bm{u}}$ for approximating the integral. Thus the integral in \cref{eq:integration_2D} can be approximated as:

\begin{equation}
\scriptsize
\begin{aligned}
    \mathcal{I}_g^\text{2D}(\bm{u}) & \approx \int_{\tilde{\Omega}_{\bm{u}}}g^\text{2D}(\bm{u}) \text{d}\bm{u}
     =\int^{\tilde{u}_x + \frac{1}{2}}_{\tilde{u}_x - \frac{1}{2}} \exp\left(
    -\frac{1}{2\lambda_1} x^2
    \right) \text{d}x
    \int^{\tilde{u}_y + \frac{1}{2}}_{\tilde{u}_y - \frac{1}{2}} \exp\left(
    -\frac{1}{2\lambda_2} y^2
    \right) \text{d}y \\
    & = 2\pi\sqrt{\lambda_1 \lambda_2} \int^{\tilde{u}_x + \frac{1}{2}}_{\tilde{u}_x - \frac{1}{2}} \frac{1}{\sqrt{2\pi\lambda_1}}\exp\left(
    -\frac{1}{2\lambda_1} x^2
    \right) \text{d}x
    \int^{\tilde{u}_y + \frac{1}{2}}_{\tilde{u}_y - \frac{1}{2}} \frac{1}{\sqrt{2\pi\lambda_2}}\exp\left(
    -\frac{1}{2\lambda_2} y^2
    \right) \text{d}y \\
    & \approx 2\pi\sigma_1\sigma_2 \left[
        S_{\sigma_1}(\tilde{u}_x + \frac{1}{2}) - S_{\sigma_1}(\tilde{u}_x - \frac{1}{2})
        \right]\left[
        S_{\sigma_2}(\tilde{u}_y + \frac{1}{2}) - S_{\sigma_2}(\tilde{u}_y - \frac{1}{2})
    \right]. \\
\end{aligned}
\label{eq:integration_2D_approximation}
\end{equation}
where $\sigma_{\ast}$ subscripts of $G$ and $S$ respectively correspond to Gaussian signals with the standard deviation $\sigma_{\ast}$. $\sigma_1 = \sqrt{\lambda_1}$ and $\sigma_2 = \sqrt{\lambda_2}$ denote the standard derivations of the independent Gaussian signals along two eigenvectors respectively. In summary, the volume shading in Analytic-Splatting is given by:
\begin{equation}
\small
\begin{aligned}
    \bm{C}(\bm{u}) & = \sum_{i\in N} T_i \mathcal{I}_{g-i}^\text{2D}(\bm{u} | \hat{\bm{\mu}_i}, \hat{\bm{\Sigma}_i})\alpha_i \bm{c}_i,\quad
    T_i = \prod^{i-1}_{j=1}(1 - \mathcal{I}_{g-j}^\text{2D}(\bm{u} | \hat{\bm{\mu}_j}, \hat{\bm{\Sigma}_j}) \alpha_j),   \\
    \mathcal{I}_{g}^\text{2D}(\bm{u}) & = 2\pi\sigma_1\sigma_2\left[
        S_{\sigma_1}(\tilde{u}_x + \frac{1}{2}) - S_{\sigma_1}(\tilde{u}_x - \frac{1}{2})
        \right]\left[
        S_{\sigma_2}(\tilde{u}_y + \frac{1}{2}) - S_{\sigma_2}(\tilde{u}_y - \frac{1}{2})
    \right].
\end{aligned}
\label{eq:summary}
\end{equation}

\section{Experiments}
\label{sec:experiments}
\subsection{Approximation Error Analysis}
\label{subsec:approximation}
In this section, we first comprehensively explore the approximation errors in our scheme and then conduct an elaborate comparison against other schemes. It is worth noting that during training, the pruning and densification schemes proposed in 3DGS tend to maintain the standard deviations of Gaussian signals within an appropriate range (\ie $\sigma \in [0.3,\ 6.6]$), and each Gaussian signal only responds to pixels within the $99\%$ confidence interval (\ie $|x| < 3\sigma$) for shading. Therefore, we only consider the Gaussian signals with standard deviations in $[0.3,\ 6.6]$, and merely discuss the approximation error of pixels within the $99\%$ confidence interval.

\begin{figure}[!t]
\centering
\begin{subfigure}[b]{0.45\textwidth}
    \centering
    \includegraphics[width=\textwidth]{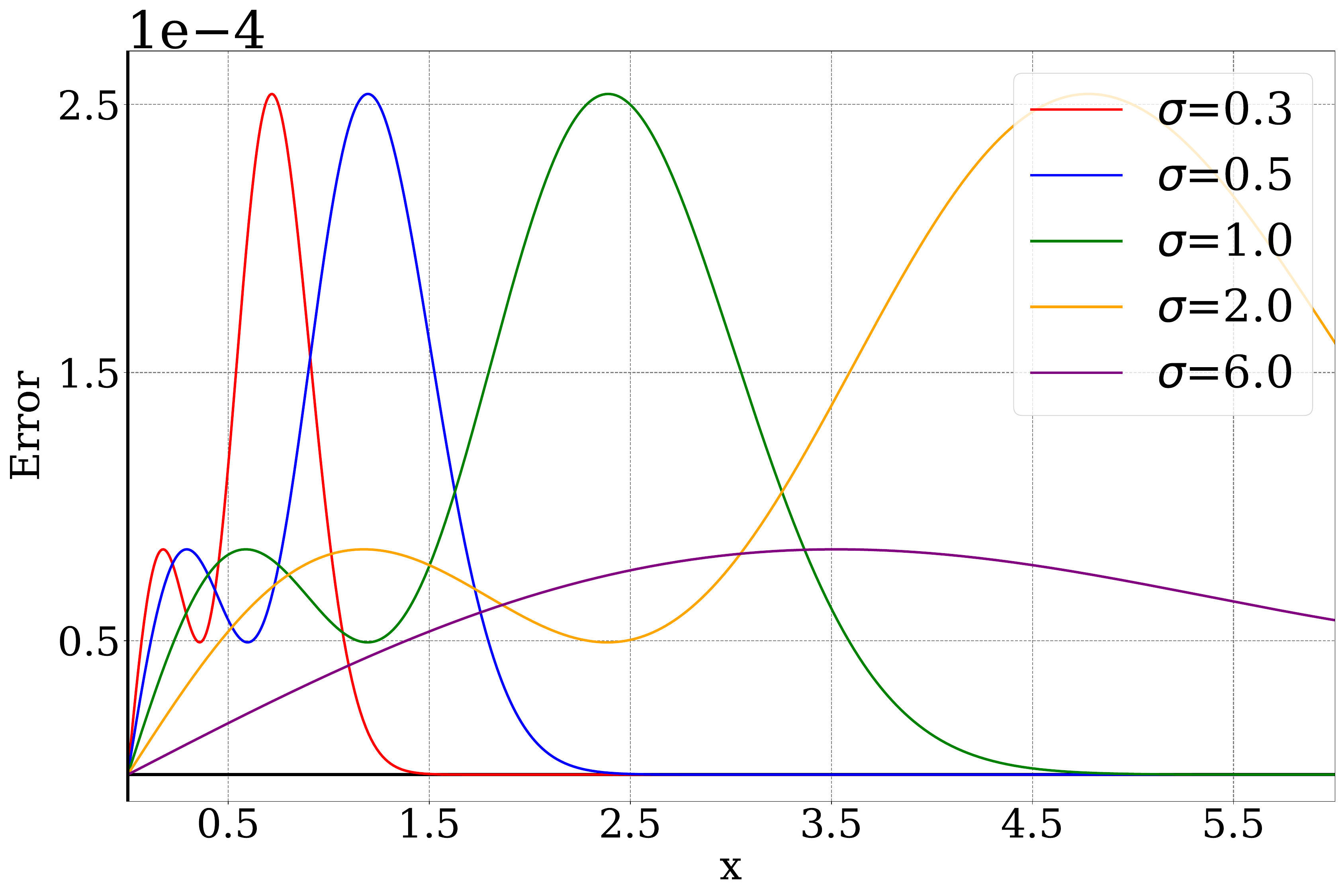}
    \caption{Approximation errors of our conditioned logistic function \cref{eq:logistic_approximation} to the CDF \cref{eq:CDF} referring to different standard deviations.\\ }
    \label{fig:CDF_error}
\end{subfigure}
\hfill
\begin{subfigure}[b]{0.45\textwidth}
    \centering
    \includegraphics[width=\textwidth]{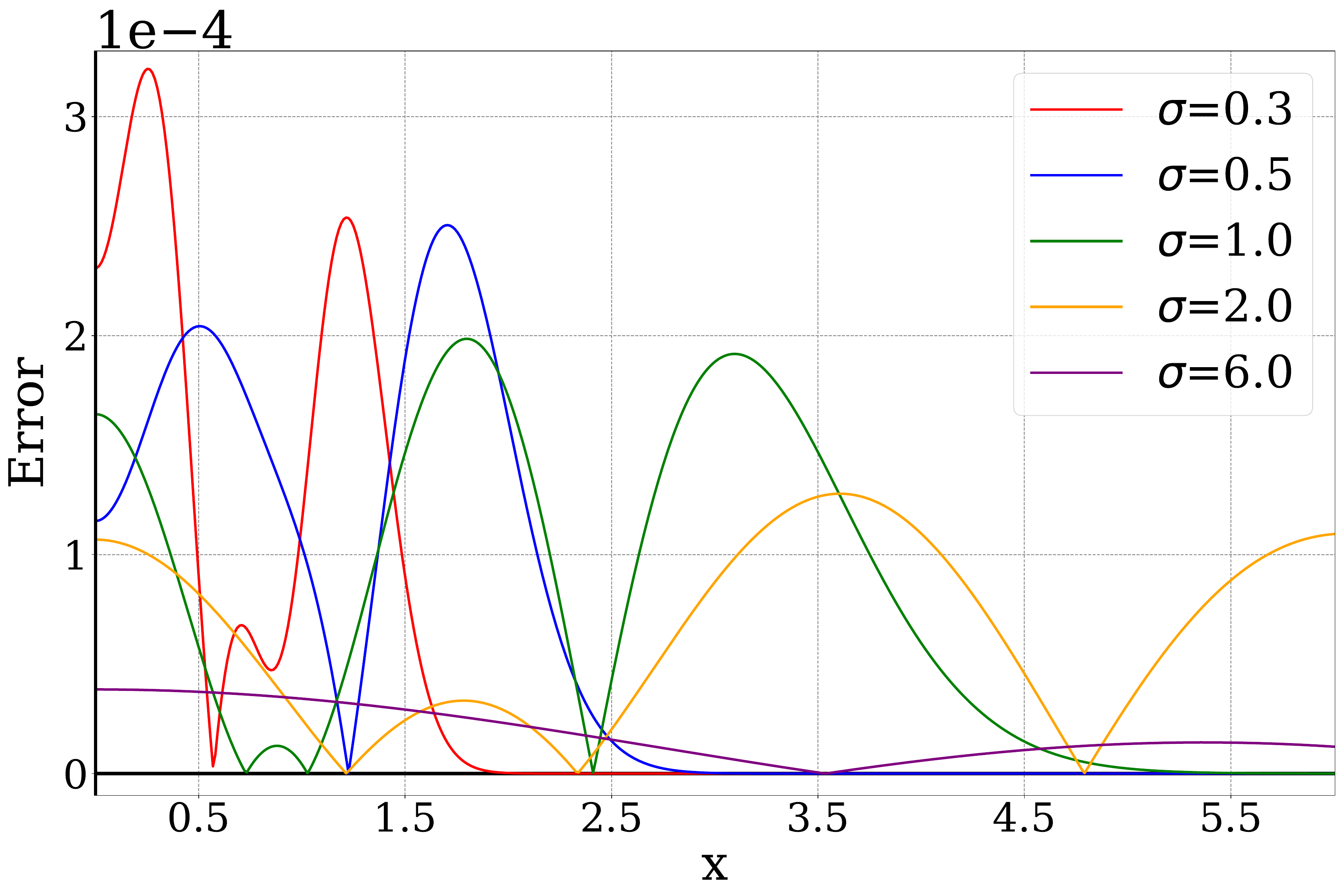}
    \caption{Approximation errors of our approximated window integral \cref{eq:window_logistic} to the ground-truth \cref{eq:window_integral} referring to different standard deviations.}
    \label{fig:Window_error}
\end{subfigure}
\vspace{-0.2cm}
\caption{Error Analysis of using our conditioned logistic function \cref{eq:logistic_approximation} to approximate CDF of Gaussian signals \cref{eq:CDF} and window integration \cref{eq:window_integral}. Note that the scaling factor of the error is $1e^{-4}$.}
\label{fig:approximation_error}
\vspace{-0.4cm}
\end{figure}

Referring to \cref{eq:CDF} and \cref{eq:logistic_approximation}, we get the approximation error function about the CDF of the Gaussian function:
\begin{equation}
\small
\mathcal{E}_{\text{CDF}}(x) = \left\lvert S(x) - G(x) \right\rvert,
\label{eq:error_CDF}
\end{equation}
and referring to \cref{eq:window_integral} and \cref{eq:window_logistic}, the approximation error function regarding the 1-width window area integral response is:
\begin{equation}
\small
\mathcal{E}_{\text{Int}}(x) = \left\lvert \left(
    S(x + \frac{1}{2}) - S(x + \frac{1}{2})
    \right) - \left(
    G(x + \frac{1}{2}) - G(x + \frac{1}{2})
\right)
\right\rvert,
\label{eq:error_window}
\end{equation}
the approximation error $\mathcal{E}_\text{CDF}$ and $\mathcal{E}_\text{Int}$ referring to different standard deviations are shown in \cref{fig:CDF_error} and \cref{fig:Window_error} respectively. \footnote{Since \cref{eq:error_CDF} and \cref{eq:error_window} are even functions, we show the results for the positive semi-axis over $0\leq x \leq 6$ in \cref{fig:approximation_error}.} Please note that the errors shown in \cref{fig:approximation_error} are scaled by a tiny factor of $1\mathrm{e}{-4}$.

\begin{figure}[tbp]
    \centering
    \begin{subfigure}[b]{0.32\textwidth}
        \centering
        \includegraphics[width=\textwidth]{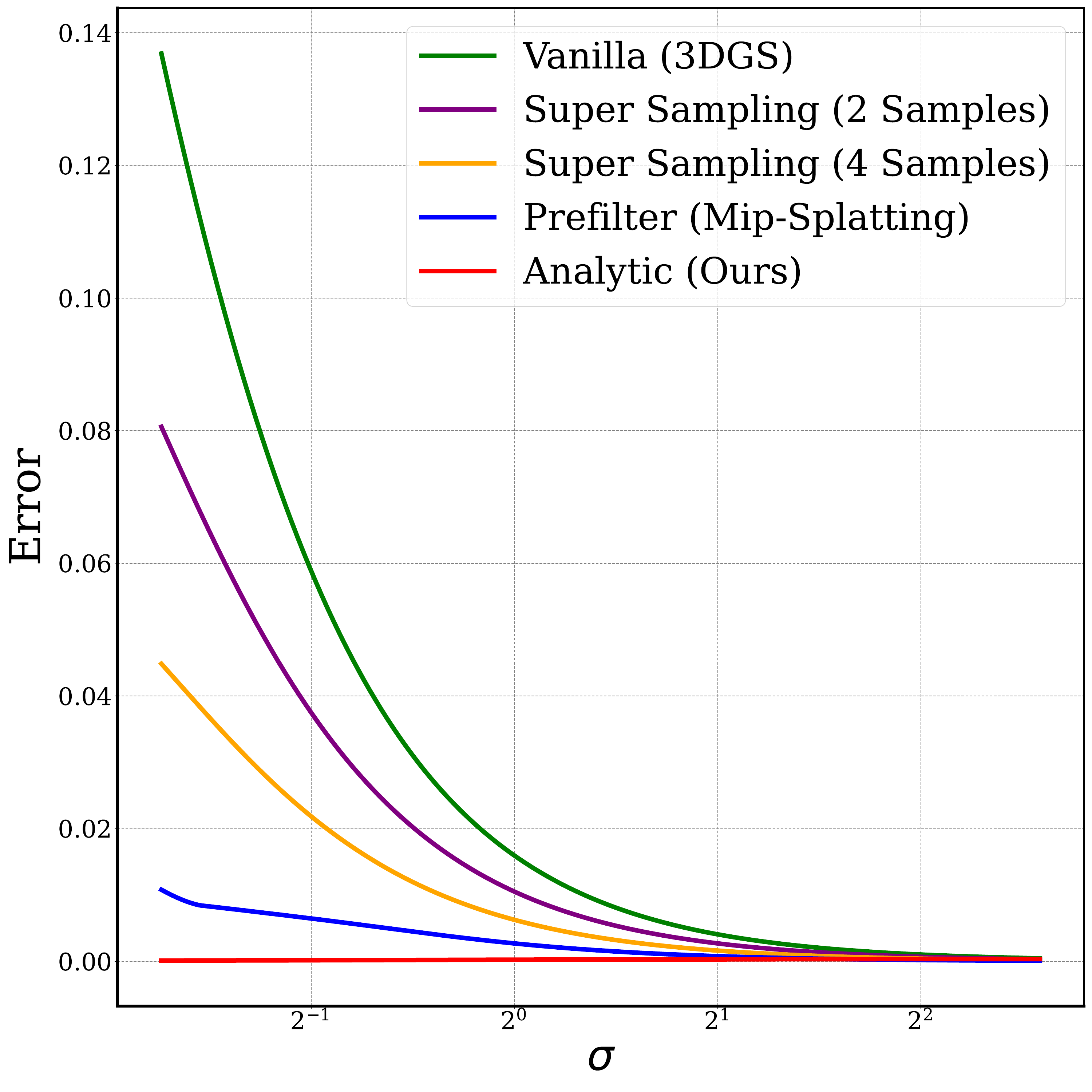}
        \caption{Approximation errors referring to different standard variations.}
        \label{fig:error_size}
    \end{subfigure}
    \hfill
    \begin{subfigure}[b]{0.32\textwidth}
        \centering
        \includegraphics[width=\textwidth]{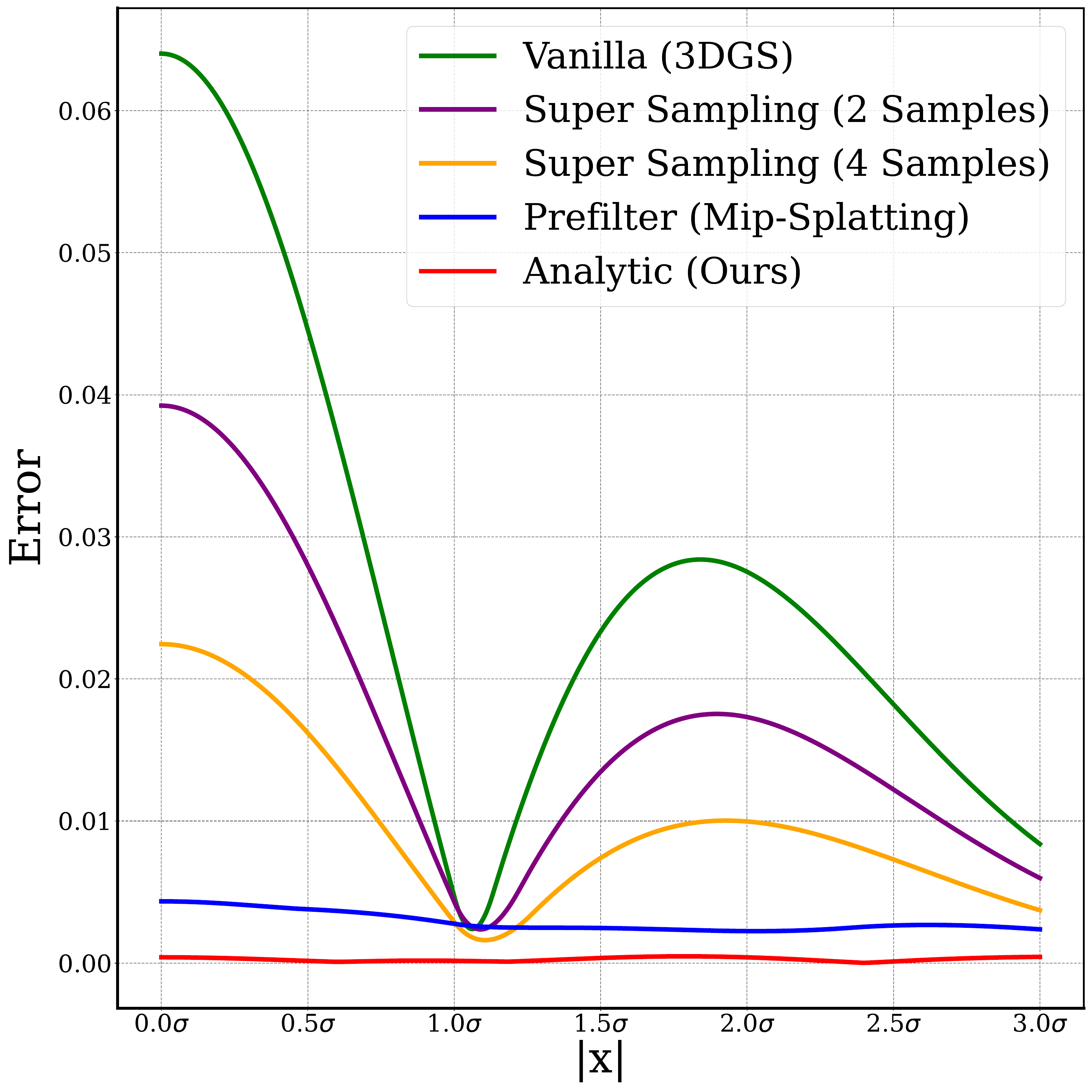}
        \caption{Approximation errors for different variable distribution.\\}
        \label{fig:error_distance}
    \end{subfigure}
    \hfill
    \begin{subfigure}[b]{0.32\textwidth}
        \centering
        \includegraphics[width=\textwidth]{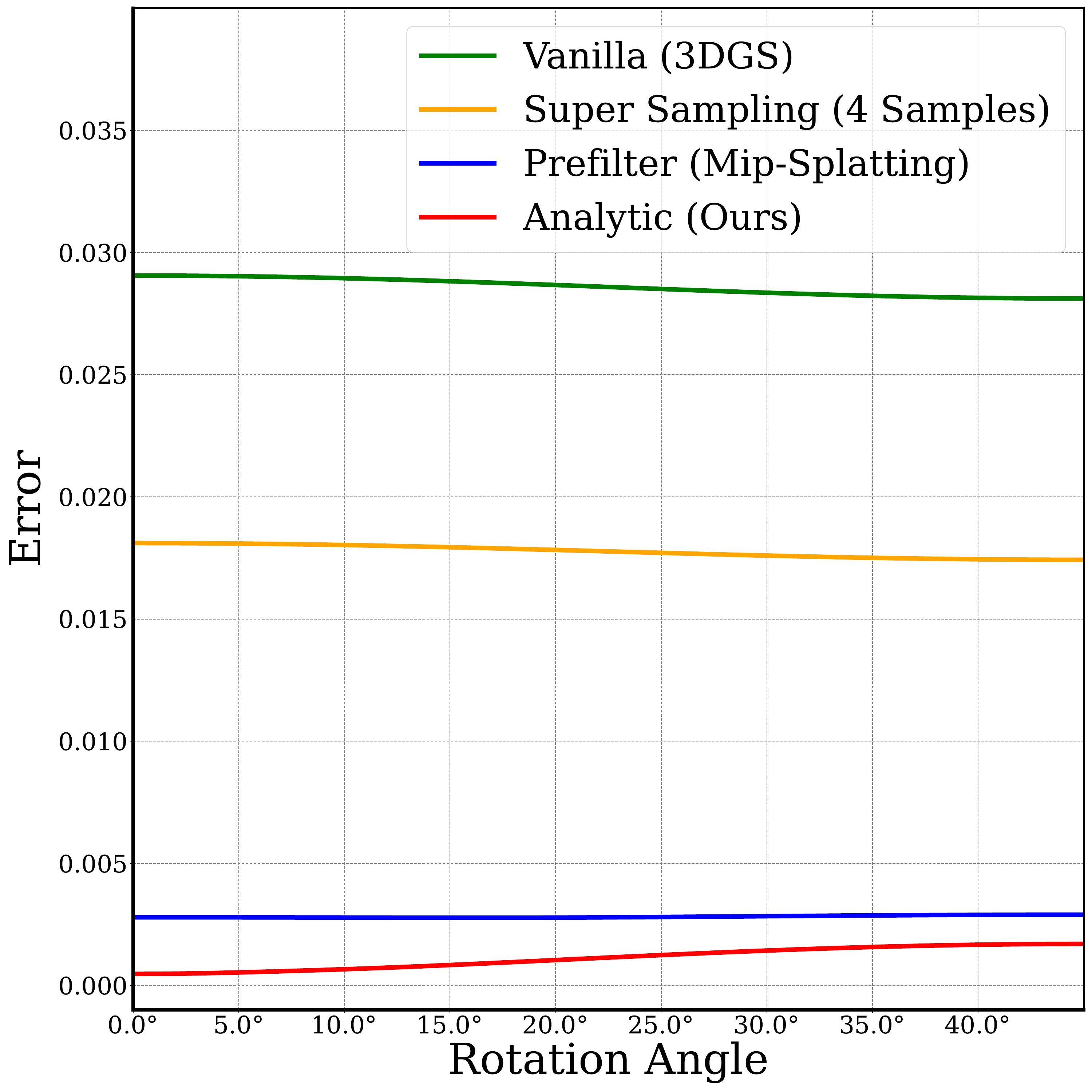}
        \caption{Approximation errors of rotating the integral domain by different angles in \cref{fig:2D_approximation}.}
        \label{fig:error_rotation}
    \end{subfigure}
    \vspace{-0.2cm}
    \caption{Error Analysis of approximating the window integral \cref{eq:window_integral} using different schemes in \cref{fig:comparison}.}
    \label{fig:error_comparison}
    \vspace{-0.6cm}
\end{figure}

For approximating the integral response over the 1-width window area, our scheme significantly outperforms other schemes. \cref{fig:error_size} and \cref{fig:error_distance} respectively show that one-dimensional approximation error with different standard deviations and variable distributions. Our scheme is robust to these two conditions, especially when the standard deviations of Gaussian signals become smaller, our advantage becomes more obvious, which means that our scheme is better at capturing the high-frequency signals (\ie details) of the scene, and our subsequent experimental results also verify this.

\begin{table}[!htbp]
\centering
\scalebox{0.6}{
\begin{tabular}{l|c c c c c|c c c c c|c c c c c}
 & \multicolumn{5}{c|}{PSNR $\uparrow$} & \multicolumn{5}{c|}{SSIM $\uparrow$} & \multicolumn{5}{c}{LPIPS $\downarrow$} \\
& Full Res. & $\nicefrac{1}{2}$ Res. & $\nicefrac{1}{4}$ Res. & $\nicefrac{1}{8}$ Res. & Avg.
& Full Res. & $\nicefrac{1}{2}$ Res. & $\nicefrac{1}{4}$ Res. & $\nicefrac{1}{8}$ Res. & Avg.
& Full Res. & $\nicefrac{1}{2}$ Res. & $\nicefrac{1}{4}$ Res. & $\nicefrac{1}{8}$ Res. & Avg. \\
\hline
NeRF w/o $\mathcal{L}_\text{area}$ &
31.20 & 30.65 & 26.25 & 22.53 & 27.66 &
0.950 & 0.956 & 0.930 & 0.871 & 0.927 &
0.055 & 0.034 & 0.043 & 0.075 & 0.052 \\
NeRF~\cite{mildenhall2020nerf} &
29.90 & 32.13 & 33.40 & 29.47 & 31.23 &
0.938 & 0.959 & 0.973 & 0.962 & 0.958 &
0.074 & 0.040 & 0.024 & 0.039 & 0.044 \\
MipNeRF~\cite{barron2021mip} &
32.63 & \cc{3}34.34 & \cc{2}35.47 & \cc{2}35.60 & \cc{3}34.51 &
0.958 & 0.970 & 0.979 & 0.983 & 0.973 &
0.047 & 0.026 & 0.017 & \cc{3}0.012 & 0.026 \\
\hline
Plenoxels~\cite{fridovich2022plenoxels} &
31.60 & 32.85 & 30.26 & 26.63 & 30.34 &
0.956 & 0.967 & 0.961 & 0.936 & 0.955 &
0.052 & 0.032 & 0.045 & 0.077 & 0.051 \\
TensoRF~\cite{chen2022tensorf} &
32.11 & 33.03 & 30.45 & 26.80 & 30.60 &
0.956 & 0.966 & 0.962 & 0.939 & 0.956 &
0.056 & 0.038 & 0.047 & 0.076 & 0.054 \\
Instant-NGP~\cite{muller2022instant} &
30.00 & 32.15 & 33.31 & 29.35 & 31.20 &
0.939 & 0.961 & 0.974 & 0.963 & 0.959 &
0.079 & 0.043 & 0.026 & 0.040 & 0.047 \\
Tri-MipRF~\cite{hu2023tri} &
\cc{3}32.65 & 34.24 & 35.02 & \cc{3}35.53 & 34.36 &
0.958 & 0.971 & \cc{3}0.980 & \cc{3}0.987 & \cc{3}0.974 &
0.047 & 0.027 & 0.018 & \cc{3}0.012 & 0.026 \\
3DGS~\cite{kerbl20233d} &
28.79 & 30.66 & 31.64 & 27.98 & 29.77 &
0.943 & 0.962 & 0.972 & 0.960 & 0.960 &
0.065 & 0.038 & 0.025 & 0.031 & 0.040 \\
3DGS-SS~\cite{kerbl20233d} &
32.05 & 33.78 & 33.92 & 31.12 & 32.71 &
\cc{3}0.964 & \cc{3}0.975 & \cc{3}0.980 & 0.977 & \cc{3}0.974 &
\cc{3}0.039 & \cc{3}0.021 & \cc{3}0.016 & 0.020 & \cc{3}0.024 \\
Mip-Splatting~\cite{yu2023mip} &
\cc{2}32.81 & \cc{2}34.49 & \cc{3}35.45 & 35.50 & \cc{2}34.56 &
\cc{1}0.967 & \cc{1}0.977 & \cc{2}0.983 & \cc{2}0.988 & \cc{1}0.979 &
\cc{2}0.035 & \cc{1}0.019 & \cc{2}0.013 & \cc{1}0.010 & \cc{2}0.019 \\
\hline
Ours &
\cc{1}33.22 & \cc{1}34.92 & \cc{1}35.98 & \cc{1}36.00 & \cc{1}35.03 & 
\cc{1}0.967 & \cc{1}0.977 & \cc{1}0.984 & \cc{1}0.989 & \cc{1}0.979 & 
\cc{1}0.033 & \cc{1}0.019 & \cc{1}0.012 & \cc{1}0.010 & \cc{1}0.018
\end{tabular}
}
\caption{\textbf{Quantatitive Comparison of Analytic-Splatting against several cutting-edge methods on the multi-scale Blender Synthetic dataset~\cite{barron2021mip}.} These methods conduct multi-scale training and testing.}
\label{tab:multi_blender}
\vspace{-1.cm}
\end{table}

Last but not least, we employ this scheme to approximate the window integral responses of two-dimensional Gaussian signals, which requires rotating the integration domain from $\Omega_{\bm{u}}$ to $\tilde{\Omega}_{\bm{u}}$ as shown in \cref{fig:2D_approximation} and inevitably introduces additional errors.
To study this error, we record the approximation errors caused by rotating the integration domain from $0^\circ$ to $45^\circ$ \footnote{Since we always hold $\sigma_1 \geq \sigma_2$ in practice, thus we only consider the approximation error caused by the rotation angle from $0^\circ$ to $45^\circ$.} under different standard deviations of Gaussian signals and distributions, as shown in \cref{fig:error_rotation}. Although the approximation error of our scheme slightly increases as the rotation angle becomes larger, our scheme still surpasses other schemes.

\subsection{Comparison}
\label{subsec:comparison}
To verify the anti-aliasing capability and versatility of Analytic-Splatting, we conduct experiments against state-of-the-art methods under the multi-scale training \& multi-scale testing (MTMT) setting on Blender Synthetic~\cite{mildenhall2020nerf, barron2021mip} and Mip-NeRF 360~\cite{barron2022mip} datasets. We further evaluate the performance of 3DGS and its variants under the super-resolution setting.

\noindent\textbf{Dataset \& Metric.} We conduct experiments using benchmark datasets of multi-scale Blender Synthetic~\cite{mildenhall2020nerf, barron2021mip} and multi-scale Mip-NeRF 360~\cite{barron2022mip}. They respectively contain 8 objects and 9 scenes, each object and scene is compiled by downscaling the original dataset with a factor of 2, 4, and 8, and combining. For the Blender Synthetic dataset, each object contains 100 images for training and 200 images for testing. For the Mip-NeRF 360 dataset, we select 1 image from every 8 images for testing and the remaining images for training.
To verify the efficacy of our method, we evaluate the synthesized novel view at multiple scales on both datasets in terms of Peak Signal-to-Noise Ratio (PSNR), Structural Similarity Index Measure (SSIM), and Learned Perceptual Image Path Similarity (LPIPS)~\cite{zhang2018unreasonable}. 

\begin{figure}[!t]
\centering
\scalebox{0.95}{
\includegraphics[width=\textwidth]{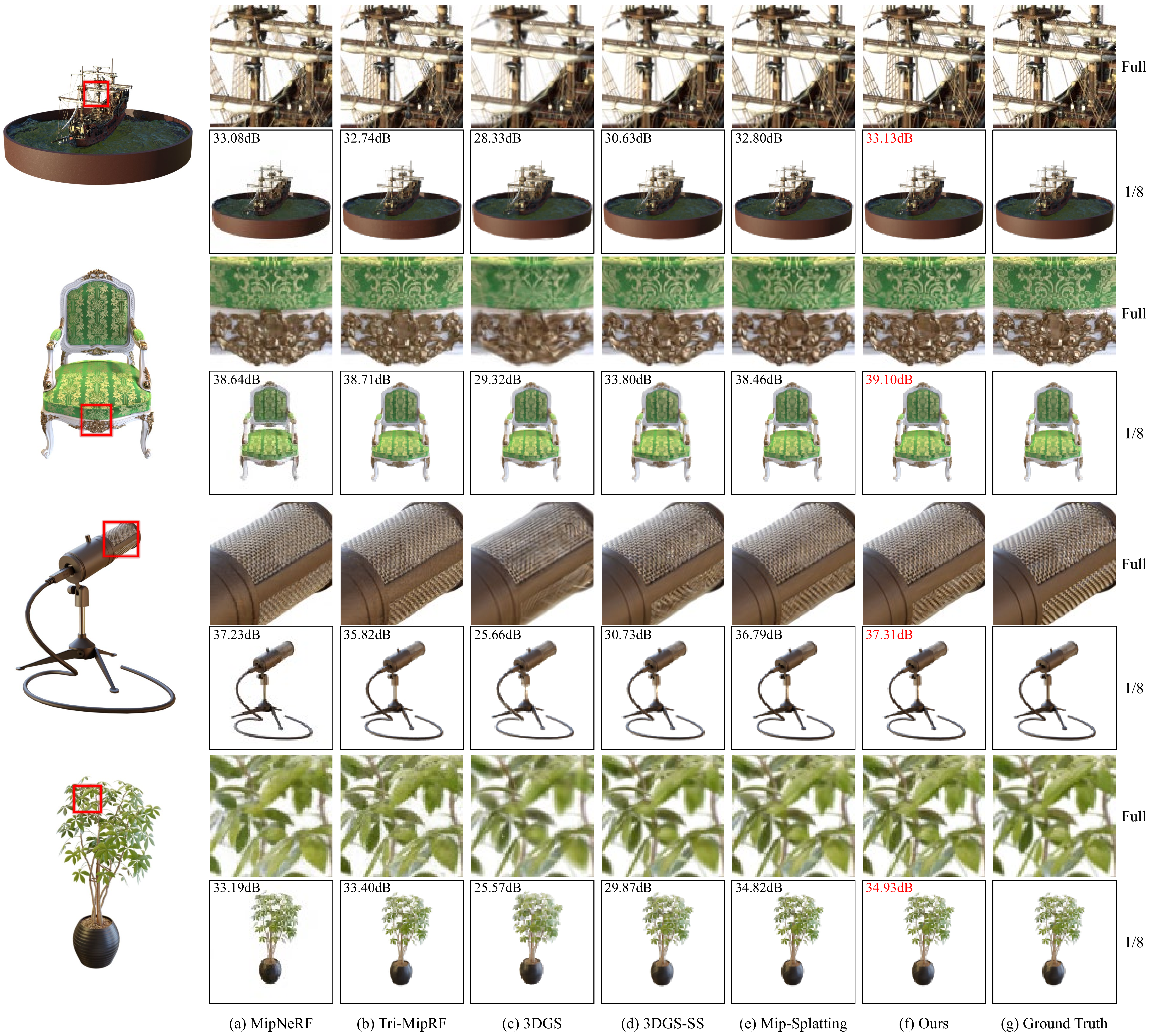}
}
\vspace{-0.2cm}
\caption{
    \textbf{Qualitative comparison of full-resolution and low-resolution ($\nicefrac{1}{8}$) on Multi-Scale Blender~\cite{barron2021mip}.} All methods are trained on images with downsampling rates covering [1, 2, 4, 8]. Our method can better overcome the artifacts in 3DGS with better fidelity of details.}
\label{fig:object_comparison}
\vspace{-.8cm}
\end{figure}

\noindent{\textbf{Implementation.}} We implement Analytic-Splatting upon 3DGS~\cite{kerbl20233d} and custom our shading module with CUDA extensions. Following 3DGS, we train Analytic-Splatting using the same parameters, training schedule, and loss functions, ensuring the efficacy of our scheme. To achieve super sampling of Gaussian signals, we implement 3DGS-SS, which first renders an image at twice the target resolution and obtains the final image at the target resolution through average pooling. For the MTMT setting, we follow previous works~\cite{barron2021mip, hu2023tri, yu2023mip} and tend to select more full-resolution images as supervision samples during training. Please refer to \textit{supplementary material} for more details on the backpropagation implementation of rendering.

\begin{table*}[tb]
\centering
\scalebox{0.58}{
    \begin{tabular}{l|c c c c c|c c c c c|c c c c c}
    & \multicolumn{5}{c|}{PSNR $\uparrow$} & \multicolumn{5}{c|}{SSIM $\uparrow$} & \multicolumn{5}{c}{LPIPS $\downarrow$} \\
    & Full Res. & $\nicefrac{1}{2}$ Res. & $\nicefrac{1}{4}$ Res.   & $\nicefrac{1}{8}$ Res. & Avg.
    & Full Res. & $\nicefrac{1}{2}$ Res. & $\nicefrac{1}{4}$ Res.   & $\nicefrac{1}{8}$ Res. & Avg.
    & Full Res. & $\nicefrac{1}{2}$ Res. & $\nicefrac{1}{4}$ Res.   & $\nicefrac{1}{8}$ Res. & Avg. \\
    \hline
    Mip-NeRF 360~\cite{barron2022mip} &
    \cc{2}27.50 & \cc{2}29.19 & \cc{2}30.45 & \cc{3}30.86 & \cc{3}29.50 &
    0.778 & 0.864 & 0.912 & 0.931 & 0.871 &
    0.254 & \cc{3}0.136 & \cc{3} 0.077 & 0.058 & \cc{3}0.131\\
    Mip-NeRF 360 + iNGP &
    26.46 & 27.92 & 27.67 & 25.58 & 26.91 &
    0.773 & 0.855 & 0.866 & 0.804 & 0.824 &
    0.253 & 0.142 & 0.117 & 0.159 & 0.167 \\
    Zip-NeRF~\cite{barron2023zip} &
    \cc{1}28.25 & \cc{1}30.01 & \cc{1}31.56 & \cc{1}32.52 & \cc{1}30.58 &
    \cc{1}0.822 & \cc{1}0.891 & \cc{1}0.933 & \cc{1}0.955 & \cc{1}0.900 &
    \cc{1}0.198 & \cc{1}0.099 & \cc{1}0.056 & \cc{1}0.038 & \cc{1}0.098\\
    \hline
    3DGS~\cite{kerbl20233d} &
    26.55 & 28.00 & 28.51 & 27.45 & 27.63 &
    0.779 & 0.854 & 0.891 & 0.888 & 0.853 &
    0.274 & 0.162 & 0.102 & 0.087 & 0.156 \\
    3DGS-SS~\cite{kerbl20233d} &
    27.20 & 28.75 & 29.89 & 29.71 & 28.89 &
    0.800 & \cc{3}0.871 & 0.914 & 0.928 & 0.878 &
    0.246 & 0.138 & 0.081 & 0.061 & \cc{3}0.131\\
    Mip-Splatting~\cite{yu2023mip} &
    27.20 & 28.74 & 29.90 & 30.66 & 29.12 &
    \cc{3}0.802 & 0.870 & \cc{3}0.915 & \cc{3}0.944 & \cc{3}0.883 &
    \cc{3}0.244 & 0.146 & 0.090 & \cc{3}0.056 & 0.134 \\
    \hline
    Ours &
    \cc{2}27.50 & \cc{3}28.99 & \cc{3}30.35 & \cc{2}31.21 & \cc{2}29.51 &
    \cc{2}0.808 & \cc{2}0.874 & \cc{2}0.919 & \cc{2}0.945 & \cc{2}0.887 &
    \cc{2}0.231 & \cc{2}0.132 & \cc{2}0.077 & \cc{2}0.051 & \cc{2}0.123
    \end{tabular}
}
\caption{\textbf{Quantatitive Comparison of Analytic-Splatting against several cutting-edge methods on the multi-scale Mip-NeRF 360 dataset~\cite{barron2022mip, barron2023zip}.} These methods conduct multi-scale training and testing.}
\label{tab:multi_360}
\vspace{-0.6cm}
\end{table*}

\noindent{\textbf{Evaluation on Blender Synthetic Dataset.}} We compare  our Analytic-Splatting with several state-of-the-art methods \ie NeRF~\cite{mildenhall2020nerf}, MipNeRF~\cite{barron2021mip}, Plenoxels~\cite{fridovich2022plenoxels}, TensoRF~\cite{chen2022tensorf}, Instant-NGP~\cite{muller2022instant}, Tri-MipRF~\cite{hu2023tri}, 3DGS~\cite{kerbl20233d} and its variants (\ie 3DGS-SS, Mip-Splatting~\cite{yu2023mip}) on the Blender Synthetic dataset. The quantitative results in \cref{tab:multi_blender} show that Analytic-Splatting outperforms other methods in all aspects. The qualitative results in \cref{fig:object_comparison} demonstrate that Analytic-Splatting can better capture high-frequency details while being anti-aliased. More results can be found in the \textit{supplementary material}.

\begin{figure}[tb]
    \centering
    \scalebox{1.0}
    {
        \includegraphics[width=\textwidth]{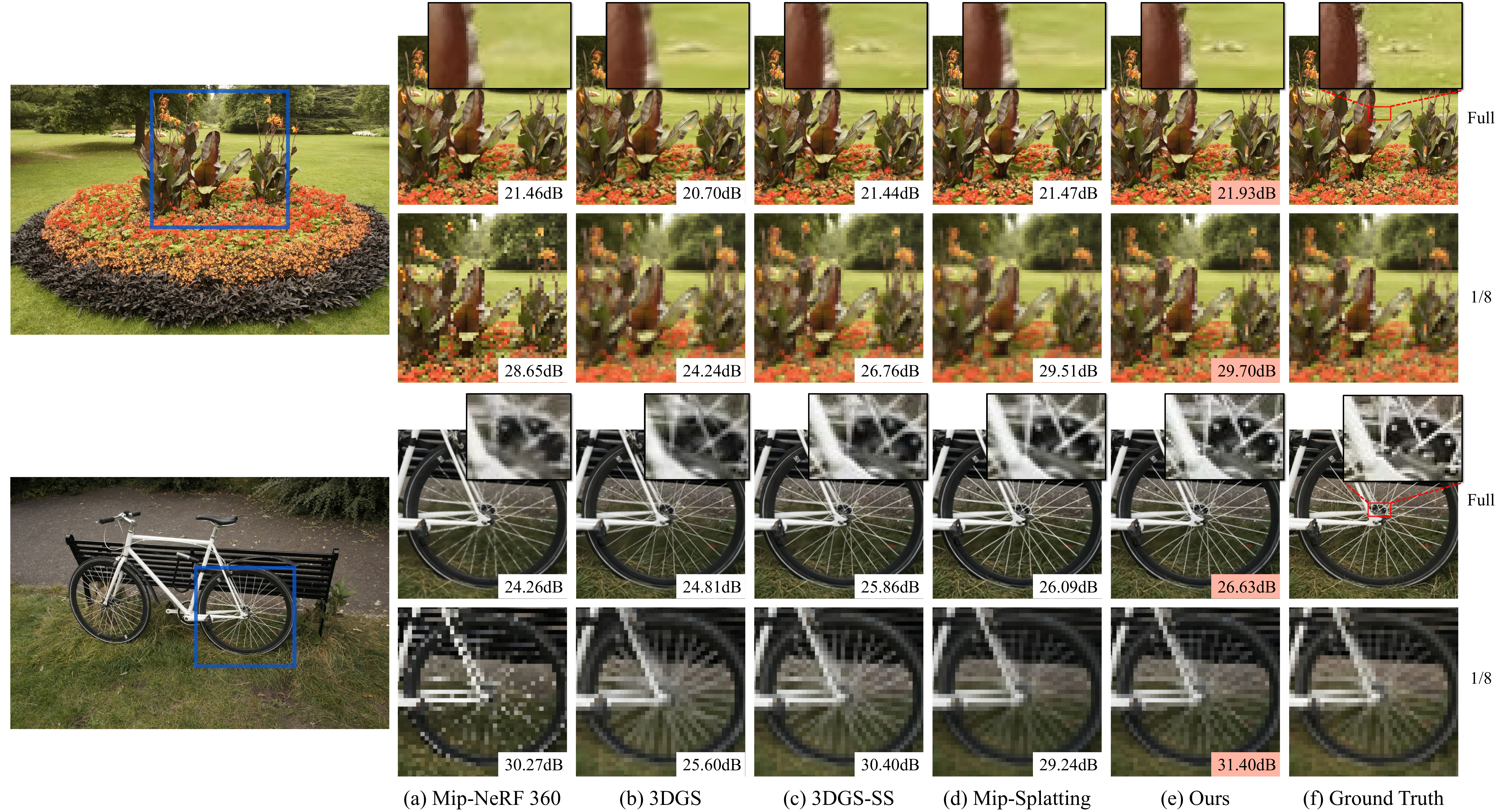}
    }
    \vspace{-0.2cm}
    \caption{\textbf{Qualitative comparisons of full-resolution and low-resolution on Multi-Scale Mip-NeRF 360.~\cite{barron2022mip, barron2023zip}} All methods are trained on images with downsampling rates covering [1, 2, 4, 8]. Our method can better overcome the artifacts with better fidelity of details. Please note that the artifacts of 3DGS/3DGS-SS become obvious at low resolutions, especially on elongated shapes (\eg wheels and flower stems), while Mip-Splatting produces over-smoothing results (\eg lobes).}
    \label{fig:scene_comparison}
    \vspace{-0.6cm}
\end{figure}

\noindent{\textbf{Evaluation on Mip-NeRF 360 Dataset.}} We compare our Analytic-Splatting with several cut-edge methods \ie Mip-NeRF 360 and its iNGP encoding version~\cite{barron2022mip}, Zip-NeRF~\cite{barron2023zip}, 3DGS~\cite{kerbl20233d} and its variants (\ie 3DGS-SS, Mip-Splatting~\cite{yu2023mip}) on the challenging Mip-NeRF 360 dataset. The results of Zip-NeRF are reported from the available official implementation \footnote{\href{https://github.com/jonbarron/camp_zipnerf}{https://github.com/jonbarron/camp\_zipnerf}}. Please note that Mip-NeRF 360 and Zip-NeRF struggle with real-time rendering, especially Zip-NeRF performs supersampling techniques in the rendering phase. The quantitative results in \cref{tab:multi_360} show that our method is second only to Zip-NeRF, and we outperform other methods with real-time rendering capability (\ie 3DGS and its variants). The qualitative comparisons in \cref{fig:scene_comparison} demonstrate that our method has better anti-aliasing capability and detail fidelity despite facing complex scenes. More results can be found in the \textit{supplementary material}.

\begin{figure}[tb]
\centering
\scalebox{0.96}{
\includegraphics[width=\textwidth]{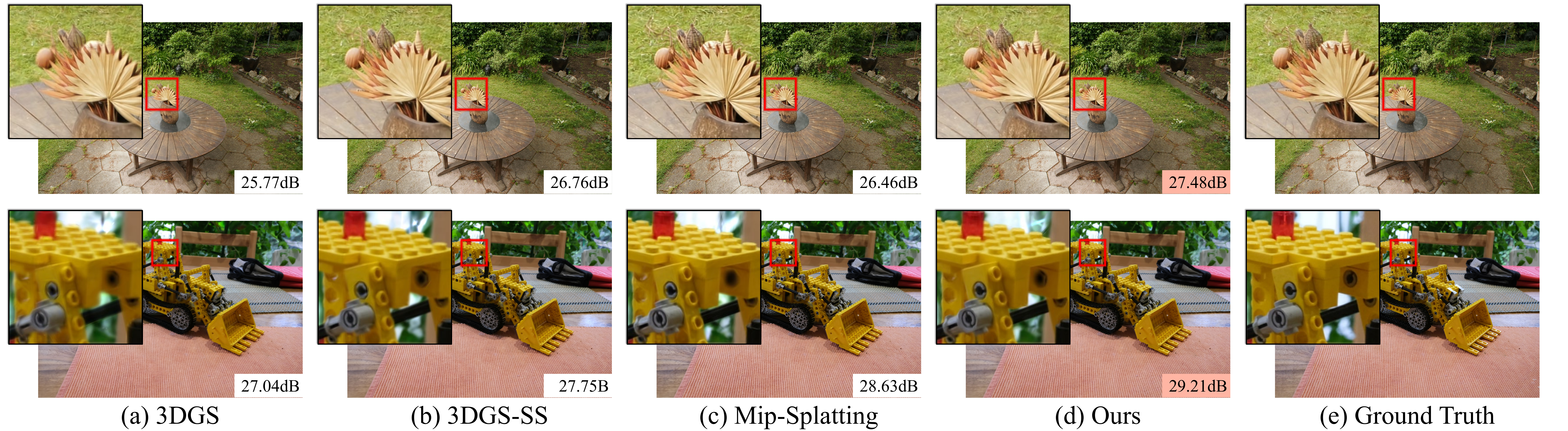}
}
\vspace{-0.2cm}
\caption{\textbf{Qualitative comparison of super-resolution (2$\times$) on Multi-Scale Mip-NeRF 360.~\cite{barron2022mip, barron2023zip}} All methods are trained on images with downsampling rates covering [1, 2, 4, 8]. Please note the high-frequency aliasing of 3DGS-SS and the over-smoothing of Mip-Splatting. Our results are closest to the ground truth.}
\label{fig:super_resolution}
\vspace{-0.6cm}
\end{figure}

\noindent{\textbf{Super-Resolution Evaluation on Mip-NeRF 360 Dataset.}}
We further evaluate our method against other 3DGS and its variants under the super-resolution setting ($2\times$ Res.) on the Mip-NeRF 360 dataset. All methods are trained on the Multi-Scale Mip-NeRF 360 dataset. The quantitative results are shown in \cref{tab:super_resolution} and the qualitative results are shown in \cref{fig:super_resolution}.
Both results demonstrate the superior performance of our method, further supporting the capability of Analytic-Splatting in capturing details. Conversely, Mip-Splatting is insufficient in capturing details due to pre-filtering.

\begin{table}[tb]
\centering
\scalebox{0.95}{
\begin{tabular}{l|c c c c c|c c c c|c}
& \multicolumn{10}{c}{\textbf{PSNR} $\uparrow$} \\
& bicycle & flowers & garden & stump & treehill
& room & counter & kitchen & bonsai & Avg. \\
\hline
3DGS & 23.14 & 20.28 & 24.62 & 25.44 & 21.90 &
30.27 & 28.08 & 29.51 & 30.34 & 25.95 \\
3DGS-SS & 23.98 & 20.84 & 25.48 & 26.24 & \textbf{22.12} &
30.90 & 28.69 & 30.53 & 31.34 & 26.68 \\
Mip-Splatting & 23.82 & 20.60 & 24.97 & 25.78 & 21.82 &
30.95 & 28.68 & 30.45 & 31.07 & 26.46 \\
Ours & \textbf{24.22} & \textbf{20.97} & \textbf{25.72} & \textbf{26.29} & 22.04 &
\textbf{31.04} & \textbf{28.90} & \textbf{31.10} & \textbf{31.83} & \textbf{26.90} \\
\multicolumn{11}{c}{} \\
& \multicolumn{10}{c}{\textbf{SSIM} $\uparrow$} \\
& bicycle & flowers & garden & stump & treehill
& room & counter & kitchen & bonsai & Avg. \\
\hline
3DGS & 0.639 & 0.492 & 0.707 & 0.706 & 0.578 &
0.902 & 0.891 & 0.893 & 0.916 & 0.747 \\
3DGS-SS & 0.675 & 0.527 & 0.747 & \textbf{0.739} & \textbf{0.596} &
0.908 & 0.901 & 0.907 & 0.926 & 0.769 \\
Mip-Splatting & 0.671 & 0.526 & 0.737 & 0.728 & 0.589 &
0.906 & 0.898 & 0.900 & 0.924 & 0.764 \\
Ours & \textbf{0.683} & \textbf{0.535} & \textbf{0.761} & \textbf{0.739} & \textbf{0.596} &
\textbf{0.910} & \textbf{0.904} & \textbf{0.911} & \textbf{0.930} & \textbf{0.774} \\
\multicolumn{11}{c}{} \\
& \multicolumn{10}{c}{\textbf{LPIPS} $\downarrow$} \\
& bicycle & flowers & garden & stump & treehill
& room & counter & kitchen & bonsai & Avg. \\
\hline
3DGS & 0.382 & 0.471 & 0.325 & 0.378 & 0.462 &
0.324 & 0.314 & 0.241 & 0.321 & 0.358 \\
3DGS-SS & 0.345 & 0.438 & 0.281 & 0.340 & 0.435 &
0.314 & 0.297 & 0.220 & 0.307 & 0.333 \\
Mip-Splatting & \textbf{0.341} & 0.433 & 0.291 & \textbf{0.338} & 0.439 &
0.309 & 0.295 & 0.216 & \textbf{0.300} & 0.329 \\
Ours & 0.342 & \textbf{0.429} & \textbf{0.268} & \textbf{0.338} & \textbf{0.434} &
\textbf{0.307} & \textbf{0.291} & \textbf{0.209} & \textbf{0.300} & \textbf{0.324} \\
\end{tabular}
}
\caption{\textbf{Quantatitive comparison of super-resolution (2$\times$) on Multi-Scale Mip-NeRF 360.~\cite{barron2022mip, barron2023zip}} All methods are trained on images with downsampling rates covering [1, 2, 4, 8]. we boldly mark the best results.}
\label{tab:super_resolution}
\vspace{-0.6cm}
\end{table}

\section{Conclusion}
In this paper, we first revisit the window response of one-dimensional Gaussian signals and reason about an analytical and accurate approximation using a conditioned logistic function. We then introduce this approximation in the two-dimensional pixel shading and present Analytic-Splatting, which approximates the pixel area integral response to achieve anti-aliasing capability and better detail fidelity. Our extensive experiments demonstrate the efficacy of Analytic-Splatting in achieving state-of-the-art novel view synthesis results under multi-scale and super-resolution settings.

\noindent\textbf{Limitations.} Compared with 3DGS and Mip-Splatting, our shading implementation introduces more root and exponential operations, which inevitably increases the computational burden and reduces the frame rate. Despite this, our frame rate is only $10\%$ lower than Mip-Splatting, which is also an anti-aliasing approach.


%
%
\bibliographystyle{splncs04}
\bibliography{main}

\appendix
\clearpage
\setcounter{page}{1}

\section{Shading Module}
\label{sec:shading}
Since we greatly improved the shading module, our implementation is quite different from the vanilla 3DGS, especially the backward part.
In this section, we give the forward and backward propagation step by step so that everyone can learn the details and reproduce our shading module easily.

\begin{figure}[h]
    \centering
\begin{subfigure}[b]{0.45\textwidth}
    \centering
    \includegraphics[width=\textwidth]{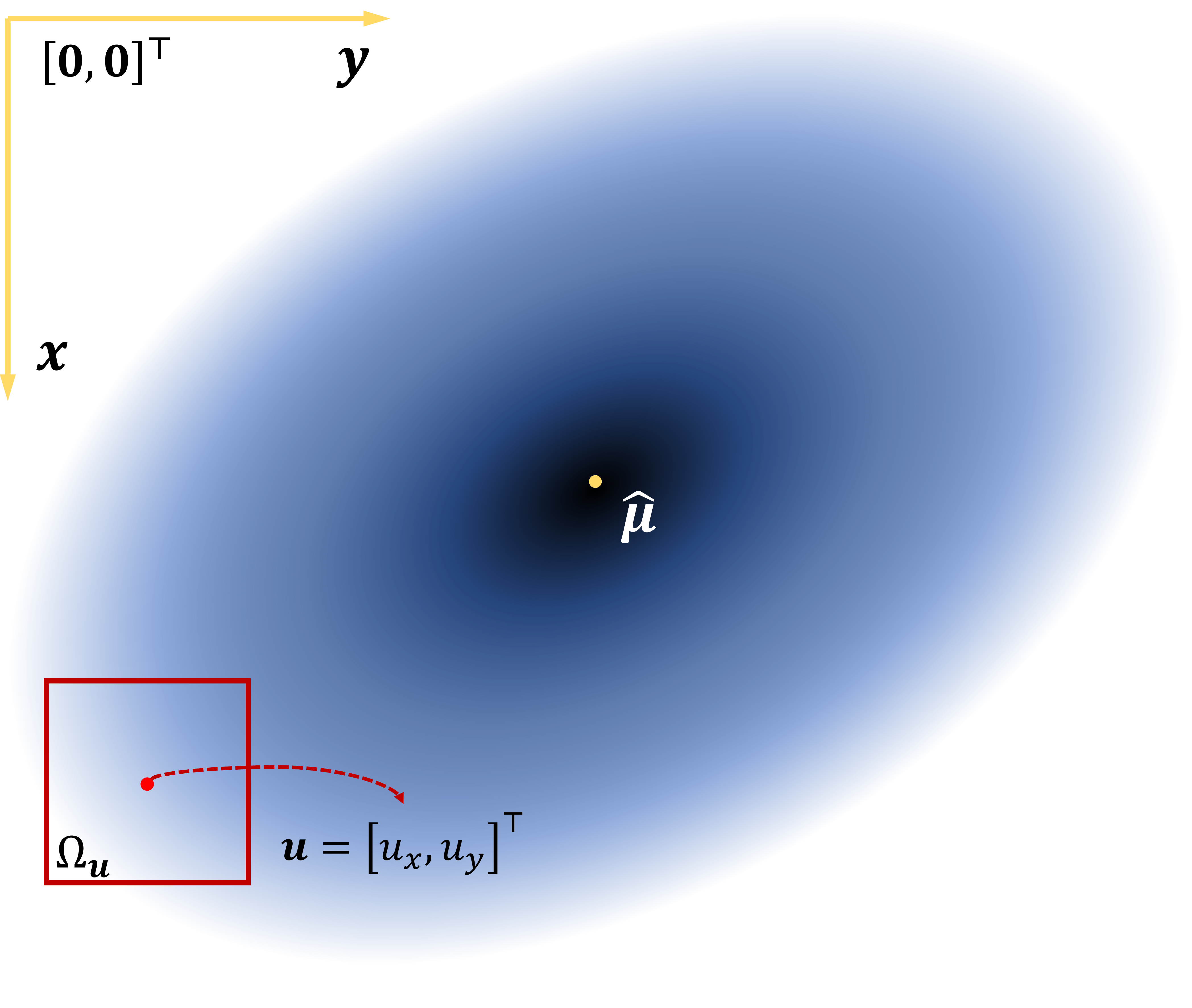}
    \caption{Integral over the original pixel area.}
    \label{fig:supp_integral_pixel}
\end{subfigure}
\hfill
\begin{subfigure}[b]{0.45\textwidth}
    \centering
    \includegraphics[width=\textwidth]{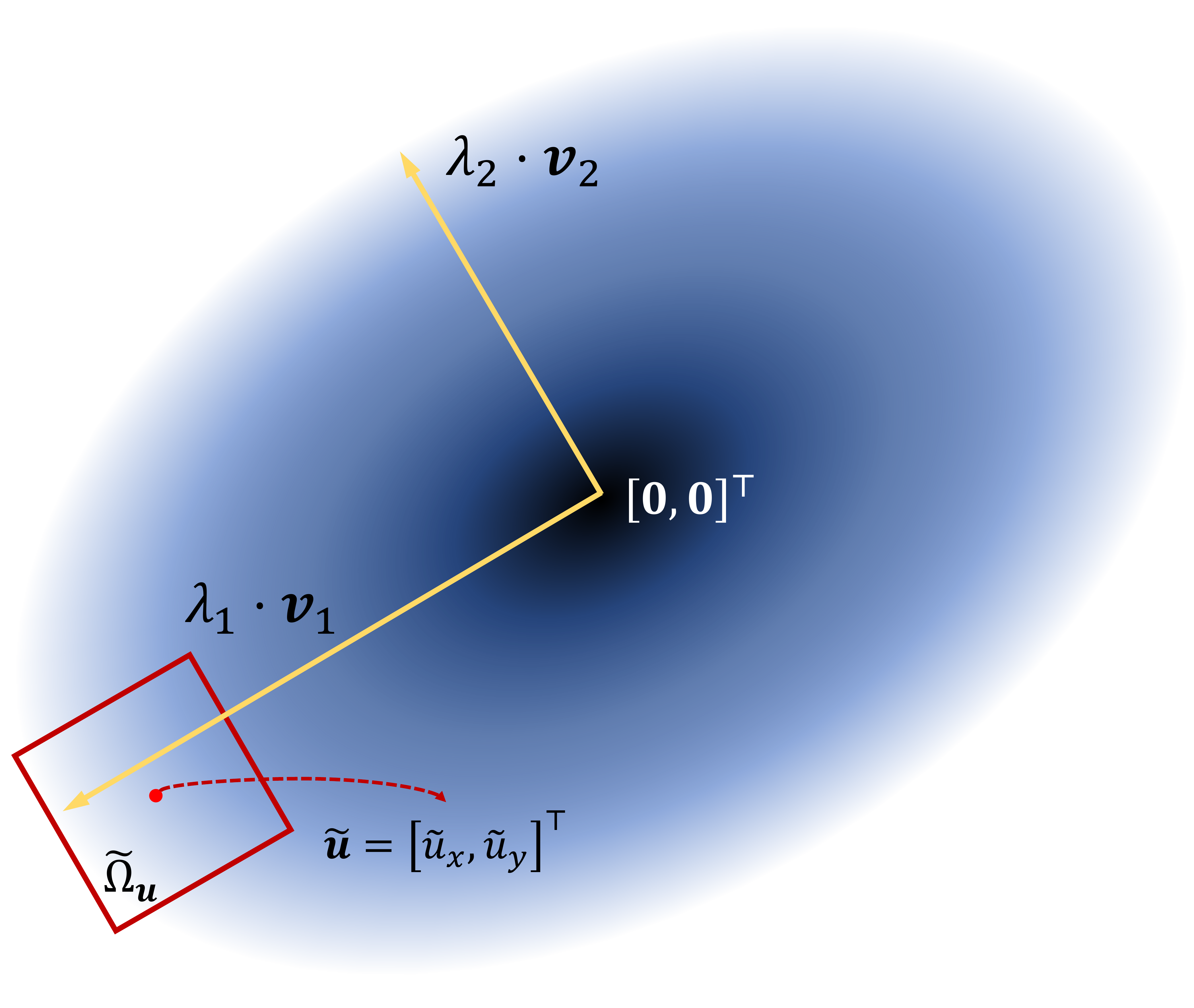}
    \caption{Integral over the rotated pixel area.}
    \label{fig:supp_integral_rotated_pixel}
\end{subfigure}
\vspace{-0.2cm}
\caption{Example diagram of the pixel integration domain and the domain after rotation. The yellow lines in \cref{fig:supp_integral_pixel} are the coordinate axes of 2D screen space; And the yellow lines in \cref{fig:supp_integral_rotated_pixel} are the eigenvectors scaled by the eigenvalues. }
\label{fig:supp_2D_approximation}
\vspace{-0.6cm}
\end{figure}

\subsection{Forward}
\label{subsec:forward}
Given a pixel with center $\bm{u} = [u_x, u_y]^\top$, we use a 2D Gaussian signal $g^\text{2D}$ to respond to the pixel and shade it.
Assuming that the 2D Gaussian signal $g^\text{2D}$ has a mean vector $\hat{\bm{\mu}} \in \mathbb{R}^2$ and a real-symmetric covariance matrix $\hat{\Sigma} \in \mathbb{R}^{2\times2}$. For better formulation, we express them as:
\begin{equation}
\hat{\bm{\mu}} =
\begin{bmatrix}
\hat{\mu}_x \\ \hat{\mu}_y
\end{bmatrix},\quad
\hat{\bm{\Sigma}} =
\begin{bmatrix}
\hat{\Sigma}_\text{11} & \hat{\Sigma}_\text{12} \\
\hat{\Sigma}_\text{12} & \hat{\Sigma}_\text{22} \\
\end{bmatrix}.
\label{eq:mu_and_Sigma}
\end{equation}

In Analytic-Splatting, we first perform eigendecomposition on $\hat{\bm{\Sigma}}$ to achieve diagonalization. After the decomposition, we obtain the eigenvalues $\{\lambda_1, \lambda_2\}$ and eigenvectors $\{\bm{v}_1, \bm{v}_2\}$ of $\hat{\bm{\Sigma}}$:
\begin{equation}
\small
\begin{aligned}
\lambda_1 = \frac{
    \text{Tr}(\hat{\bm{\Sigma}}) + \sqrt{
        \text{Tr}(\hat{\bm{\Sigma}})^2 - 4\det(\hat{\bm{\Sigma}})
    }
}{2},\quad&
\lambda_2 = \frac{
    \text{Tr}(\hat{\bm{\Sigma}}) - \sqrt{
        \text{Tr}(\hat{\bm{\Sigma}})^2 - 4\det(\hat{\bm{\Sigma}})
    }
}{2}, \\
\hat{\bm{v}}_1 =
\begin{bmatrix}
\hat{\Sigma}_\text{12} \\
\lambda_1 - \hat{\Sigma}_\text{11}
\end{bmatrix},\quad&
\hat{\bm{v}}_2 =
\begin{bmatrix}
\lambda_2 - \hat{\Sigma}_\text{22} \\
\hat{\Sigma}_\text{12}
\end{bmatrix}, \\
\bm{v}_1 =
\begin{bmatrix}
v_{1x} \\ v_{1y}
\end{bmatrix}
= \frac{\hat{\bm{v}}_1}{\Vert \hat{\bm{v}}_1 \Vert} = 
\begin{bmatrix}
\frac{\hat{\Sigma}_\text{12}}{\Vert \hat{\bm{v}}_1 \Vert} \\
\frac{\lambda_1 - \hat{\Sigma}_\text{11}}{\Vert \hat{\bm{v}}_1 \Vert}
\end{bmatrix},\quad&
\bm{v}_2 = 
\begin{bmatrix}
v_{2x} \\ v_{2y}
\end{bmatrix}
= \frac{\hat{\bm{v}}_2}{\Vert \hat{\bm{v}}_2 \Vert}=
\begin{bmatrix}
\frac{\lambda_2 - \hat{\Sigma}_\text{22}}{\Vert \hat{\bm{v}}_1 \Vert} \\
\frac{\hat{\Sigma}_\text{12}}{\Vert \hat{\bm{v}}_1 \Vert}
\end{bmatrix},\\
\end{aligned}
\label{eq:eigendecomposition}
\end{equation}
where
$\text{Tr}(\hat{\bm{\Sigma}}) = \hat{\Sigma}_{11} + \hat{\Sigma}_{22}$, 
$\det(\hat{\bm{\Sigma}}) = \hat{\Sigma}_{11} \hat{\Sigma}_{22} - \hat{\Sigma}_{12}^2$.

Then we use the eigenvectors $\{\bm{v}_1, \bm{v}_2\}$ to construct a new coordinate system with its origin at $\hat{\bm{\mu}} = [\hat{\bm{\mu}}_x, \hat{\bm{\mu}}_y]^\top$ (refer to the yellow lines in \cref{fig:integral_rotated_pixel}) to unravel the correlation in the covariance $\hat{\bm{\Sigma}}$. In this way, the coordinate of the pixel center $\bm{u}$ is transformed into $\tilde{\bm{u}}$:

\begin{equation}
\begin{aligned}
\tilde{\bm{u}} &=
\begin{bmatrix}
\tilde{u}_x \\ \tilde{u}_y
\end{bmatrix} = 
\begin{bmatrix}
- & \bm{v}_1 & - \\
- & \bm{v}_2 & -
\end{bmatrix}
\left(\bm{u} - \hat{\bm{\mu}}\right)
=
\begin{bmatrix}
v_{1x} & v_{1y} \\
v_{2x} & v_{2y}
\end{bmatrix}
\begin{bmatrix}
u_x - \hat{\mu}_x \\
u_y - \hat{\mu}_y
\end{bmatrix} \\
&=
\begin{bmatrix}
v_{1x} (u_x - \hat{\mu}_x) + v_{1y} (u_y - \hat{\mu}_y)
 \\
v_{2x} (u_x - \hat{\mu}_x) + v_{2y} (u_y - \hat{\mu}_y)
\end{bmatrix}
\end{aligned},
\label{eq:transform_center}
\end{equation}
then the intensity response of the pixel center can be written as:
\begin{equation}
g^\text{2D}(\bm{u}) = \exp\left(
-\frac{1}{2\lambda_1} \tilde{u}_x^2
\right) \exp\left(
-\frac{1}{2\lambda_2} \tilde{u}_y^2
\right)
\label{eq:center_response}.
\end{equation}

In Sec. 4.1 of the main page, we propose to use a conditioned logistic function  to approximate the cumulative distribution function (CDF) of the standard Gaussian distribution $g(x)$ as:
\begin{equation}
\scriptsize
\begin{aligned}
G(x) = \int^{x}_{-\infty} g(u) \text{d}u
= \int^{x}_{-\infty} \frac{1}{\sqrt{2\pi}}\exp\left(-\frac{u^2}{2}\right) \text{d}u
\approx S(x) = \frac{1}{1 + \exp(-1.6\cdot x - 0.07\cdot x^3)}
\end{aligned},
\label{eq:standard_CDF_approximation}
\end{equation}
and for the Gaussian distribution with standard deviation $\sigma \neq 1$, we use the reciprocal of $\sigma$ to scale $x$ and express the logistic function $S_\sigma$ as:
\begin{equation}
\begin{aligned}
G_\sigma(x) &=
\int^x_{-\infty} \frac{1}{\sigma\sqrt{2\pi}}\exp\left(
-\frac{u^2}{2\sigma^2}
\right) \text{d}u \\
&\approx S_\sigma(x)
= S\left(\frac{x}{\sigma}\right)
= \frac{1}{1 + \exp\left(
-1.6\cdot\frac{x}{\sigma} - 0.07 \cdot \left(
\frac{x}{\sigma}
\right)^3
\right)}.
\end{aligned}
\label{eq:CDF_approximation}
\end{equation}

Given the CDF approximation \cref{eq:CDF_approximation}, we approximate the response of 1-width window around sample $x$ as:
\begin{equation}
\begin{aligned}
\mathcal{I}_g(x) &=
\int^{x + \frac{1}{2}}_{x - \frac{1}{2}} \frac{1}{\sigma\sqrt{2\pi}}\exp\left(
-\frac{u^2}{2\sigma^2}
\right) \text{d}u \\
&= G_\sigma\left(
x + \frac{1}{2}
\right) - G_\sigma\left(
x - \frac{1}{2}
\right)
\approx S_\sigma\left(
x + \frac{1}{2}
\right) - S_\sigma\left(
x - \frac{1}{2}
\right)
\end{aligned}.
\label{eq:window_approximation}
\end{equation}

In Analytic-Splatting, we calculate the intensity response $\mathcal{I}^\text{2D}_g$ by approximating the integral of Gaussian signal \cref{eq:center_response} over the domain $\tilde{\Omega}_{\bm{u}}$ in \cref{fig:supp_2D_approximation}, and the integral over 2D domain $\tilde{\Omega}_{\bm{u}}$ can be represented as the product of the integrals of two one-dimensional Gaussian signals:
\begin{equation}
\scriptsize
\begin{aligned}
\mathcal{I}^\text{2D}_g
&\approx \int_{\tilde{\Omega}_{\bm{u}}} g^\text{2D}(\bm{u}) \text{d}\bm{u} \\
&=\int^{\tilde{u}_x + \frac{1}{2}}_{\tilde{u}_x - \frac{1}{2}} \exp\left(
-\frac{1}{2\lambda_1} x^2
\right) \text{d}x
\int^{\tilde{u}_y + \frac{1}{2}}_{\tilde{u}_y - \frac{1}{2}} \exp\left(
-\frac{1}{2\lambda_2} y^2
\right) \text{d}y \\
& = \sqrt{2\pi\lambda_1} \int^{\tilde{u}_x + \frac{1}{2}}_{\tilde{u}_x - \frac{1}{2}} \frac{1}{\sqrt{2\pi\lambda_1}}\exp\left(
-\frac{1}{2\lambda_1} x^2
\right) \text{d}x\cdot
\sqrt{2\pi\lambda_2}
\int^{\tilde{u}_y + \frac{1}{2}}_{\tilde{u}_y - \frac{1}{2}} \frac{1}{\sqrt{2\pi\lambda_2}}\exp\left(
-\frac{1}{2\lambda_2} y^2
\right) \text{d}y \\
& \approx 2\pi
\underbrace{
\sigma_1 \left[
    S_{\sigma_1}(\tilde{u}_x + \frac{1}{2}) - S_{\sigma_1}(\tilde{u}_x - \frac{1}{2})
\right]
}_{\mathcal{I}_{\sigma_1}}
\underbrace{
\sigma_2\left[
    S_{\sigma_2}(\tilde{u}_y + \frac{1}{2}) - S_{\sigma_2}(\tilde{u}_y - \frac{1}{2})
\right]
}_{\mathcal{I}_{\sigma_2}},
\end{aligned}
\label{eq:intensity_approximation}
\end{equation}
where $\sigma_1 = \sqrt{\lambda_1}, \sigma_2 = \sqrt{\lambda_2}$ denote the standard derivations of the independent Gaussian signals along two eigenvectors respectively. In summary, the volume shading in Analytic-Splatting is given by:
\begin{equation}
\small
\begin{aligned}
    \bm{C}(\bm{u}) & = \sum_{i\in N} T_i \mathcal{I}_{g-i}^\text{2D}(\bm{u} | \hat{\bm{\mu}_i}, \hat{\bm{\Sigma}_i})\alpha_i \bm{c}_i,\quad
    T_i = \prod^{i-1}_{j=1}(1 - \mathcal{I}_{g-j}^\text{2D}(\bm{u} | \hat{\bm{\mu}_j}, \hat{\bm{\Sigma}_j}) \alpha_j),   \\
    \mathcal{I}_{g}^\text{2D}(\bm{u})
    &= 2\pi\sigma_1\sigma_2\left[
        S_{\sigma_1}(\tilde{u}_x + \frac{1}{2}) - S_{\sigma_1}(\tilde{u}_x - \frac{1}{2})
        \right]\left[
        S_{\sigma_2}(\tilde{u}_y + \frac{1}{2}) - S_{\sigma_2}(\tilde{u}_y - \frac{1}{2})
    \right]\\
    &= 2\pi \mathcal{I}_{\sigma_1} \mathcal{I}_{\sigma_2}.
\end{aligned}
\label{eq:summary_forward}
\end{equation}

\subsection{Backward}
\label{subsec:backward}
\noindent\textbf{Derivation of the Conditioned Logistic Function}
Before introducing the backpropagation in Analytic-Splatting, we first give the derivation of the conditioned logistic function $S(x)$ in \cref{eq:standard_CDF_approximation}:
\begin{equation}
\begin{aligned}
\frac{\partial S(x)}{\partial x} &=
\left(1.6 + 0.21\cdot x^2\right)
\frac{
    \exp\left(
        -1.6\cdot x - 0.07 \cdot x^3
    \right)
}{\left[
    1 + \exp\left(
        -1.6\cdot x - 0.07 \cdot x^3
    \right)
\right]^2} \\
&= (1.6 + 0.21\cdot x^2)
S(x)[1 - S(x)]
\end{aligned},
\label{eq:original_logistic_backward}
\end{equation}
further for the derivation of $S_\sigma(x)$ in \cref{eq:CDF_approximation}, we can get
$\frac{\partial S_\sigma(x)}{\partial x}$ and
$\frac{\partial S_\sigma(x)}{\partial \sigma}$
through the chain rule:
\begin{equation}
\small
\begin{aligned}
\frac{\partial S_\sigma(x)}{\partial x}
&= \frac{\partial S\left(\frac{x}{\sigma}\right)}{\partial \frac{x}{\sigma}}
\frac{\partial \frac{x}{\sigma}}{\partial x} 
= \left(
    1.6 + 0.21\cdot \left(\frac{x}{\sigma}\right)^2
\right)
S_\sigma\left(\frac{x}{\sigma}\right)
\left[1 - S_\sigma\left(\frac{x}{\sigma}\right)\right]\cdot
\frac{1}{\sigma} \\
\frac{\partial S_\sigma(x)}{\partial \sigma}
&= \frac{\partial S\left(\frac{x}{\sigma}\right)}{\partial \frac{x}{\sigma}}
\frac{\partial \frac{x}{\sigma}}{\partial \sigma}
= \left(
    1.6 + 0.21\cdot \left(\frac{x}{\sigma}\right)^2
\right)
S_\sigma\left(\frac{x}{\sigma}\right)
\left[1 - S_\sigma\left(\frac{x}{\sigma}\right)\right]\cdot
-\frac{x}{\sigma^2}
\end{aligned}
\label{eq:logsitic_backward}
\end{equation}

\noindent\textbf{Backpropogation in Shading}
Our backpropagation aims to derive the gradients of $\mathcal{I}_g^\text{2D}$ with respect to $\hat{\bm{\mu}}$ and $\hat{\bm{\Sigma}}$, as $\frac{\partial \mathcal{I}_g^\text{2D}}{\partial \hat{\bm{\mu}}}$ and $\frac{\partial \mathcal{I}_g^\text{2D}}{\partial \hat{\bm{\Sigma}}}$:
\begin{equation}
\frac{\partial \mathcal{I}_g^\text{2D}}{\partial \hat{\bm{\mu}}} =
\begin{bmatrix}
\nicefrac{\partial \mathcal{I}_g^\text{2D}}{\partial \hat{\mu}_x} \\
\nicefrac{\partial \mathcal{I}_g^\text{2D}}{\partial \hat{\mu}_y}
\end{bmatrix} \in \mathbb{R}^2,\quad
\frac{\partial \mathcal{I}_g^\text{2D}}{\partial \hat{\bm{\Sigma}}} =
\begin{bmatrix}
\nicefrac{\partial \mathcal{I}_g^\text{2D}}{\partial \hat{\Sigma}_\text{11}} & &
\nicefrac{\partial \mathcal{I}_g^\text{2D}}{\partial \hat{\Sigma}_\text{12}} \\
\nicefrac{\partial \mathcal{I}_g^\text{2D}}{\partial \hat{\Sigma}_\text{12}} & &
\nicefrac{\partial \mathcal{I}_g^\text{2D}}{\partial \hat{\Sigma}_\text{22}} \\
\end{bmatrix} \in \mathbb{R}^{2\times2}.
\label{eq:mu_and_Sigma_gradient}
\end{equation}

It is quite difficult to express the above gradient directly. Still, we can use the chain rule to boil down the above results layer by layer to be available.
We note that our key insight is to construct a new coordinate system using the mean vector $\hat{\bm{\mu}}$, eigenvalues $\{\lambda_1, \lambda_2\}$ and eigenvectors $\{\bm{v}_1, \bm{v}_2\}$.
Therefore, we can use them as an intermediary to solve the final gradient by the chain rule.
For the gradients of $\mathcal{I}_g^\text{2D}$ with respect to the mean vector $\hat{\bm{\mu}}$, according to \cref{eq:transform_center} and \cref{eq:summary_forward}, we have:
\begin{equation}
\begin{aligned}
\frac{\partial \mathcal{I}^\text{2D}_{g}}{\partial \hat{\bm{\mu}}}
=& 2\pi \left(
\mathcal{I}_{\sigma_2} \frac{
    \partial \mathcal{I}_{\sigma_1}
}{
    \partial \hat{\bm{\mu}}
} + \mathcal{I}_{\sigma_1} \frac{
    \partial \mathcal{I}_{\sigma_2}
}{
    \partial \hat{\bm{\mu}}
}
\right)
= 2\pi \left(
\mathcal{I}_{\sigma_2} \frac{
    \partial \mathcal{I}_{\sigma_1}
}{
    \partial \tilde{u}_x
} \frac{
    \partial \tilde{u}_x
}{
    \partial \hat{\bm{\mu}}
} + \mathcal{I}_{\sigma_1} \frac{
    \partial \mathcal{I}_{\sigma_2}
}{
    \partial \tilde{u}_y
} \frac{
    \partial \tilde{u}_y
}{
    \partial \hat{\bm{\mu}}
}
\right) \\
=& 2\pi
\mathcal{I}_{\sigma_2} \left[
\frac{\partial S_{\sigma_1}(\tilde{u}_x + \frac{1}{2})}{\partial \tilde{u}_x}
\frac{\partial \tilde{u}_x}{\partial \hat{\bm{\mu}}} - 
\frac{\partial S_{\sigma_1}(\tilde{u}_x - \frac{1}{2})}{\partial \tilde{u}_x}
\frac{\partial \tilde{u}_x}{\partial \hat{\bm{\mu}}}
\right] + \\
&2\pi \mathcal{I}_{\sigma_1} \left[
\frac{\partial S_{\sigma_2}(\tilde{u}_y + \frac{1}{2})}{\partial \tilde{u}_y}
\frac{\partial \tilde{u}_y}{\partial \hat{\bm{\mu}}} - 
\frac{\partial S_{\sigma_2}(\tilde{u}_y - \frac{1}{2})}{\partial \tilde{u}_y}
\frac{\partial \tilde{u}_y}{\partial \hat{\bm{\mu}}}
\right],
\end{aligned}
\label{eq:gradient_meanvector}
\end{equation}
where $\frac{\partial \tilde{u}_x}{\partial \hat{\bm{\mu}}} = [-v_{1x}, -v_{1y}]^\top$, $\frac{\partial \tilde{u}_y}{\partial \hat{\bm{\mu}}} = [-v_{2x}, -v_{2y}]^\top$, and the gradient $\frac{\partial S_\sigma(x)}{\partial x}$ has been solved in \cref{eq:logsitic_backward}.

According to \cref{eq:summary_forward}, we have the gradients of $\mathcal{I}_g^\text{2D}$ to eigenvalues $\{\lambda_1, \lambda_2\}$:
\begin{equation}
\scriptsize
\begin{aligned}
\frac{\partial \mathcal{I}^\text{2D}_{g}}{\partial \lambda_1} &=
\frac{\partial \mathcal{I}^\text{2D}_{g}}{\partial \sigma_1}
\frac{\partial \sigma_1}{\partial \lambda_1} =
2\pi\mathcal{I}_{\sigma_2}\frac{\partial \mathcal{I}_{\sigma_1}}{\partial \sigma_1} \cdot \frac{1}{2\sqrt{\lambda_1}} \\
&= \frac{\pi \mathcal{I}_{\sigma_2}}{\sqrt{\lambda_1}} \left(
\left[
    S_{\sigma_1}(\tilde{u}_x + \frac{1}{2}) - S_{\sigma_1}(\tilde{u}_x - \frac{1}{2})
\right] + \sigma_1 \left[
    \frac{\partial S_{\sigma_1}(\tilde{u}_x + \frac{1}{2})}
    {\partial \sigma_1} -
    \frac{\partial S_{\sigma_1}(\tilde{u}_x - \frac{1}{2})}
    {\partial \sigma_1}
\right]
\right), \\
\frac{\partial \mathcal{I}^\text{2D}_{g}}{\partial \lambda_2} &=
\frac{\partial \mathcal{I}^\text{2D}_{g}}{\partial \sigma_2}
\frac{\partial \sigma_2}{\partial \lambda_2} =
2\pi\mathcal{I}_{\sigma_1}\frac{\partial \mathcal{I}_{\sigma_2}}{\partial \sigma_2} \cdot \frac{1}{2\sqrt{\lambda_2}} \\
&= \frac{\pi \mathcal{I}_{\sigma_1}}{\sqrt{\lambda_2}} \left(
\left[
    S_{\sigma_2}(\tilde{u}_y + \frac{1}{2}) - S_{\sigma_2}(\tilde{u}_y - \frac{1}{2})
\right] + \sigma_2 \left[
    \frac{\partial S_{\sigma_2}(\tilde{u}_y + \frac{1}{2})}
    {\partial \sigma_2} -
    \frac{\partial S_{\sigma_2}(\tilde{u}_y - \frac{1}{2})}
    {\partial \sigma_2}
\right]
\right), \\
\end{aligned}
\label{eq:gradient_eigenvalues}
\end{equation}
please note that the gradient $\frac{\partial S_\sigma(x)}{\partial \sigma}$ has been solved in \cref{eq:logsitic_backward}. Given \cref{eq:transform_center} and \cref{eq:summary_forward}, the gradients of $\mathcal{I}_g^\text{2D}$ with respect to the eigenvectors $\{\bm{v}_1, \bm{v}_2\}$ are:
\begin{equation}
\small
\begin{aligned}
\frac{\partial \mathcal{I}^\text{2D}_{g}}{\partial \bm{v}_1}
=& \frac{\partial \mathcal{I}^\text{2D}_{g}}{\partial \tilde{u}_x}
\frac{\partial \tilde{u}_x}{\partial \bm{v}_1} = 2\pi\mathcal{I}_{\sigma2} \frac{\partial \mathcal{I}_{\sigma_1}}{\partial \tilde{u}_x}
\begin{bmatrix}
u_x - \hat{\mu}_x \\ u_y - \hat{\mu}_y
\end{bmatrix} \\
=& 2\pi\mathcal{I}_{\sigma2}
\left(
\frac{\partial S_{\sigma_1}(\tilde{u}_x + \frac{1}{2})}{\partial \tilde{u}_x} -
\frac{\partial S_{\sigma_1}(\tilde{u}_x - \frac{1}{2})}{\partial \tilde{u}_x}
\right)
\begin{bmatrix}
u_x - \hat{\mu}_x \\ u_y - \hat{\mu}_y
\end{bmatrix} \in \mathbb{R}^2,
\\
\frac{\partial \mathcal{I}^\text{2D}_{g}}{\partial \bm{v}_2}
=& \frac{\partial \mathcal{I}^\text{2D}_{g}}{\partial \tilde{u}_y}
\frac{\partial \tilde{u}_y}{\partial \bm{v}_2} = 2\pi\mathcal{I}_{\sigma1} \frac{\partial \mathcal{I}_{\sigma_2}}{\partial \tilde{u}_y}
\begin{bmatrix}
u_x - \hat{\mu}_x \\ u_y - \hat{\mu}_y
\end{bmatrix} \\
=& 2\pi\mathcal{I}_{\sigma1}
\left(
\frac{\partial S_{\sigma_2}(\tilde{u}_y + \frac{1}{2})}{\partial \tilde{u}_y} -
\frac{\partial S_{\sigma_2}(\tilde{u}_y - \frac{1}{2})}{\partial \tilde{u}_y}
\right)
\begin{bmatrix}
u_x - \hat{\mu}_x \\ u_y - \hat{\mu}_y
\end{bmatrix} \in \mathbb{R}^2. \\
\end{aligned}
\label{eq:gradient_eigenvectors}
\end{equation}

In addition, please refer to \cite{petersen2008matrix} and learn the gradients of eigenvalues with respect to the covariance $\{\frac{\partial\lambda_1}{\partial\hat{\Sigma}_{ij}}, \frac{\partial\lambda_2}{\partial\hat{\Sigma}_{ij}}\}$ and the gradients of eigenvectors with respect to the covariance $\{\frac{\partial\bm{v}_1}{\partial\hat{\Sigma}_{ij}}, \frac{\partial\bm{v}_2}{\partial\hat{\Sigma}_{ij}}\}$. According to \cref{eq:gradient_eigenvalues} and \cref{eq:gradient_eigenvectors}, we get the final gradient:
\begin{equation}
\small
\begin{aligned}
\frac{\partial \mathcal{I}^\text{2D}_g}{\partial\hat{\Sigma}_{11}}
&= \frac{\partial \mathcal{I}^\text{2D}_{g}}{\partial \lambda_1}
\frac{\partial \lambda_1}{\partial \hat{\Sigma}_{11}} +
\frac{\partial \mathcal{I}^\text{2D}_{g}}{\partial \lambda_2}
\frac{\partial \lambda_2}{\partial \hat{\Sigma}_{11}} +
\left(\frac{\partial \mathcal{I}^\text{2D}_{g}}{\partial \bm{v}_1}\right)^\top
\frac{\partial \bm{v}_1}{\partial \hat{\Sigma}_{11}} +
\left(\frac{\partial \mathcal{I}^\text{2D}_{g}}{\partial \bm{v}_2}\right)^\top
\frac{\partial \bm{v}_2}{\partial \hat{\Sigma}_{11}} \\
\frac{\partial \mathcal{I}^\text{2D}_g}{\partial\hat{\Sigma}_{12}}
&= \frac{\partial \mathcal{I}^\text{2D}_{g}}{\partial \lambda_1}
\frac{\partial \lambda_1}{\partial \hat{\Sigma}_{12}} +
\frac{\partial \mathcal{I}^\text{2D}_{g}}{\partial \lambda_2}
\frac{\partial \lambda_2}{\partial \hat{\Sigma}_{12}} +
\left(\frac{\partial \mathcal{I}^\text{2D}_{g}}{\partial \bm{v}_1}\right)^\top
\frac{\partial \bm{v}_1}{\partial \hat{\Sigma}_{12}} +
\left(\frac{\partial \mathcal{I}^\text{2D}_{g}}{\partial \bm{v}_2}\right)^\top
\frac{\partial \bm{v}_2}{\partial \hat{\Sigma}_{12}} \\
\frac{\partial \mathcal{I}^\text{2D}_g}{\partial\hat{\Sigma}_{22}}
&= \frac{\partial \mathcal{I}^\text{2D}_{g}}{\partial \lambda_1}
\frac{\partial \lambda_1}{\partial \hat{\Sigma}_{22}} +
\frac{\partial \mathcal{I}^\text{2D}_{g}}{\partial \lambda_2}
\frac{\partial \lambda_2}{\partial \hat{\Sigma}_{22}} +
\left(\frac{\partial \mathcal{I}^\text{2D}_{g}}{\partial \bm{v}_1}\right)^\top
\frac{\partial \bm{v}_1}{\partial \hat{\Sigma}_{22}} +
\left(\frac{\partial \mathcal{I}^\text{2D}_{g}}{\partial \bm{v}_2}\right)^\top
\frac{\partial \bm{v}_2}{\partial \hat{\Sigma}_{22}} \\
\end{aligned}.
\end{equation}

\section{Additional Results}
\label{sec:additional}
In this section, we report more quantitative and qualitative results in detail. In addition, we provide the results of our method combined with 3D smoothing filter~\cite{yu2023mip} for better study.
According to the quantitative results of our experiments, the conclusion is that the 3D smoothing filter slightly improves the single-scale training and testing results. Still, it has no significant impact on the multi-scale training and testing results.

\begin{table}[!htbp]
\centering
\scalebox{0.87}{
\begin{tabular}{l|c c c|c c c|c c c}
 & \multicolumn{3}{c|}{Mip-NeRF 360}
 & \multicolumn{3}{c|}{Tanks$\&$Temples}
 & \multicolumn{3}{c}{Deep Blending} \\
 & PSNR $\uparrow$ & SSIM $\uparrow$ & LPIPS $\downarrow$
 & PSNR $\uparrow$ & SSIM $\uparrow$ & LPIPS $\downarrow$
 & PSNR $\uparrow$ & SSIM $\uparrow$ & LPIPS $\downarrow$ \\
\hline
Plenoxels~\cite{fridovich2022plenoxels}
& 23.08 & 0.625 & 0.463
& 21.07 & 0.721 & 0.379
& 23.06 & 0.795 & 0.510 \\
INGP-Base~\cite{muller2022instant}
& 25.30 & 0.671 & 0.371
& 21.72 & 0.734 & 0.330
& 23.62 & 0.797 & 0.423 \\
INGP-Big~\cite{muller2022instant}
& 25.59 & 0.699 & 0.331
& 21.92 & 0.752 & 0.305
& 24.96 & 0.817 & 0.390 \\
Mip-NeRF 360~\cite{barron2022mip}
& \cc{1}27.69 & 0.792 & 0.237
& 22.22 & 0.800 & 0.257
& 29.40 & 0.901 & \cc{2}0.245 \\
3DGS~\cite{kerbl20233d}
& 27.21 & 0.815 & \cc{1}0.214
& 23.14 & 0.844 & 0.183
& 29.41 & 0.903 & \cc{1}0.243 \\
Mip-Splatting~\cite{yu2023mip}
& 27.57 & \cc{2}0.817 & 0.218
& \cc{3}23.78 & \cc{2}0.851 & \cc{3}0.178
& \cc{3}29.69 & \cc{3}0.904 & 0.248 \\
\hline
Ours
& \cc{2}27.58 & \cc{3}0.816 & \cc{2}0.217
& \cc{2}23.84 & \cc{2}0.851 & \cc{1}0.177
& \cc{1}29.75 & \cc{2}0.905 & 0.248 \\
Ours + 3D filter
& \cc{2}27.58 & \cc{1}0.818 & \cc{2}0.217
& \cc{1}23.91 & \cc{1}0.853 & \cc{1}0.177
& \cc{2}29.71 & \cc{1}0.906 & \cc{3}0.247 \\
\end{tabular}
}
\caption{\textbf{Quantitative comparison of our method against previous methods over three datasets~\cite{barron2022mip, knapitsch2017tanks, hedman2018deep}.} All methods are trained on full-resolution images and tested on the same-resolution images.}
\label{tab:single_scene_summary}
\vspace{-0.8cm}
\end{table}

\begin{table}[h]
\centering
\scalebox{0.72}{
\begin{tabular}{|l||c c|c c||c c|c c||c c|c c|}
\hline
& \multicolumn{4}{c||}{\textbf{PSNR} $\uparrow$}
& \multicolumn{4}{c||}{\textbf{SSIM} $\uparrow$}
& \multicolumn{4}{c|}{\textbf{LPIPS} $\downarrow$} \\
\cline{2-13}
& Truck & Train & Johnson & Playroom
& Truck & Train & Johnson & Playroom
& Truck & Train & Johnson & Playroom \\
\hline
Plenoxels~\cite{fridovich2022plenoxels}
& 23.22 & 18.93 & 23.14 & 22.98
& 0.774 & 0.663 & 0.787 & 0.802
& 0.335 & 0.422 & 0.521 & 0.499 \\
INGP-Base~\cite{muller2022instant}
& 23.26 & 20.17 & 27.75 & 19.48
& 0.779 & 0.666 & 0.839 & 0.754
& 0.274 & 0.386 & 0.381 & 0.465 \\
INGP-Big~\cite{muller2022instant}
& 23.83 & 20.46 & 28.26 & 21.67
& 0.800 & 0.689 & 0.854 & 0.779
& 0.249 & 0.360 & 0.352 & 0.428 \\
Mip-NeRF 360~\cite{barron2022mip}
& 24.91 & 19.52 & \cc{3}29.14 & 29.66
& 0.857 & 0.660 & \cc{2}0.901 & 0.900
& 0.159 & 0.354 & \cc{1}0.237 & 0.252 \\
3DGS~\cite{kerbl20233d}
& 25.19 & 21.09 & 28.77 & 30.04
& 0.879 & 0.802 & 0.899 & 0.906
& \cc{2}0.148 & 0.218 & \cc{2}0.244 & \cc{1}0.241 \\
Mip-Splatting~\cite{yu2023mip}
& \cc{2}25.48 & \cc{3}22.09 & 29.09 & \cc{2}30.28
& \cc{2}0.884 & \cc{3}0.818 & \cc{2}0.901 & \cc{3}0.908
& 0.149 & \cc{3}0.208 & 0.248 & \cc{2}0.248 \\
\hline
Ours
& \cc{2}25.48 & \cc{2}22.20 & \cc{1}29.18 & \cc{1}30.32
& \cc{3}0.883 & \cc{2}0.819 & \cc{2}0.901 & \cc{1}0.909
& \cc{2}0.148 & \cc{2}0.207 & \cc{3}0.247 & \cc{2}0.248 \\
Ours + 3D filter
& \cc{1}25.60 & \cc{1}22.22 & \cc{2}29.17 & \cc{3}30.25
& \cc{1}0.887 & \cc{1}0.820 & \cc{1}0.903 & \cc{1}0.909
& \cc{1}0.147 & \cc{1}0.206 & \cc{3}0.247 & \cc{2}0.248 \\
\hline
\end{tabular}
}
\caption{\textbf{Quantitative comparison of our method against previous methods on Tanks\&Temples~\cite{knapitsch2017tanks} and Deep Blending~\cite{hedman2018deep}.} All methods are trained on full-resolution images and tested on the same-resolution images.}
\label{tab:tan_db_single}
\vspace{-0.8cm}
\end{table}

\begin{table}[!tb]
\centering
\scalebox{0.9}{
\begin{tabular}{l|c c c c c|c c c c|c}
& \multicolumn{10}{c}{\textbf{PSNR} $\uparrow$} \\
& bicycle & flowers & garden & stump & treehill
& room & counter & kitchen & bonsai & Avg. \\
\hline
Plenoxels~\cite{fridovich2022plenoxels} &
21.91 & 20.10 & 23.49 & 20.66 & 22.25 &
27.59 & 23.62 & 23.42 & 24.67 & 23.08 \\
INGP-Base~\cite{muller2022instant} &
22.19 & 20.35 & 24.60 & 23.63 & 22.36 &
29.27 & 26.44 & 28.55 & 30.34 & 25.30 \\
INGP-Big~\cite{muller2022instant} &
22.17 & 20.65 & 25.07 & 23.47 & 22.37 &
29.69 & 26.69 & 29.48 & 30.69 & 25.59 \\
Mip-NeRF 360~\cite{barron2022mip} &
24.37 & \cc{1}21.73 & 26.98 & 26.40 & \cc{1}22.87 &
\cc{3}31.63 & \cc{1}29.55 & \cc{1}32.23 & \cc{1}33.46 & \cc{1}27.69 \\
3DGS~\cite{kerbl20233d} &
\cc{3}25.25 & 21.52 & \cc{3}27.41 & 26.55 & 22.49 &
30.63 & 28.70 & 30.32 & 31.98 & 27.21 \\
Mip-Splatting~\cite{yu2023mip} &
\cc{2}25.31 & \cc{3}21.62 & \cc{2}27.45 & \cc{3}26.62 & \cc{2}22.62 &
31.62 & \cc{2}29.11 & \cc{3}31.53 & \cc{3}32.30 & 27.57 \\
\hline
Ours &
25.18 & 21.61 & 27.39 & \cc{2}26.65 & 22.54 &
\cc{1}31.75 & \cc{2}29.11 & \cc{2}31.56 & \cc{2}32.43 & \cc{2}27.58 \\
Ours + 3D filter &
\cc{1}25.32 & \cc{2}21.64 & \cc{1}27.51 & \cc{1}26.68 & \cc{3}22.59 &
\cc{2}31.66 & 29.04 & 31.50 & \cc{3}32.30 & \cc{2}27.58 \\
\Xhline{3\arrayrulewidth}
3DGS*~\cite{kerbl20233d} &
25.63 & 21.77 & 27.70 & 26.87 & 22.75 &
31.69 & 29.08 & 31.56 & 32.29 & 27.76 \\
Mip-Splatting*~\cite{yu2023mip} &
25.72 & 21.93 & 27.76 & 26.94 & 22.98 &
31.74 & 29.16 & 31.55 & 32.31 & 27.79 \\
Ours* &
25.63 & 21.92 & 27.73 & 26.92 & 22.79 &
31.89 & 29.24 & 31.66 & 32.60 & 27.82 \\
Ours + 3D filter* &
25.70 & 21.92 & 27.78 & 26.98 & 22.95 &
31.86 & 29.17 & 31.74 & 32.55 & 27.85 \\
\multicolumn{11}{c}{} \\
& \multicolumn{10}{c}{\textbf{SSIM} $\uparrow$} \\
& bicycle & flowers & garden & stump & treehill
& room & counter & kitchen & bonsai & Avg. \\
\hline
Plenoxels~\cite{fridovich2022plenoxels} &
0.496 & 0.431 & 0.606 & 0.523 & 0.509 &
0.842 & 0.759 & 0.648 & 0.814 & 0.625 \\
INGP-Base~\cite{muller2022instant} &
0.491 & 0.450 & 0.649 & 0.574 & 0.518 &
0.855 & 0.798 & 0.818 & 0.890 & 0.671 \\
INGP-Big~\cite{muller2022instant} &
0.512 & 0.486 & 0.701 & 0.594 & 0.542 &
0.871 & 0.817 & 0.858 & 0.906 & 0.699 \\
Mip-NeRF 360~\cite{barron2022mip} &
0.685 & 0.583 & 0.813 & 0.744 & 0.632 &
0.913 & 0.894 & 0.920 & 0.941 & 0.792 \\
3DGS~\cite{kerbl20233d} &
\cc{1}0.771 & 0.605 & \cc{2}0.868 & \cc{2}0.775 & \cc{1}0.638 &
0.914 & 0.905 & 0.922 & 0.938 & 0.815 \\
Mip-Splatting~\cite{yu2023mip} &
\cc{3}0.767 & \cc{1}0.608 & \cc{2}0.868 & \cc{1}0.776 & \cc{2}0.636 &
\cc{3}0.920 & \cc{2}0.909 & \cc{1}0.928 & \cc{1}0.943 & \cc{1}0.817 \\
\hline
Ours &
0.763 & \cc{3}0.606 & 0.866 & 0.772 & 0.633 &
\cc{1}0.921 & \cc{1}0.910 & \cc{1}0.928 & \cc{1}0.943 & \cc{3}0.816 \\
Ours + 3D filter &
\cc{2}0.769 & \cc{1}0.608 & \cc{1}0.869 & \cc{3}0.773 & \cc{2}0.636 &
\cc{1}0.921 & \cc{2}0.909 & \cc{1}0.928 & \cc{1}0.943 & \cc{1}0.817 \\
\Xhline{3\arrayrulewidth}
3DGS*~\cite{kerbl20233d} &
0.777 & 0.620 & 0.871 & 0.784 & 0.655 &
0.927 & 0.916 & 0.933 & 0.948 & 0.825 \\
Mip-Splatting*~\cite{yu2023mip} &
0.780 & 0.623 & 0.875 & 0.786 & 0.655 &
0.928 & 0.916 & 0.933 & 0.948 & 0.827 \\
Ours* &
0.777 & 0.623 & 0.873 & 0.783 & 0.651 &
0.929 & 0.917 & 0.933 & 0.943 & 0.826 \\
Ours + 3D filter* &
0.780 & 0.624 & 0.876 & 0.785 & 0.654 &
0.929 & 0.917 & 0.934 & 0.949 & 0.827 \\
\multicolumn{11}{c}{} \\
& \multicolumn{10}{c}{\textbf{LPIPS} $\downarrow$} \\
& bicycle & flowers & garden & stump & treehill
& room & counter & kitchen & bonsai & Avg. \\
\hline
Plenoxels~\cite{fridovich2022plenoxels} &
0.506 & 0.521 & 0.386 & 0.503 & 0.540 &
0.419 & 0.441 & 0.447 & 0.398 & 0.463 \\
INGP-Base~\cite{muller2022instant} &
0.487 & 0.481 & 0.312 & 0.450 & 0.489 &
0.301 & 0.342 & 0.254 & 0.227 & 0.371 \\
INGP-Big~\cite{muller2022instant} &
0.446 & 0.441 & 0.257 & 0.421 & 0.450 &
0.261 & 0.306 & 0.195 & \cc{2}0.205 & 0.331 \\
Mip-NeRF 360~\cite{barron2022mip} &
0.301 & 0.344 & 0.170 & 0.261 & 0.339 &
0.221 & 0.204 & \cc{1}0.127 & \cc{1}0.176 & 0.237 \\
3DGS~\cite{kerbl20233d} &
\cc{1}0.205 & \cc{1}0.336 & \cc{1}0.103 & \cc{1}0.210 & \cc{1}0.317 &
\cc{1}0.220 & 0.204 & 0.129 & \cc{2}0.205 & \cc{1}0.214 \\
Mip-Splatting~\cite{yu2023mip} &
0.213 & \cc{3}0.340 & \cc{2}0.108 & \cc{2}0.216 & 0.329 &
\cc{3}0.221 & \cc{2}0.201 & \cc{1}0.127 & 0.208 & \cc{3}0.218 \\
\hline
Ours &
\cc{3}0.212 & \cc{1}0.336 & 0.110 & \cc{3}0.218 & \cc{3}0.328 &
\cc{3}0.221 & \cc{1}0.200 & \cc{1}0.127 & 0.206 & \cc{2}0.217 \\
Ours + 3D filter &
\cc{2}0.211 & \cc{3}0.340 & \cc{2}0.108 & \cc{3}0.218 & \cc{2}0.327 &
\cc{1}0.220 & \cc{3}0.202 & \cc{1}0.127 & 0.207 & \cc{3}0.218 \\
\Xhline{3\arrayrulewidth}
3DGS*~\cite{kerbl20233d} &
0.205 & 0.329 & 0.103 & 0.208 & 0.318 &
0.192 & 0.178 & 0.113 & 0.174 & 0.202 \\
Mip-Splatting*~\cite{yu2023mip} &
0.206 & 0.331 & 0.103 & 0.209 & 0.320 &
0.192 & 0.179 & 0.113 & 0.173 & 0.203 \\
Ours* &
0.207 & 0.329 & 0.105 & 0.210 & 0.320 &
0.194 & 0.180 & 0.114 & 0.176 & 0.204 \\
Ours + 3D filter* &
0.206 & 0.333 & 0.104 & 0.210 & 0.321 &
0.194 & 0.181 & 0.114 & 0.177 & 0.204 \\
\end{tabular}
}
\caption{\textbf{Quantative results of Single-Scale Training and Single-Scale Testing on the Mip-NeRF 360~\cite{barron2022mip} dataset.} All methods are trained on full-resolution images and tested on the same-resolution images.}
\label{tab:single_360}
\vspace{-0.6cm}
\end{table}

\subsection{Single-scale Training and Single-scale Testing on Scene Datasets.}
\label{sec:object_multi}
We evaluate our Analytic-Splatting against other methods on complex scene datasets (\ie Mip-NeRF 360~\cite{barron2022mip}, Tanks\&Temples~\cite{knapitsch2017tanks}, and Deep Blending~\cite{hedman2018deep}) under the single-scale training and testing setting, which is the most widely used setting. The overall results are shown in \cref{tab:single_scene_summary}, our method shows great generalization across different datasets and almost outperforms other methods.

Moreover, we report per-scene metrics for Tanks\&Temples~\cite{knapitsch2017tanks} and Deep Blending~\cite{hedman2018deep} in \cref{tab:tan_db_single}, and for Mip-NeRF 360~\cite{barron2022mip} in \cref{tab:single_360}.

For Mip-NeRF 360, images from indoor and outdoor scenes are downsampled by $2\times$ and $4\times$, respectively, as full-resolution input for training and testing. The official dataset provides the downsampled images and stores them in different folders.
Specifically, the results reported in Mip-Splatting did not use officially provided downsampled images as input, but use the bicubically downsampled images as input. The quantitative results in \cref{tab:single_360} show these two downsampling schemes greatly affect the metrics. Therefore, for fairness, we mark the methods that use bicubically downsampled images for training with * (\ie 3DGS*, Mip-Splatting*, and Ours*). And the remaining methods without * marks use officially provided downsampled images for training.

\subsection{Multi-scale Training and Multi-scale Testing on the Multi-scale Blender Synthetic Dataset}
\label{sec:scene_single}
\begin{figure}[!t]
\centering
\scalebox{0.95}{
\includegraphics[width=\textwidth]{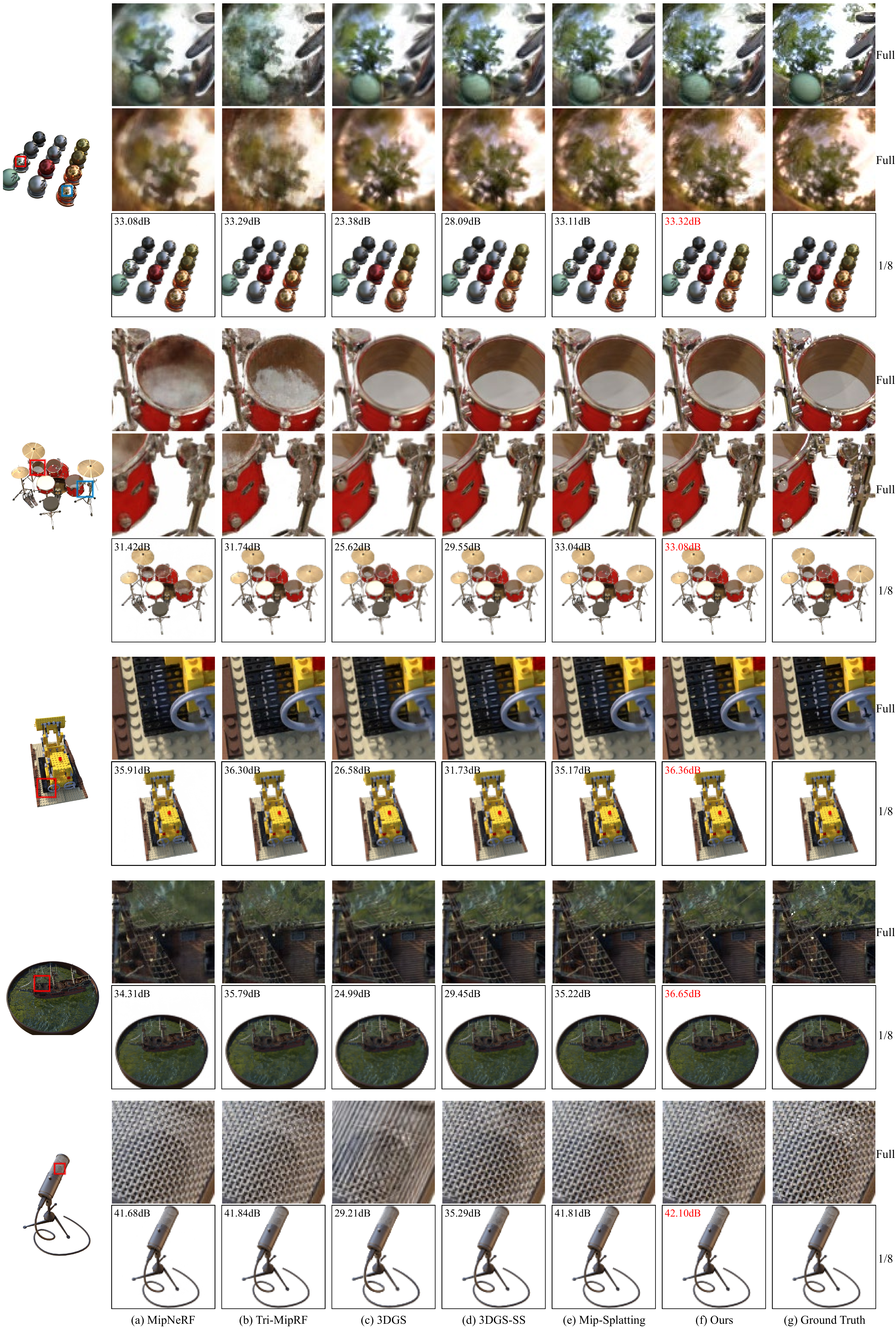}
}
\vspace{-0.2cm}
\caption{
    \textbf{Qualitative comparison of full-resolution and low-resolution ($\nicefrac{1}{8}$) on Multi-Scale Blender~\cite{barron2021mip}.} All methods are trained on images only from the training set with downsampling rates covering [1, 2, 4, 8]. Our method can better overcome the artifacts in 3DGS with better fidelity of details.}
\label{fig:supp_object_comparison}
\vspace{-.8cm}
\end{figure}

We evaluate our Analytic-Splatting against other cutting-edge methods on the Blender Synthetic dataset under the multi-scale training and testing setting. Since per-resolution metrics have been mentioned in the main paper, we report per-object metrics in \cref{tab:blender_multi_per} for more comprehensive comparisons. More qualitative results are shown in \cref{fig:supp_object_comparison}. Our method almost surpasses other methods and performs better than other methods in terms of detail capturing and anti-aliasing.
We further provide our method's per-resolution and per-object metrics in \cref{tab:blender_multi_my} so that subsequent methods can better refer to our results.

\begin{table}[!b]
\centering
\resizebox{0.95\linewidth}{!}{
\begin{tabular}
{l|c c c c c c c c|c}
& \multicolumn{9}{c}{\textbf{PSNR} $\uparrow$} \\
& chair & drums & ficus & hotdog
& lego & materials & mic & ship & Avg. \\
\hline
Full Res.
& 35.76 & 26.16 & 35.76 & 37.54 
& 35.06 & 29.59 & 35.18 & 30.76 & 33.22 \\
\nicefrac{1}{2} Res.
& 38.48 & 27.30 & 36.46 & 39.46 
& 36.46 & 31.05 & 37.61 & 32.54 & 34.92 \\
\nicefrac{1}{4} Res.
& 39.58 & 28.61 & 36.21 & 40.64 
& 36.53 & 32.77 & 39.33 & 34.12 & 35.97 \\
\nicefrac{1}{8} Res.
& 39.24 & 29.87 & 36.01 & 40.24 
& 34.93 & 33.54 & 39.02 & 35.10 & 35.99 \\
All.
& 38.26 & 27.98 & 36.11 & 39.47 
& 35.75 & 31.74 & 37.78 & 33.13 & 35.03 \\
\multicolumn{10}{c}{} \\
& \multicolumn{9}{c}{\textbf{SSIM} $\uparrow$} \\
& chair & drums & ficus & hotdog
& lego & materials & mic & ship & Avg. \\
\hline
Full Res.
& 0.986 & 0.952 & 0.988 & 0.984 
& 0.980 & 0.958 & 0.990 & 0.901 & 0.967 \\
\nicefrac{1}{2} Res.
& 0.993 & 0.960 & 0.993 & 0.990
& 0.989 & 0.974 & 0.994 & 0.926 & 0.977 \\
\nicefrac{1}{4} Res.
& 0.995 & 0.952 & 0.988 & 0.984 
& 0.992 & 0.986 & 0.995 & 0.948 & 0.984 \\
\nicefrac{1}{8} Res.
& 0.995 & 0.977 & 0.994 & 0.995 
& 0.992 & 0.992 & 0.997 & 0.967 & 0.989 \\
All.
& 0.992 & 0.964 & 0.992 & 0.991 
& 0.988 & 0.977 & 0.994 & 0.936 & 0.979 \\
\multicolumn{10}{c}{} \\
& \multicolumn{9}{c}{\textbf{LPIPS} $\downarrow$} \\
& chair & drums & ficus & hotdog
& lego & materials & mic & ship & Avg. \\
\hline
Full Res.
& 0.015 & 0.040 & 0.011 & 0.023 
& 0.020 & 0.039 & 0.007 & 0.111 & 0.033 \\
\nicefrac{1}{2} Res.
& 0.007 & 0.029 & 0.006 & 0.011 
& 0.009 & 0.018 & 0.004 & 0.065 & 0.019 \\
\nicefrac{1}{4} Res.
& 0.006 & 0.026 & 0.006 & 0.006 
& 0.007 & 0.010 & 0.004 & 0.036 & 0.013 \\
\nicefrac{1}{8} Res.
& 0.005 & 0.022 & 0.006 & 0.005
& 0.008 & 0.008 & 0.004 & 0.021 & 0.010 \\
All.
& 0.008 & 0.029 & 0.007 & 0.011 
& 0.011 & 0.018 & 0.005 & 0.058 & 0.018 \\
\hline
\end{tabular}
}
\vspace{0.2cm}
\caption{\textbf{Quantatitive Comparison of Analytic-Splatting against 3DGS and its variants under the $2\times$ super-resolution setting on the Mip-NeRF 360 dataset~\cite{barron2022mip}.} These methods are trained on images with down-sampling rates covering [1, 2, 4, 8].}
\label{tab:blender_multi_my}
\vspace{-0.4cm}
\end{table}

\begin{table}[!t]
\centering
\scalebox{1.0}{
\begin{tabular}{l|c c c c c c c c|c}
& \multicolumn{9}{c}{\textbf{PSNR} $\uparrow$} \\
& chair & drums & ficus & hotdog
& lego & materials & mic & ship & Avg. \\
\hline
NeRF w/o $\mathcal{L}_\text{area}$
& 29.92 & 23.27 & 27.15 & 32.00
& 27.75 & 26.30 & 28.40 & 26.46 & 27.66 \\
NeRF~\cite{mildenhall2020nerf}
& 33.39 & 25.87 & 30.37 & 35.64
& 31.65 & 30.18 & 32.60 & 30.09 & 31.23 \\
MipNeRF~\cite{barron2021mip}
& 37.14 & 27.02 & 33.19 & \cc{2}39.31
& \cc{2}35.74 & \cc{1}32.56 & \cc{1}38.04 & \cc{2}33.08 & 34.51 \\
\hline
Plenoxels~\cite{fridovich2022plenoxels}
& 32.79 & 25.25 & 30.28 & 34.65
& 31.26 & 28.33 & 31.53 & 28.59 & 30.34 \\
TensoRF~\cite{chen2022tensorf}
& 32.47 & 25.37 & 31.16 & 34.96
& 31.73 & 28.53 & 31.48 & 29.08 & 30.60 \\
Instant-NGP~\cite{muller2022instant}
& 32.95 & 26.43 & 30.41 & 35.87
& 31.83 & 29.31 & 32.58 & 30.23 & 31.20 \\
Tri-MipRF~\cite{hu2023tri}
& \cc{2}37.67 & 27.35 & 33.57 & 38.78
& 35.72 & 31.42 & 37.63 & 32.74 & 34.36 \\
3DGS~\cite{kerbl20233d}
& 32.73 & 25.30 & 29.00 & 35.03
& 29.44 & 27.13 & 31.17 & 28.33 & 29.77 \\
3DGS-SS~\cite{kerbl20233d}
& 35.62 & 27.02 & 33.12 & 37.46
& 33.27 & 29.90 & 34.69 & 30.63 & 32.71 \\
Mip-Splatting~\cite{yu2023mip}
& 37.48 & \cc{3}27.74 & \cc{3}34.71 & 39.15
& 35.07 & \cc{2}31.88 & \cc{3}37.68 & 32.80 & \cc{3}34.56 \\
\hline
Ours
& \cc{1}38.26 & \cc{1}27.98 & \cc{1}36.11 & \cc{1}39.47
& \cc{1}35.75 & 31.74 & \cc{2}37.78 & \cc{1}33.13 & \cc{1}35.03 \\
Ours + 3D filter
& \cc{3}37.53 & \cc{2}27.77 & \cc{2}35.85 & \cc{3}39.17
& \cc{3}35.26 & \cc{3}31.80 & 37.61 & \cc{3}32.95 & \cc{2}34.74 \\
\multicolumn{10}{c}{} \\
& \multicolumn{9}{c}{\textbf{SSIM} $\uparrow$} \\
& chair & drums & ficus & hotdog
& lego & materials & mic & ship & Avg. \\
\hline
NeRF w/o $\mathcal{L}_\text{area}$
& 0.944 & 0.891 & 0.942 & 0.959 & 0.926 & 0.934 & 0.958 & 0.861 & 0.927 \\
NeRF~\cite{mildenhall2020nerf}
& 0.971 & 0.932 & 0.971 & 0.979 & 0.965 & 0.967 & 0.980 & 0.900 & 0.958 \\
MipNeRF~\cite{barron2021mip}
& 0.988 & 0.945 & 0.984 & 0.988
& 0.984 & \cc{2}0.977 & 0.993 & 0.922 & 0.973 \\
\hline
Plenoxels~\cite{fridovich2022plenoxels}
& 0.968 & 0.929 & 0.972 & 0.976 & 0.964 & 0.959 & 0.979 & 0.892 & 0.955 \\
TensoRF~\cite{chen2022tensorf}
& 0.967 & 0.930 & 0.972 & 0.976 & 0.964 & 0.959 & 0.979 & 0.892 & 0.955 \\
Instant-NGP~\cite{muller2022instant}
& 0.971 & 0.940 & 0.973 & 0.979 & 0.966 & 0.959 & 0.981 & 0.904 & 0.959 \\
Tri-MipRF~\cite{hu2023tri}
& 0.990 & 0.951 & 0.985 & 0.988
& 0.986 & 0.969 & 0.992 & 0.929 & 0.974 \\
3DGS~\cite{kerbl20233d}
& 0.976 & 0.941 & 0.968 & 0.982 & 0.964 & 0.956 & 0.979 & 0.910 & 0.960 \\
3DGS-SS~\cite{kerbl20233d}
& 0.988 & 0.958 & 0.985
& 0.988 & 0.982 & 0.973 & 0.990 & 0.928 & 0.974 \\
Mip-Splatting~\cite{yu2023mip}
& \cc{2}0.991 & \cc{2}0.963 & \cc{2}0.990 & \cc{2}0.990
& \cc{2}0.987 & \cc{1}0.978 & \cc{1}0.994 & \cc{1}0.936 & \cc{1}0.979 \\
\hline
Ours
& \cc{1}0.992 & \cc{1}0.964 & \cc{1}0.992 & \cc{1}0.991
& \cc{1}0.988 & \cc{2}0.977 & \cc{1}0.994 & \cc{1}0.936 & \cc{1}0.979 \\
Ours + 3D filter
& \cc{2}0.991 & \cc{2}0.963 & \cc{2}0.990 & \cc{2}0.990
& \cc{2}0.987 & \cc{2}0.977 & \cc{1}0.994 & \cc{1}0.936 & \cc{1}0.979 \\
\multicolumn{10}{c}{} \\
& \multicolumn{9}{c}{\textbf{LPIPS} $\downarrow$} \\
& chair & drums & ficus & hotdog
& lego & materials & mic & ship & Avg. \\
\hline
NeRF w/o $\mathcal{L}_\text{area}$
& 0.035 & 0.069 & 0.032 & 0.028
& 0.041 & 0.045 & 0.031 & 0.095 & 0.052 \\
NeRF~\cite{mildenhall2020nerf}
& 0.028 & 0.059 & 0.026 & 0.024
& 0.035 & 0.033 & 0.025 & 0.085 & 0.044 \\
MipNeRF~\cite{barron2021mip}
& 0.011 & 0.044 & 0.014 & 0.012
& 0.013 & \cc{3}0.019 & 0.007 & 0.062 & 0.026 \\
\hline
Plenoxels~\cite{fridovich2022plenoxels}
& 0.040 & 0.070 & 0.032 & 0.037
& 0.038 & 0.055 & 0.036 & 0.104 & 0.051 \\
TensoRF~\cite{chen2022tensorf}
& 0.042 & 0.070 & 0.032 & 0.037
& 0.038 & 0.055 & 0.036 & 0.104 & 0.051 \\
Instant-NGP~\cite{muller2022instant}
& 0.035 & 0.066 & 0.029 & 0.028
& 0.040 & 0.051 & 0.032 & 0.095 & 0.047 \\
Tri-MipRF~\cite{hu2023tri}
& 0.011 & 0.046 & 0.016 & 0.014
& 0.013 & 0.033 & 0.008 & 0.069 & 0.026 \\
3DGS~\cite{kerbl20233d}
& 0.025 & 0.056 & 0.030 & 0.022
& 0.038 & 0.040 & 0.023 & 0.086 & 0.040 \\
3DGS-SS~\cite{kerbl20233d}
& 0.013 & 0.036 & 0.014 & 0.014
& 0.017 & 0.023 & 0.008 & 0.068 & 0.024 \\
Mip-Splatting~\cite{yu2023mip}
& \cc{3}0.010 & \cc{2}0.031 & \cc{2}0.009 & \cc{1}0.011
& \cc{2}0.012 & \cc{1}0.018 & \cc{1}0.005 & \cc{2}0.059 & \cc{2}0.019 \\
\hline
Ours
& \cc{1}0.008 & \cc{1}0.029 & \cc{1}0.007 & \cc{1}0.011
& \cc{1}0.011 & \cc{1}0.018 & \cc{1}0.005 & \cc{1}0.058 & \cc{1}0.018 \\
Ours + 3D filter
& \cc{2}0.009 & \cc{2}0.031 & \cc{1}0.009 & \cc{1}0.011
& \cc{2}0.012 & \cc{3}0.019 & \cc{1}0.005 & \cc{2}0.059 & \cc{2}0.019 \\
\end{tabular}
}
\vspace{0.2cm}
\caption{\textbf{Quantatitive Comparison of Analytic-Splatting against several cutting-edge methods on the Multi-scale Blender Synthetic dataset~\cite{barron2021mip}.} We report the metrics for each object in this table and all methods are trained on images only from the training set with downsampling rates covering [1, 2, 4, 8].}
\label{tab:blender_multi_per}
\vspace{-0.4cm}
\end{table}

\subsection{Multi-scale Training and Multi-scale Testing on the Mip-NeRF 360 Dataset}
\label{sec:scene_multi}
\begin{figure}[!b]
\centering
\scalebox{1.0}{
\includegraphics[width=\textwidth]{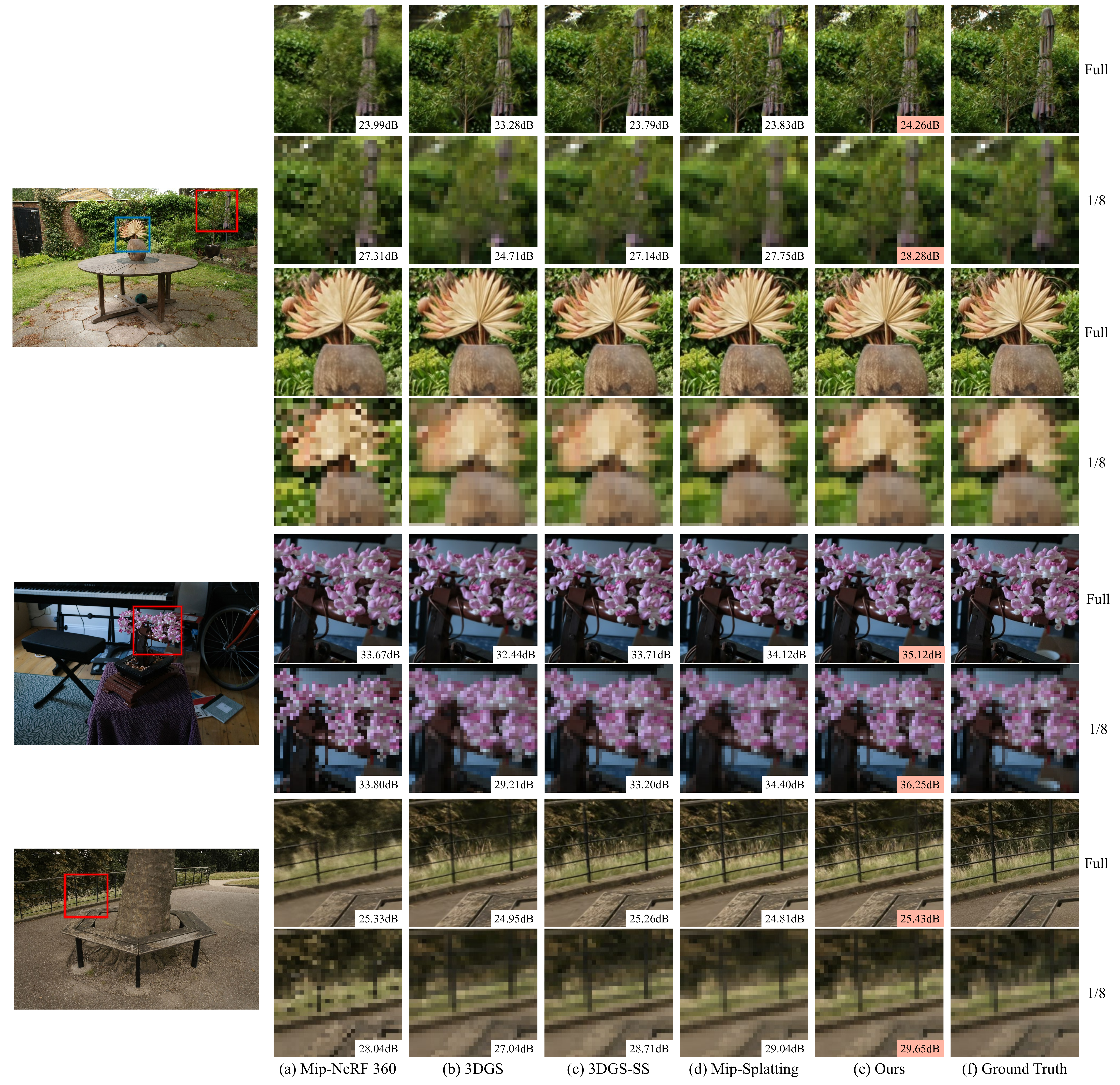}
}
\vspace{-0.2cm}
\caption{
    \textbf{Qualitative comparison of full-resolution and low-resolution ($\nicefrac{1}{8}$) on Multi-Scale Mip-NeRF 360~\cite{barron2022mip}.} All methods are trained on images with downsampling rates covering [1, 2, 4, 8]. Our method can better overcome the artifacts in 3DGS with better fidelity of details.}
\label{fig:supp_scene_comparison}
\vspace{-0.6cm}
\end{figure}

We evaluate our Analytic-Splatting against other methods on the Mip-NeRF 360 dataset under the multi-scale training and testing setting. As mentioned in \cref{sec:scene_single}, we use the officially provided downsampled images ($2\times$ for indoor scenes, and $4\times$ for outdoor scenes) as full-resolution images.
Under the multi-scale training and testing setting, we convert each full-resolution image into a set of four images for training via bicubically downsampling it by $[1\times, 2\times, 4\times, 8\times]$. Some qualitative results are shown in \cref{fig:supp_scene_comparison}. Our method performs better anti-aliasing capability and detail fidelity.

\begin{table*}[!t]
\centering
\resizebox{\linewidth}{!}{
\begin{tabular}{|l||@{\,\,\,}c@{\,\,\,}c@{\,\,\,}c@{\,\,\,}c@{\,\,\,}|@{\,\,\,}c@{\,\,\,}c@{\,\,\,}c@{\,\,\,}c@{\,\,\,}|@{\,\,\,}c@{\,\,\,}c@{\,\,\,}c@{\,\,\,}c@{\,\,\,}|@{\,\,\,}c@{\,\,\,}c@{\,\,\,}c@{\,\,\,}c@{\,\,\,}|@{\,\,\,}c@{\,\,\,}c@{\,\,\,}c@{\,\,\,}c@{\,\,\,}|}
\multicolumn{21}{c}{Outdoors} \\
\hline
\multirow{2}*{\textbf{PSNR}$\uparrow$}
& \multicolumn{4}{c@{\,\,\,}|@{\,\,\,}}{bicycle} 
& \multicolumn{4}{c@{\,\,\,}|@{\,\,\,}}{flowers}
& \multicolumn{4}{c@{\,\,\,}|@{\,\,\,}}{garden}
& \multicolumn{4}{c@{\,\,\,}|@{\,\,\,}}{stump}
& \multicolumn{4}{c@{\,\,\,}|}{treehill}  \\
& $1\times$ & $2\times$ & $4\times$ & $8\times$
& $1\times$ & $2\times$ & $4\times$ & $8\times$
& $1\times$ & $2\times$ & $4\times$ & $8\times$
& $1\times$ & $2\times$ & $4\times$ & $8\times$
& $1\times$ & $2\times$ & $4\times$ & $8\times$ \\
\hline
Mip-NeRF 360~\cite{barron2022mip}
& 24.51 & 26.93 & 28.53 & 29.24
& 21.64 & 23.90 & 26.01 & 27.35
& 26.71 & 29.59 & 31.35 & 32.52
& 26.27 & 27.68 & 28.82 & 29.27
& 22.93 & 24.63 & 26.06 & 27.12 \\
Mip-NeRF 360 + iNGP
& 24.61 & 26.98 & 26.69 & 24.50
& 21.93 & 24.14 & 24.90 & 23.19
& 26.48 & 29.06 & 27.54 & 24.85
& 26.41 & 27.63 & 27.62 & 25.64
& 23.19 & 24.86 & 25.55 & 24.81 \\
Zip-NeRF~\cite{barron2023zip}
& 25.57 & 28.25 & 30.20 & 31.37
& 22.37 & 24.91 & 27.51 & 29.50
& 27.71 & 30.53 & 32.60 & 33.83
& 27.17 & 28.62 & 30.30 & 31.73
& 23.63 & 25.47 & 27.27 & 28.84 \\
\hline
3DGS~\cite{kerbl20233d}
& 24.19 & 26.23 & 26.46 & 25.83
& 20.96 & 23.07 & 24.54 & 23.93
& 26.16 & 28.54 & 29.13 & 28.66
& 25.84 & 27.24 & 27.96 & 27.64
& 22.50 & 24.13 & 25.31 & 25.35 \\
3DGS-SS
& 20.96 & 27.16 & 28.25 & 27.95
& 21.51 & 23.07 & 25.89 & 26.11
& 26.81 & 28.54 & 30.71 & 30.91
& 26.56 & 27.24 & 29.36 & 29.69
& 22.67 & 24.13 & 25.81 & 26.62 \\
Mip-Splatting~\cite{yu2023mip}
& 24.90 & 27.24 & 28.81 & 29.10
& 21.42 & 23.75 & 26.13 & 28.19
& 26.69 & 29.37 & 30.92 & 31.66
& 26.49 & 27.94 & 29.58 & 31.17
& 22.52 & 24.36 & 26.22 & 27.84 \\
\hline
Ours
& 25.20 & 27.39 & 28.97 & 29.81
& 21.76 & 24.05 & 26.52 & 28.57
& 27.13 & 29.53 & 31.20 & 32.19
& 26.74 & 28.15 & 29.90 & 31.50
& 22.70 & 24.41 & 26.13 & 27.67 \\
Ours + 3D filter
& 25.32 & 27.50 & 29.04 & 29.88
& 21.79 & 24.04 & 26.49 & 28.52
& 27.19 & 29.52 & 31.14 & 32.08
& 26.80 & 28.15 & 29.87 & 31.48
& 22.72 & 24.33 & 25.97 & 27.49 \\
\hline
\multicolumn{10}{c}{} \\
\hline
\multirow{2}*{\textbf{SSIM}$\uparrow$}
& \multicolumn{4}{c@{\,\,\,}|@{\,\,\,}}{bicycle} 
& \multicolumn{4}{c@{\,\,\,}|@{\,\,\,}}{flowers}
& \multicolumn{4}{c@{\,\,\,}|@{\,\,\,}}{garden}
& \multicolumn{4}{c@{\,\,\,}|@{\,\,\,}}{stump}
& \multicolumn{4}{c@{\,\,\,}|}{treehill}  \\
& $1\times$ & $2\times$ & $4\times$ & $8\times$
& $1\times$ & $2\times$ & $4\times$ & $8\times$
& $1\times$ & $2\times$ & $4\times$ & $8\times$
& $1\times$ & $2\times$ & $4\times$ & $8\times$
& $1\times$ & $2\times$ & $4\times$ & $8\times$ \\ \hline
Mip-NeRF 360~\cite{barron2022mip}
& 0.666 & 0.815 & 0.890 & 0.912
& 0.567 & 0.727 & 0.834 & 0.881
& 0.791 & 0.903 & 0.939 & 0.959
& 0.726 & 0.819 & 0.874 & 0.882
& 0.615 & 0.748 & 0.839 & 0.893 \\
Mip-NeRF 360 + iNGP
& 0.673 & 0.825 & 0.857 & 0.773
& 0.592 & 0.742 & 0.805 & 0.763
& 0.786 & 0.904 & 0.864 & 0.767
& 0.748 & 0.830 & 0.849 & 0.770
& 0.616 & 0.736 & 0.785 & 0.762 \\
Zip-NeRF~\cite{barron2023zip}
& 0.758 & 0.872 & 0.926 & 0.948
& 0.635 & 0.774 & 0.864 & 0.914
& 0.850 & 0.929 & 0.960 & 0.974
& 0.791 & 0.865 & 0.914 & 0.939
& 0.671 & 0.780 & 0.865 & 0.922 \\
\hline
3DGS~\cite{kerbl20233d}
& 0.703 & 0.831 & 0.864 & 0.855
& 0.545 & 0.690 & 0.784 & 0.795
& 0.810 & 0.904 & 0.920 & 0.919
& 0.729 & 0.810 & 0.850 & 0.830
& 0.602 & 0.725 & 0.811 & 0.839 \\
3DGS-SS
& 0.736 & 0.849 & 0.902 & 0.911
& 0.585 & 0.724 & 0.824 & 0.864
& 0.834 & 0.920 & 0.947 & 0.956
& 0.763 & 0.837 & 0.887 & 0.891
& 0.620 & 0.738 & 0.830 & 0.878 \\
Mip-Splatting~\cite{yu2023mip}
& 0.739 & 0.849 & 0.912 & 0.940
& 0.591 & 0.724 & 0.825 & 0.891
& 0.832 & 0.917 & 0.949 & 0.966
& 0.768 & 0.837 & 0.892 & 0.932
& 0.619 & 0.737 & 0.839 & 0.905 \\
\hline
Ours
& 0.750 & 0.855 & 0.913 & 0.940
& 0.601 & 0.732 & 0.834 & 0.898
& 0.847 & 0.921 & 0.951 & 0.966
& 0.772 & 0.842 & 0.899 & 0.933
& 0.627 & 0.739 & 0.835 & 0.899 \\
Ours + 3D filter
& 0.754 & 0.858 & 0.915 & 0.941
& 0.602 & 0.733 & 0.835 & 0.897
& 0.848 & 0.922 & 0.951 & 0.965
& 0.773 & 0.842 & 0.898 & 0.933
& 0.628 & 0.739 & 0.834 & 0.898 \\
\hline
\multicolumn{10}{c}{} \\
\hline
\multirow{2}*{\textbf{LPIPS}$\downarrow$}
& \multicolumn{4}{c@{\,\,\,}|@{\,\,\,}}{bicycle} 
& \multicolumn{4}{c@{\,\,\,}|@{\,\,\,}}{flowers}
& \multicolumn{4}{c@{\,\,\,}|@{\,\,\,}}{garden}
& \multicolumn{4}{c@{\,\,\,}|@{\,\,\,}}{stump}
& \multicolumn{4}{c@{\,\,\,}|}{treehill}  \\
& $1\times$ & $2\times$ & $4\times$ & $8\times$
& $1\times$ & $2\times$ & $4\times$ & $8\times$
& $1\times$ & $2\times$ & $4\times$ & $8\times$
& $1\times$ & $2\times$ & $4\times$ & $8\times$
& $1\times$ & $2\times$ & $4\times$ & $8\times$ \\
\hline
Mip-NeRF 360~\cite{barron2022mip}
& 0.322 & 0.177 & 0.089 & 0.066
& 0.367 & 0.215 & 0.114 & 0.071
& 0.194 & 0.079 & 0.045 & 0.029
& 0.279 & 0.171 & 0.114 & 0.107
& 0.362 & 0.236 & 0.144 & 0.096 \\
Mip-NeRF 360 + iNGP
& 0.313 & 0.166 & 0.128 & 0.169
& 0.344 & 0.192 & 0.124 & 0.137
& 0.192 & 0.079 & 0.107 & 0.176
& 0.254 & 0.156 & 0.137 & 0.180
& 0.344 & 0.223 & 0.171 & 0.182 \\
Zip-NeRF~\cite{barron2023zip}
& 0.222 & 0.112 & 0.061 & 0.041
& 0.287 & 0.156 & 0.083 & 0.050
& 0.129 & 0.055 & 0.030 & 0.020
& 0.206 & 0.122 & 0.077 & 0.057
& 0.263 & 0.163 & 0.103 & 0.068 \\
\hline
3DGS~\cite{kerbl20233d}
& 0.295 & 0.180 & 0.113 & 0.100
& 0.404 & 0.288 & 0.184 & 0.147
& 0.197 & 0.085 & 0.059 & 0.054
& 0.284 & 0.186 & 0.130 & 0.134
& 0.398 & 0.279 & 0.185 & 0.140 \\
3DGS-SS
& 0.257 & 0.144 & 0.078 & 0.067
& 0.365 & 0.251 & 0.154 & 0.107
& 0.165 & 0.065 & 0.038 & 0.032
& 0.243 & 0.150 & 0.097 & 0.088
& 0.364 & 0.250 & 0.160 & 0.110 \\
Mip-Splatting~\cite{yu2023mip}
& 0.258 & 0.153 & 0.083 & 0.050
& 0.363 & 0.262 & 0.165 & 0.093
& 0.169 & 0.073 & 0.045 & 0.027
& 0.233 & 0.156 & 0.106 & 0.072
& 0.373 & 0.265 & 0.172 & 0.105 \\
\hline
Ours
& 0.239 & 0.134 & 0.070 & 0.047
& 0.344 & 0.239 & 0.148 & 0.084
& 0.141 & 0.061 & 0.037 & 0.027
& 0.224 & 0.138 & 0.087 & 0.062
& 0.349 & 0.244 & 0.160 & 0.098 \\
Ours + 3D filter
& 0.237 & 0.134 & 0.070 & 0.048
& 0.347 & 0.241 & 0.147 & 0.086
& 0.141 & 0.061 & 0.037 & 0.027
& 0.224 & 0.140 & 0.089 & 0.063
& 0.350 & 0.245 & 0.161 & 0.100 \\
\hline
\end{tabular}
}
\resizebox{\linewidth}{!}{
\begin{tabular}{|l||@{\,\,\,}c@{\,\,\,}c@{\,\,\,}c@{\,\,\,}c@{\,\,\,}|@{\,\,\,}c@{\,\,\,}c@{\,\,\,}c@{\,\,\,}c@{\,\,\,}|@{\,\,\,}c@{\,\,\,}c@{\,\,\,}c@{\,\,\,}c@{\,\,\,}|@{\,\,\,}c@{\,\,\,}c@{\,\,\,}c@{\,\,\,}c@{\,\,\,}|}
\multicolumn{17}{c}{} \\
\multicolumn{17}{c}{} \\
\multicolumn{17}{c}{Indoors} \\
\hline
\multirow{2}*{\textbf{PSNR}$\uparrow$}
& \multicolumn{4}{c@{\,\,\,}|@{\,\,\,}}{room}
& \multicolumn{4}{c@{\,\,\,}|@{\,\,\,}}{counter}
& \multicolumn{4}{c@{\,\,\,}|@{\,\,\,}}{kitchen}
& \multicolumn{4}{c@{\,\,\,}|}{bonsai}  \\
& $1\times$ & $2\times$ & $4\times$ & $8\times$
& $1\times$ & $2\times$ & $4\times$ & $8\times$
& $1\times$ & $2\times$ & $4\times$ & $8\times$
& $1\times$ & $2\times$ & $4\times$ & $8\times$ \\
\hline
Mip-NeRF 360~\cite{barron2022mip}
& 31.44 & 32.53 & 33.17 & 32.96
& 29.30 & 30.12 & 30.81 & 30.52
& 31.90 & 33.39 & 34.69 & 34.92
& 32.85 & 33.97 & 34.63 & 33.80 \\
Mip-NeRF 360 + iNGP
& 30.93 & 31.83 & 31.66 & 29.52
& 24.30 & 24.66 & 24.81 & 24.06
& 30.13 & 31.25 & 29.85 & 26.14
& 30.20 & 30.90 & 30.39 & 27.49 \\
Zip-NeRF~\cite{barron2023zip}
& 32.20 & 33.33 & 34.12 & 34.26
& 29.17 & 29.93 & 30.70 & 31.11
& 32.33 & 33.76 & 35.20 & 35.71
& 34.08 & 35.25 & 36.18 & 36.32 \\
\hline
3DGS~\cite{kerbl20233d}
& 30.53 & 31.42 & 31.46 & 29.81
& 28.25 & 28.91 & 29.21 & 27.66
& 29.90 & 31.04 & 31.50 & 29.57
& 30.63 & 31.42 & 31.02 & 31.58 \\
3DGS-SS
& 31.12 & 32.13 & 32.75 & 32.12
& 28.81 & 29.47 & 30.16 & 29.85
& 30.84 & 32.05 & 33.13 & 32.59
& 31.57 & 32.44 & 32.96 & 31.58 \\
Mip-Splatting~\cite{yu2023mip}
& 31.32 & 32.26 & 32.79 & 32.88
& 28.91 & 29.50 & 30.03 & 30.32
& 31.11 & 32.05 & 32.33 & 32.79
& 31.48 & 32.21 & 32.26 & 31.97 \\
\hline
Ours
& 31.26 & 32.23 & 32.92 & 33.07
& 29.03 & 29.65 & 30.39 & 30.90
& 31.44 & 32.57 & 33.56 & 33.78
& 32.12 & 32.92 & 33.60 & 33.39 \\
Ours + 3D filter
& 31.32 & 32.28 & 32.98 & 33.13
& 29.03 & 29.64 & 30.38 & 30.88
& 31.07 & 32.15 & 33.09 & 33.25
& 31.97 & 32.73 & 33.42 & 33.26 \\
\hline
\multicolumn{10}{c}{} \\
\hline
\multirow{2}*{\textbf{SSIM}$\uparrow$}
& \multicolumn{4}{c@{\,\,\,}|@{\,\,\,}}{room}
& \multicolumn{4}{c@{\,\,\,}|@{\,\,\,}}{counter}
& \multicolumn{4}{c@{\,\,\,}|@{\,\,\,}}{kitchen}
& \multicolumn{4}{c@{\,\,\,}|}{bonsai}  \\
& $1\times$ & $2\times$ & $4\times$ & $8\times$
& $1\times$ & $2\times$ & $4\times$ & $8\times$
& $1\times$ & $2\times$ & $4\times$ & $8\times$
& $1\times$ & $2\times$ & $4\times$ & $8\times$ \\
\hline
Mip-NeRF 360~\cite{barron2022mip}
& 0.906 & 0.944 & 0.963 & 0.967
& 0.887 & 0.916 & 0.936 & 0.942
& 0.916 & 0.949 & 0.968 & 0.975
& 0.935 & 0.959 & 0.969 & 0.968 \\
Mip-NeRF 360 + iNGP
& 0.904 & 0.941 & 0.950 & 0.932
& 0.816 & 0.837 & 0.843 & 0.819
& 0.903 & 0.938 & 0.904 & 0.773
& 0.920 & 0.941 & 0.937 & 0.874 \\
Zip-NeRF~\cite{barron2023zip}
& 0.921 & 0.955 & 0.971 & 0.977
& 0.899 & 0.926 & 0.944 & 0.955
& 0.926 & 0.956 & 0.975 & 0.982
& 0.947 & 0.968 & 0.978 & 0.980 \\
\hline
3DGS~\cite{kerbl20233d}
& 0.903 & 0.936 & 0.952 & 0.947
& 0.886 & 0.912 & 0.928 & 0.920
& 0.907 & 0.941 & 0.956 & 0.950
& 0.924 & 0.948 & 0.954 & 0.938 \\
3DGS-SS
& 0.911 & 0.943 & 0.962 & 0.965
& 0.898 & 0.922 & 0.941 & 0.947
& 0.919 & 0.949 & 0.968 & 0.938
& 0.933 & 0.955 & 0.967 & 0.966 \\
Mip-Splatting~\cite{yu2023mip}
& 0.913 & 0.944 & 0.962 & 0.969
& 0.899 & 0.921 & 0.936 & 0.949
& 0.920 & 0.948 & 0.960 & 0.974
& 0.935 & 0.954 & 0.960 & 0.966 \\
\hline
Ours
& 0.914 & 0.946 & 0.964 & 0.971
& 0.902 & 0.924 & 0.942 & 0.955
& 0.924 & 0.951 & 0.967 & 0.974
& 0.939 & 0.958 & 0.969 & 0.972 \\
Ours + 3D filter
& 0.915 & 0.946 & 0.964 & 0.971
& 0.903 & 0.924 & 0.943 & 0.955
& 0.924 & 0.951 & 0.967 & 0.973
& 0.939 & 0.958 & 0.969 & 0.972 \\
\hline
\multicolumn{10}{c}{} \\
\hline
\multirow{2}*{\textbf{LPIPS}$\downarrow$}
& \multicolumn{4}{c@{\,\,\,}|@{\,\,\,}}{room}
& \multicolumn{4}{c@{\,\,\,}|@{\,\,\,}}{counter}
& \multicolumn{4}{c@{\,\,\,}|@{\,\,\,}}{kitchen}
& \multicolumn{4}{c@{\,\,\,}|}{bonsai}  \\
& $1\times$ & $2\times$ & $4\times$ & $8\times$
& $1\times$ & $2\times$ & $4\times$ & $8\times$
& $1\times$ & $2\times$ & $4\times$ & $8\times$
& $1\times$ & $2\times$ & $4\times$ & $8\times$ \\
\hline
Mip-NeRF 360~\cite{barron2022mip}
& 0.227 & 0.101 & 0.052 & 0.042
& 0.216 & 0.114 & 0.068 & 0.059
& 0.134 & 0.063 & 0.033 & 0.023
& 0.185 & 0.065 & 0.033 & 0.033 \\
Mip-NeRF 360 + iNGP
& 0.220 & 0.105 & 0.072 & 0.095
& 0.275 & 0.195 & 0.163 & 0.182
& 0.145 & 0.074 & 0.084 & 0.188
& 0.190 & 0.082 & 0.063 & 0.126 \\
Zip-NeRF~\cite{barron2023zip}
& 0.199 & 0.084 & 0.041 & 0.028
& 0.189 & 0.095 & 0.055 & 0.039
& 0.117 & 0.055 & 0.028 & 0.018
& 0.173 & 0.052 & 0.023 & 0.017 \\
\hline
3DGS~\cite{kerbl20233d}
& 0.254 & 0.127 & 0.066 & 0.053
& 0.235 & 0.128 & 0.078 & 0.067
& 0.159 & 0.081 & 0.046 & 0.041
& 0.234 & 0.104 & 0.056 & 0.051 \\
3DGS-SS
& 0.241 & 0.114 & 0.054 & 0.039
& 0.217 & 0.112 & 0.066 & 0.050
& 0.142 & 0.068 & 0.034 & 0.024
& 0.220 & 0.092 & 0.044 & 0.032 \\
Mip-Splatting~\cite{yu2023mip}
& 0.235 & 0.115 & 0.062 & 0.040
& 0.213 & 0.116 & 0.077 & 0.043
& 0.138 & 0.075 & 0.047 & 0.027
& 0.214 & 0.095 & 0.058 & 0.037 \\
\hline
Ours
& 0.234 & 0.111 & 0.052 & 0.035
& 0.208 & 0.109 & 0.065 & 0.045
& 0.134 & 0.065 & 0.037 & 0.028
& 0.210 & 0.088 & 0.042 & 0.029 \\
Ours + 3D filter
& 0.233 & 0.110 & 0.052 & 0.034
& 0.209 & 0.110 & 0.065 & 0.045
& 0.134 & 0.065 & 0.037 & 0.029
& 0.209 & 0.088 & 0.042 & 0.030 \\
\hline
\end{tabular}
}
\caption{\textbf{Quantative comparisons of Analyti-Splatting against several cutting-edge methods on the multi-scale Mip-NeRF 360 dataset \cite{barron2022mip, barron2023zip}.} These methods conduct multi-scale training and testing.
}
\vspace{-1.0cm}
\label{tab:scene_360_results_multi}
\end{table*}

We further provide per-resolution and per-scene metrics in \cref{tab:scene_360_results_multi}. The results of Mip-NeRF 360~\cite{barron2022mip} and Zip-NeRF~\cite{barron2023zip} are reported from the official Zip-NeRF paper~\cite{barron2023zip}. Please note that Mip-NeRF 360 and Zip-NeRF struggle with real-time rendering, and our Analytic-Splatting, like 3DGS and its variants, is capable of real-time rendering.

\subsection{Approximation Error Analysis}
\label{sec:approx_error}
In Sec. 5.1 of the main page, we study the approximation error produced by different schemes. In this analysis, we concentrate on the standard deviation $\sigma \in [0.3, 6.6]$ and the samples within the $99\%$ confidence interval (\ie $\Vert x\Vert < 3\sigma$). We plot the curve of the approximation error under different conditions.

In detail, we sample uniformly in the logarithmic coordinate system of $[0.3, 6.6]$ to provide the standard deviation samples and plot the approximation error to the standard deviation as Fig. 5(a) in the main page. For the approximation error caused by the rotation of the integral domain, to calculate the integral of the original integral domain (\cref{fig:supp_integral_pixel}) as reference results, we perform Monte Carlo sampling in the original integral domain and calculate the average response of $65536$ samples as the integration reference.

\end{document}